\documentclass[runningheads]{llncs}

\usepackage{eccv}
\usepackage{multirow}
\usepackage{array}
\usepackage{booktabs}
\usepackage{makecell}
\usepackage{adjustbox}
\usepackage{amsmath}
\usepackage{mathrsfs}
\usepackage{soul}
\usepackage{rotating}
\usepackage{caption}
\usepackage{dblfloatfix}
\usepackage[dvipsnames]{xcolor}
\usepackage{comment}
\usepackage{graphicx}
\usepackage{subcaption}
\usepackage{wrapfig}
\usepackage{gensymb}

\usepackage{eccvabbrv}

\usepackage{graphicx}
\usepackage{booktabs}

\usepackage[accsupp]{axessibility}  

\usepackage{hyperref}

\usepackage{orcidlink}

\begin{document}

\title{LaWa: Using Latent Space for In-Generation Image Watermarking} 

\titlerunning{LaWa}

\author{Ahmad Rezaei\inst{1}\thanks{Work done during an internship at Huawei Technologies Canada Co. Ltd.} \and
Mohammad Akbari\inst{2} \and
Saeed Ranjbar Alvar\inst{2} \and
Arezou Fatemi\inst{2} \and
Yong Zhang\inst{2}}

\authorrunning{A.~Rezaei et al.}

\institute{
University of British Columbia\\
\email{ahnr@mail.ubc.ca} \and
Huawei Technologies Canada Co. Ltd. \\ 
\email{\{mohammad.akbari,saeed.ranjbar.alvar1,yong.zhang3\}@huawei.com}}

\maketitle

\begin{abstract}
With generative models producing high quality images that are indistinguishable from real ones, there is growing concern regarding the malicious usage of AI-generated images. Imperceptible image watermarking is one viable solution towards such concerns. Prior watermarking methods map the image to a latent space for adding the watermark. Moreover, Latent Diffusion Models (LDM) generate the image in the latent space of a pre-trained autoencoder. We argue that this latent space can be used to integrate watermarking into the generation process. To this end, we present LaWa, an in-generation image watermarking method designed for LDMs. By using coarse-to-fine watermark embedding modules, LaWa modifies the latent space of pre-trained autoencoders and achieves high robustness against a wide range of image transformations while preserving perceptual quality of the image. We show that LaWa can also be used as a general image watermarking method. Through extensive experiments, we demonstrate that LaWa outperforms previous works in perceptual quality, robustness against attacks, and computational complexity, while having very low false positive rate. Code is available \href{https://github.com/vbdi/LaWa}{here}\footnote{\url{https://github.com/vbdi/LaWa}}.
\end{abstract}

\section{Introduction}
\label{sec:intro}

With rapid advancements in generative models, AI-generated content (AIGC) in different modalities including images~\cite{nichol2021glide, ramesh2022hierarchical, rombach2022high, saharia2022photorealistic}, video~\cite{ho2022imagen, singer2022make}, text~\cite{brown2020language, chowdhery2022palm, touvron2023llama}, and 3D~\cite{lin2023magic3d, poole2022dreamfusion} can be generated with high quality. Text-to-image diffusion models such as Stable Diffusion~\cite{rombach2022high} and DALL·E 2~\cite{ramesh2022hierarchical} are open to public and have shown stunning performance in generation of photo-realistic images that are indistinguishable from real ones. Such tools can be misused in different ways such as faking AI-generated images as human-created artworks, generating fake news, impersonation, and copyright infringement~\cite{brundage2018malicious, vincent2020online, zohny2023ethics}. Such threats raise concerns about our confidence and trust in the authenticity of photo-realistic images. Thus, responsible implementation of generative AI services needs to be considered by service providers. Specifically, two problems including detection and attribution should be addressed. In detection, model developers detect if an image is generated by their model, i.e., if the image is AI-generated or not. In attribution, the developers attribute an image to the user who generated the image using their service.

\begin{figure}[t]
  \centering
     \includegraphics[width=0.98\textwidth]{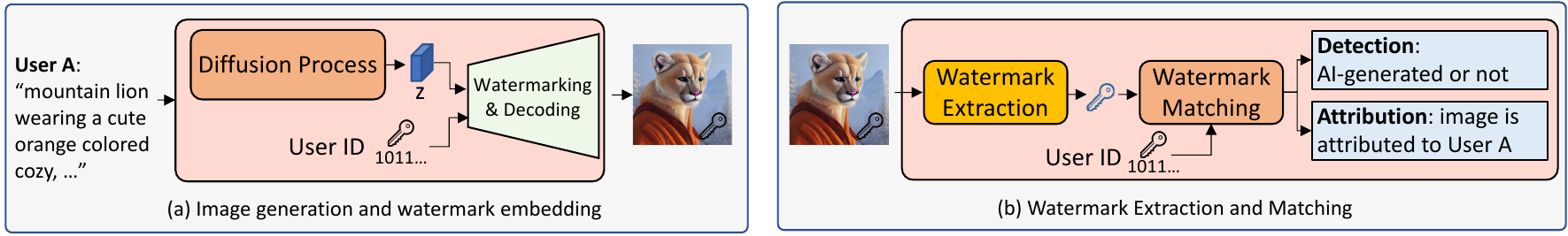}
     \vspace{-3pt}
  \caption{An overview of in-generation image watermarking for LDMs. (a) generation and watermark (i.e., User ID) embedding. (b) Watermark extraction and matching for detection and attribution.}
  \label{fig:teaser}
  \vspace{-15pt}
\end{figure}

Imperceptible image watermarking is a potential solution for the aforementioned problems. It involves embedding a message in the image without damaging its utility. A watermark extraction module is also required to accurately extract the embedded message from the watermarked image even if the it is modified. In the case of AI-generated images, the extracted message is used to detect such images and attribute them to the responsible user who created them (\cref{fig:teaser}).

For AIGC images, existing watermarking methods can be applied after the images are generated (called post-generation methods in this paper)~\cite{fernandez2022watermarking, zhang2019robust}. Such post-generation watermarking introduces an overhead on the generation service due to the extra watermark post-processing. The watermarking process generally includes embedding the watermark into a latent feature representation of an image~\cite{fernandez2022watermarking, luo2020distortion, vukotic2018deep, zhang2020udh}. The latent feature is either learned end-to-end in encoder-decoder-based watermarking methods~\cite{luo2020distortion, zhang2020udh} or it is extracted from the latent space of a pre-trained neural network~\cite{fernandez2022watermarking, vukotic2018deep}. In the case of Latent Diffusion Models (LDMs), generated images are in the latent space of a pre-trained autoencoder model. Thus, the latent features corresponding to the generation and watermarking procedures are de-coupled, which is a sub-optimal solution for combined image generation and watermarking problem.

In this paper, we propose LaWa, an in-generation watermarking method that effectively changes the latent feature of pre-trained LDMs to integrate watermarking into the generation process. Thus, generated images already conceal a watermark. Only a few prior works have addressed this problem~\cite{fernandez2023stable, wen2023tree}. Stable Signature~\cite{fernandez2023stable} is a model watermarking method for the decoder of LDMs. Since it requires fine-tuning a new decoder model for each watermark message, its application for in-generation watermarking has high computational cost and limited scalibility for an image generation service. Tree-Ring~\cite{wen2023tree} watermarking modifies the initial noise vector used for image generation. It is only applicable to deterministic sampling~\cite{nichol2021improved} and requires  image-to-noise inversion process for watermark extraction. Such process for text-guided diffusion does not always yield to the real noise vector~\cite{hertz2022prompt,mokady2023null}.


To address these limitations, LaWa incorporates novel coarse-to-fine multi-scale embedding modules into the frozen intermediate layers of the LDM decoder to ensure robustness against severe geometrical attacks. LaWa achieves high-payload in-generation watermarking with only one decoder model. Thus, it can handle many users for an image generation service with low computational cost. The focus of LaWa is on black-box watermarking, where users do not have access to the image generation model. In comparison with post-generation watermarking, we show that using the same latent features for image generation and watermarking improves the quality vs. robustness trade-off. Moreover, we show that LaWa can convert a pre-trained autoencoder into a robust general image watermarking method. In summary, our contributions are as follows:
\begin{itemize}
    \item A multi-scale latent modification mechanism for pre-trained generative autoencoders that is compatible with any autoencoder and is robust to a broad range of image modifications
    \item A simple yet effective spatial watermark coding that improves the trade-off between robustness and perceptual quality
    \item An in-generation image watermarking approach that can be used for any pre-trained LDM and any image generation task without further fine-tuning of the LDM
    \item Extending the proposed in-generation watermarking method to work as a general post-generation image watermarking technique
    \item Achieving state-of-the-art post- and in-generation results in perceptual quality, robustness, and computational complexity.
\end{itemize}
\section{Related Work}

\textbf{Detection of AI-Generated Images.} Considering the risks of AIGC, many works focus on the passive detection of generated/manipulated images. These methods are well-studied for deep-fakes using inconsistencies in generated images~\cite{chai2020makes, gragnaniello2021gan, li2018exposing} as well as generator traces in the spatial~\cite{marra2019gans, yu2019attributing} or frequency~\cite{frank2020leveraging, zhang2019detecting} domains. However, they have poor performance because they fall behind the rapid evolution of generative models. Similar approaches are proposed for diffusion models~\cite{corvi2023detection, sha2022fake}, but they are also shown to suffer from low detection accuracy and high false rate~\cite{fernandez2023stable}.

\textbf{Image Watermarking.} There are three main categories of image watermarking methods. Transform-based methods embed the watermark in a spatial~\cite{ghazanfari2011lsb++, taha2022high} or frequency~\cite{holub2012designing, holub2014universal, navas2008dwt} domain, which can achieve great imperceptibility, but have low bit extraction robustness to even minor image modifications. Encoder-decoder-based methods use the encoder to concurrently create a latent feature and add the watermark before recreating the marked image~\cite{bui2023rosteals, lee2020convolutional, luo2020distortion, zhang2020udh, zhang2019robust, zhu2018hidden, alvar2024amuse}. Despite their good robustness, encoder-decoder networks may not generalize well to images out of the training data distribution and have strong trade-off between payload and utility. The third group uses the latent space of a fixed pre-trained network to add the watermark in several iterations~\cite{fernandez2022watermarking, kishore2021fixed, vukotic2018deep, zhang2024robust, ranjbar2023nft}, which substantially increases the watermarking time. RoSteALS proposes using latent space of autoencoders for steganography~\cite{bui2023rosteals}, but it does not consider image watermarking attack constraints. The proposed solution has no robustness to geometrical image modifications. Such limitation is critically important in the AIGC application scenario as users can easily evade the watermark.

\textbf{In-Generation Image Watermarking.} Previous works on integrating watermarking and image generation are mostly for model watermarking~\cite{uchida2017embedding}. Some methods watermark the entire training dataset with one message~\cite{wu2020watermarking, yu2021artificial, zhao2023recipe} so that the output of the network carries the same message as well. Such methods are not extensible to new messages and require new training for every new message. Specifically for GANs,~\cite{yu2022responsible} proposes a solution to merge watermarking into the training procedure of the network. At the inference time, the generated images conceal a fingerprint specific to the user creating the image. This solution requires training the GAN model from scratch and is not applicable to pre-trained models.

\begin{figure*}[t]
  \centering
      \includegraphics[width=0.94\textwidth]{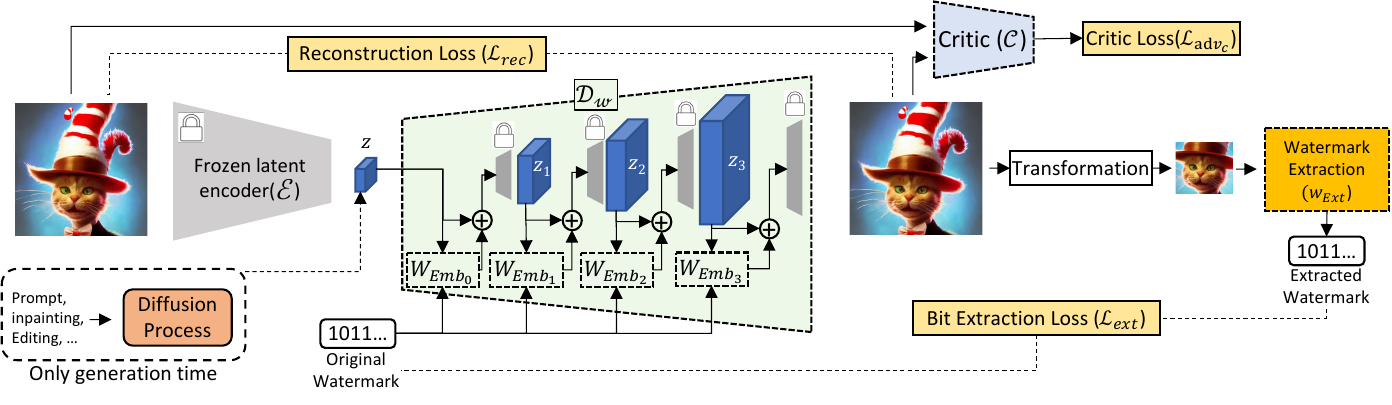}
      \vspace{-3pt}
  \caption{Overall framework including the procedure of injecting the watermark embedding modules ($W_{Emb}$) to LDM’s decoder ($\mathcal{D}$) as well as the end-to-end training process. During training, LDM's frozen encoder is used to generate latent $z$ while at inference time, $z$ is generated by the diffusion process.
  }
  \label{fig:fig2}
  \vspace{-15pt}
\end{figure*}

In-generation watermarking for diffusion models is a recent research topic in the literature~\cite{fernandez2023stable, wen2023tree}. Tree-Ring watermarking~\cite{wen2023tree} modifies the initial noise vector used in image generation. This method cannot conceal a specific bit string into the image and is only applicable to deterministic sampling~\cite{nichol2021improved} in the diffusion process. For watermark extraction, an inversion process should accurately map the watermarked image to the initial noise.~\cite{hertz2022prompt,mokady2023null} show that such inversion for text-guided diffusion does not always yield to the real noise vector, which can damage watermark extraction performance.
Stable Signature~\cite{fernandez2023stable} is a model watermarking method for the image decoder of a pre-trained LDM. If used for in-generation watermarking, a copy of the decoder model should be fine-tuned and stored for each watermark message, substantially increasing the computational cost and limiting the scalibility of this solution.


\section{Methodology}
The overall framework of how LaWa embeds and extracts messages is shown in \cref{fig:teaser} (in-generation watermarking) and Fig. 4 (image watermarking). In the embedding phase, a latent is first generated by the diffusion process. Given the watermark message (i.e., user ID), LaWa enables the pre-trained LDM’s latent decoder to simultaneously watermark and decode the latent, which results in generating the watermarked image. The watermark message can later be ex-tracted using the extraction module to be used for detection and attribution by matching the message with one of existing watermarks.

\Cref{fig:fig2} illustrates the high-level procedure of modifying the LDM’s decoder by injecting watermark embedding modules at intermediate layers of the decoder. The end-to-end training process of the added embedding and extraction networks is also shown, which will be described in details in the following subsections.

\subsection{Modified Latent Decoder}
In general, diffusion models are trained either in the image space~\cite{nichol2021glide, saharia2022photorealistic} or in a compact latent space~\cite{gu2022vector, ramesh2022hierarchical, rombach2022high} for more computational efficiency. For LDMs, the output of the diffusion process is in this latent space. Besides the diffusion process, LDMs also include an image autoencoder. The image encoder, $\mathcal{E}$, is used to downsample the input image $x \in \mathbb{R}^{H \times W \times 3}$ by a factor of $f$ to a latent feature, $z$, where $z=\mathcal{E}(x) \in \mathbb{R}^{H / f \times W / f \times C}$~\cite{rombach2022high}. The decoder model, $\mathcal{D}$, then upsamples $z$ to create the reconstructed image $\hat{x}=\mathcal{D}(\mathcal{E}(x))$. At the generation time, the diffusion process generates the latent feature $z$, which is mapped to the pixel space using the decoder: $\hat{x}=\mathcal{D}(z)$. The upsampling in $\mathcal{D}$ is a multi-step process in which latent size is scaled up by a factor of 2 after each step.

To embed the watermark information, we feed the watermark message into the latent feature before each upsampling step of $\mathcal{D}$. This results in a multiscale coarse-to-fine process of embedding the watermark into the latent space features. More specifically, assuming a $k$-bit watermark message $m \in\{0,1\}^{k}$, our watermark embedding module at $i$th upsampling step, $W_{E m b_{i}}$, accepts $m$ and the latent $z_{i}$ as input and generates a residual latent, $\delta z_{i}$, with the same size as $z_{i}: \delta z_{i}=W_{E m b_{i}}\left(z_{i}, m\right)$, where $i \in\left\{0, \ldots, \frac{f}{2}-1\right\}$ shows the upsampling step of $\mathcal{D}$. The watermarked latent, $z_{w_{i}}$ is then constructed as:
\begin{equation}
    \small
  z_{w_i} = z_i + \delta z_i. 
  \label{eq:watermarked_latent}
  \vspace{-2pt}
\end{equation}

\begin{figure}[t]
    \centering
    \begin{minipage}[t]{0.38\textwidth}
        \centering
        \includegraphics[width=\linewidth]{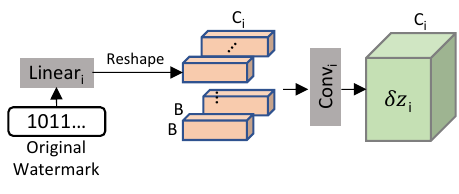}
        \caption{Structure of the watermark embedding module $W_{Emb_{i}}$}
        \label{fig:fig3}
        \vspace{-15pt}
    \end{minipage}
    \hfill
    \begin{minipage}[t]{0.55\textwidth}
        \centering
        \includegraphics[width=\linewidth]{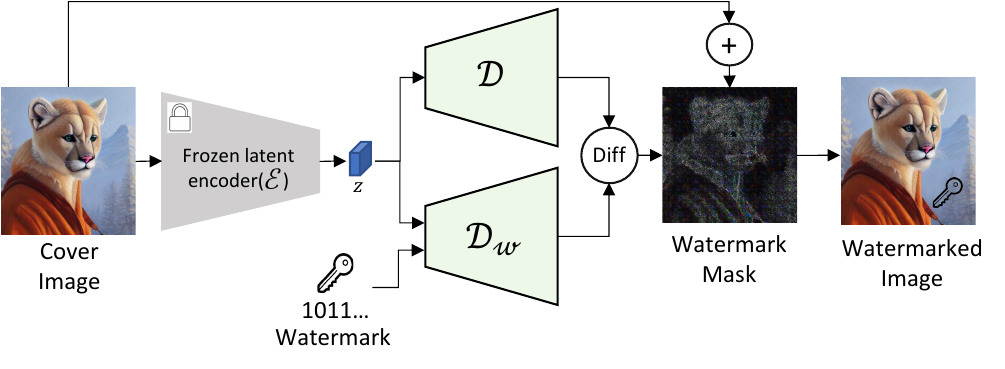}
        \caption{Diagram of general image watermarking process with LaWa}
        \label{fig:fig4}
        \vspace{-15pt}
    \end{minipage}
\end{figure}

As illustrated in \cref{fig:fig3}, $W_{E m b_{i}}$ comprises of a linear layer that maps $m$ to a noise block $b_{i} \in \mathbb{R}^{B \times B \times C_{i}}$. $C_{i}$ is the latent channel size at the $i$th step and $B$ is the height and width size of $b_{i}$ that is kept constant for all upsampling steps. After $b_{i}$ is obtained, we perform spatial watermark coding where $b_{i}$ is repeated along the height and width dimensions to match the size of $z_{i}$ 
As a result of the applied spatial watermark coding, the watermark noise carries repeated watermark message 
with the idea to create sufficient redundancy in the noise to improve robustness to geometrical attacks while minimally damaging the perceptual quality. The watermark noise is then passed through a convolution layer to generate $\delta z_{i}$. To ensure $W_{E m b_{i}}$ does not initially change the latent feature, we follow~\cite{zhang2023adding} and initialize the weights and biases of the convolution layer to 0.

Unlike the original decoder $\mathcal{D}$ that reconstructs the image without watermark $\hat{x}=\mathcal{D}(z)$, the modified latent decoder $\mathcal{D}_{w}$ generates the watermarked image $\hat{x}_{w}=\mathcal{D}_{w}(z)$.

\subsection{Training}
In the training process, the LDM's pre-trained image encoder as well as the original layers of the decoder are kept frozen. This ensures that training the watermark embedding modules do not degrade the image generation quality.

To ensure the perceptual similarity of the watermarked image $\hat{x}_{w}$ and the original generated image $\hat{x}$, we use a combination of pixel-wise distortion and perceptual loss functions. For the distortion loss, MSE ($l_{2}$ distance) loss defined as $\mathcal{L}_{I}=\left\|\hat{x}_{w}-\hat{x}\right\|^{2}$ is used. For the perceptual loss function, we employ LPIPS loss~\cite{zhang2018unreasonable} to minimize the perceptual distortion between $\hat{x}_{w}$ and $\hat{x}$.

To further improve the quality of the watermarked image, an adversarial critic network $\mathcal{C}$ using Wasserstein GAN~\cite{arjovsky2017wasserstein} is also implemented. Thus, the corresponding adversarial training includes two loss functions as follows:
\begin{align} 
\vspace{-2pt}
\mathcal{L}_{\textit{adv}\mathcal{D}_w} & = - \mathbb{E}_{\hat{x}_w \sim P_{\hat{x}_w}}[\mathcal{C}(\hat{x}_w)], \label{eq:adv_loss_1}\\
\mathcal{L}_{\textit{adv}\mathcal{C}} & = \mathbb{E}_{\hat{x}_w \sim P_{\hat{x}_w}}[\mathcal{C}(\hat{x}_w)] - \mathbb{E}_{\hat{x} \sim P_{\hat{x}}} [\mathcal{C}(\hat{x})], \label{eq:adv_loss_2}
\vspace{-2pt}
\end{align}
where $\mathcal{L}_{a d v \mathcal{C}}$ and $\mathcal{L}_{a d v \mathcal{D}_{w}}$ are the adversarial losses for the critic $\mathcal{C}$ and the modified decoder $\mathcal{D}_{w}$, respectively. $P_{\hat{x}_{w}}$ and $P_{\hat{x}}$ are the distribution of the watermarked and original images. The overall reconstruction loss is then defined as:
\vspace{-5pt}
\begin{equation}
  \mathcal{L}_{\textit{rec}} = \lambda_{\textit{I}} \mathcal{L}_{\textit{I}} + \lambda_{\textit{LPIPS}} \mathcal{L}_{\textit{LPIPS}}(\hat{x}_w, \hat{x}) + \lambda_{\textit{adv}} \mathcal{L}_{\textit{adv}\mathcal{D}_w},
  \label{eq:image_rec_loss}
\end{equation}
where $\lambda_{\textit{I}}$, $\lambda_{\textit{LPIPS}}$, and $\lambda_{\textit{adv}}$ are the relative loss weights.

Our watermark extractor, denoted by $W_{E x t}$, should be robust to different image modifications and attacks. As a result, after the watermarked image is generated, a transformation $T$ is randomly sampled from a set of differentiable transformations, $\mathcal{T}$, including common image processing attacks (see supplementary material for more details). For the differentiability issue of JPEG compression, we used the forward simulation attack introduced in~\cite{zhang2021towards}. The selected transformation is then applied to the image and the transformed image is passed through the watermark extraction network, which is based on ResNet50~\cite{he2016deep} architecture with the last linear layer changed to output a $k$-bit message. The output extracted message, $\hat{m}=W_{E x t}(T(\hat{x}))$, should match the original embedded message $m$. We define the extraction loss as the binary cross entropy (BCE) loss between $m$ and $\hat{m}$, $\mathcal{L}_{\text {ext}}=\mathcal{L}_{B C E}(m, \hat{m})$. The final watermark embedding and extraction networks are then optimized to minimize the following total loss:
\begin{equation}
\vspace{-2pt}
  \mathcal{L} = \mathcal{L}_{\textit{rec}} + \lambda \mathcal{L}_{\textit{ext}},
  \label{eq:total_loss}
  \vspace{-2pt}
\end{equation}
where $\lambda$ is the loss weight to control the trade-off between extraction accuracy and image reconstruction quality.

\subsection{Image Watermarking}
\Cref{fig:fig4} shows the general image watermarking process. Corresponding latent feature $z$ of the cover image $x$ is calculated using the frozen encoder, $z=\mathcal{E}(x)$. The feature $z$ is separately processed by $\mathcal{D}$ and $\mathcal{D}_w$ to create a watermark mask that is computed as $\mathcal{D}(z)-\mathcal{D}_{w}(z)$. Using this watermark mask, the cover image is then watermarked as $\hat{x}_{w}=x+\mathcal{D}_{w}(z)-\mathcal{D}(z)$.

\subsection{Watermark Matching}
\label{watermark_matching}
Given an image at the watermark extraction time, our watermark extractor $W_{Ext}$ tries to decode the watermark message $\hat{m}$. To address the detection and attribution problems, $\hat{m}$ needs to be matched with one of existing original messages (i.e., $m$ ) stored in the database. Following~\cite{fernandez2023stable, yu2021artificial} and to increase the matching robustness against image modifications, we use a soft matching method defined as:
\begin{equation}
M(\hat{m}, m), \geq n \quad \quad n \in\{0, \ldots, k\},
\end{equation}
where $M(\hat{m}, m)$ counts the number of matching bits in $\hat{m}$ and $m$, and $n$ is the threshold for flagging a match.

For a successful watermark matching, the rate of false detection and attribution should be at an acceptable level. Potential errors include detecting a vanilla image as watermarked (detection FPR), missing a watermarked image (detection false negative rate), and attributing an image to the wrong user (attribution FPR). Having low error rate is an important practical aspect of this problem.

Assuming that extracted bits from vanilla images are i.i.d Bernoulli random variables with parameter 0.5, the theoretical upper bound of FPR is as follows:
\begin{align}
FPR_{det}(n) & = Pr\left(M(\tilde{m}, m)>n \mid H_{0}\right)=\sum_{i=n+1}^{k}\binom{k}{i} 0.5^{k}, \label{eq:fpr_det}\\
FPR_{a t t}(N, n) & = 1-\left(1 - FPR_{det}(n)\right)^{N}, \label{eq:fpr_att}
\end{align}
where $FPR_{det}$ is the detection FPR, $FPR_{att}$ is the attribution FPR, $N$ is the total number of users, $\tilde{m}$ is any random message, and $H_{0}$ is the null hypothesis: "any random message, $\tilde{m}$, match with $m$ in more than $n$ bits". Clearly, the more the number of users, the higher the attribution FPR. By setting the $FPR_{att}$ and total number of users to a desired level, we can obtain the minimum required $n$ for watermark matching by using \cref{eq:fpr_det,eq:fpr_att} reversely. More information is provided in the supplementary material.

\section{Experiments}
In this section, the performance of the proposed method is quantitatively and qualitatively evaluated and compared with the previous works. We also study the generality of LaWa for different image generation tasks. A false positive analysis along with an extensive test for the trade-off between capacity, quality, and robustness is also presented. At the end, we will provide an ablation study to verify the effect of each loss term in our objective function.

\subsection{Experimental Settings}
To train LaWa’s watermark embedding and extraction modules, we utilize 100K images from MIRFlickR dataset~\cite{huiskes2008MIR}, where the images are randomly cropped to $256 \times 256$ resolution. The training is performed with an AdamW optimizer with the learning rate of $6 \mathrm{e}-5$, 40 epochs, and the batch size of 8. For all our experiments, we set the weights in \cref{eq:image_rec_loss,eq:total_loss} to $\lambda_{I}=0.1$, $\lambda_{LPIPS}=1.0$, $\lambda_{adv}=1.0$, and $\lambda=2.0$, which are obtained experimentally. For the main experiments, we use the KL-f8 auto-encoder of Stable Diffusion~\cite{rombach2022high} and 48-bit messages to train the LaWa method unless mentioned otherwise.

To numerically evaluate the distortion of the watermarked and original generated images, we use Peak Signal-to-Noise Ratio (PSNR) and Structural Similarity Index Measure (SSIM)~\cite{wang2004image}. Moreover, we evaluate the perceptual quality of images using LPIPS similarity score~\cite{zhang2018unreasonable} and Single Image Frechet Inception Distance (SIFID)~\cite{shaham2019singan}.

\subsection{Comparison Results}
We compare the performance of LaWa with previous in-generation and image watermarking methods using AI-generated and natural image datasets. For AI-generated dataset, we consider using LaWa as both in-generation and post-generation watermarking methods. For the baseline of in-generation methods, we used Stable Signature (SS)~\cite{fernandez2023stable}, which is a model watermarking method for in-generation image watermarking by creating different copies of the decoder. Alternatively, image watermarking methods can be used to watermark the generated images in a post-generation manner. We provide the comparison results with different post-generation techniques including encoder-decoder based (HiDDeN~\cite{zhu2018hidden}, RivaGAN~\cite{zhang2019robust}, and RoSteALS~\cite{bui2023rosteals}), optimization-based (SSL~\cite{fernandez2022watermarking} and FNNS~\cite{kishore2021fixed}), and frequency-based (DCT-DWT~\cite{cox2007digital}) methods. For HiDDeN and RivaGAN, 30-bit and 32-bit pre-trained weights in~\cite{hiddenpretrained} and~\cite{rivaGANpretrained} are respectively used. For RoSteALS, we trained their model for 48- and 32-bit message lengths with the same dataset and attacks as in LaWa. For SSL and FNNS, we used their publicly available codes and their pre-trained modules to obtain the results.

\def\rot{\rotatebox}
\newcolumntype{M}[1]{>{\centering\arraybackslash}m{#1}}
\begin{table*}[t]
\centering
\caption{
Comparison results of LaWa with existing post- and in-generation image watermarking methods in terms of quality, embedding time, and robustness against various attacks. \textbf{*}: in-generation methods.}
\vspace{-3pt}
\label{tab:tab1}
\resizebox{0.99\textwidth}{!}{
\begin{tabular}{l|cc|M{21pt}|M{23pt}M{27pt}M{27pt}M{27pt}M{27pt}M{27pt}M{27pt}M{27pt}M{27pt}M{27pt}|M{16pt}}
\toprule  
\multirow{2}{*}[-15pt]{\thead{Method (bit\#)}} & \multicolumn{2}{c|}{Image quality} & &\multicolumn{10}{c}{Bit accuracy $\uparrow$} & \\
& PSNR/SSIM $\uparrow$ & LPIPS/SIFID $\downarrow$ & Emb. time (ms) & \rot{45}{None} & \rot{45}{C. Crop 0.1} & \rot{45}{R. Crop 0.1} & \rot{45}{Resize 0.7} & \rot{45}{Rot. 15} & \rot{45}{Blur} & \rot{45}{Contr. 2.0} & \rot{45}{Bright. 2.0} & \rot{45}{JPEG 70} & \rot{45}{Comb.} & \rot{45}{Ave.} \\
\midrule
\multicolumn{15}{c}{\textbf{AI-generated Images}} \\
\midrule
DCT-DWT (32)~\cite{cox2007digital} & 39.47/0.97 & 0.03/0.02 & 139 &0.91 & 0.51 & 0.51 & 0.52 & 0.52 & 0.51 & 0.52 & 0.51 & 0.51 & 0.51 & 0.55\\
Hidden (30)~\cite{hiddenpretrained} & 32.59/0.95 & 0.03/0.05 & 17 & 0.91 & 0.91 & 0.91 & 0.82 & 0.79 & 0.76 & 0.75 & 0.74 & 0.53 & 0.59 & 0.77\\
SSL (32)~\cite{fernandez2022watermarking} & 33.23/0.89 & 0.15/0.28 & 870 & \textbf{1.00} & 0.74 & 0.72 & 0.99 & 0.99 & \textbf{1.00} & 0.96 & 0.95 & 0.99 & 0.85 & 0.92 \\
RoSteALS (32)~\cite{bui2023rosteals} & 29.31/0.93 & 0.04/0.06 & 144 & \textbf{1.00} & 0.50 & 0.51 & \textbf{1.00} & 0.47 & \textbf{1.00} & 0.88 & 0.86 & \textbf{1.00} &  0.50 & 0.77 \\
RivaGan (32)~\cite{zhang2019robust} & \textbf{40.53/0.98} & 0.03/0.04 & 57 & 0.99 & 0.98 & \textbf{0.98} & 0.87 & 0.91 & 0.99 & 0.81 & 0.80 & 0.98 & 0.93 & 0.92 \\
\textbf{LaWa}$^*$ (32) & 34.25/0.89 & \textbf{0.03/0.01} & 1 & \textbf{1.00} & \textbf{1.00} & \textbf{0.98} & \textbf{1.00} & \textbf{1.00} & \textbf{1.00} & \textbf{1.00} & \textbf{1.00} & \textbf{1.00} & 0.97 & \textbf{0.99}\\
\textbf{LaWa-post-gen} (32) & 34.28/0.90 & 0.03/0.02 & 33 & \textbf{1.00} & \textbf{1.00} & 0.96 & \textbf{1.00} & \textbf{1.00} & \textbf{1.00} & \textbf{1.00} & \textbf{1.00} & \textbf{1.00} & \textbf{0.98} & \textbf{0.99} \\

\midrule 
SSL (48)~\cite{fernandez2022watermarking} & 33.25/0.89 & 0.15/0.27 & 870 & \textbf{1.00} & 0.72 & 0.70 & 0.99 & \textbf{0.99} & \textbf{1.00} & 0.94 & 0.94 & 0.99 & 0.82 & 0.91 \\
FNNS (48)~\cite{kishore2021fixed} & \textbf{36.84/0.97} & 0.04/0.05 & 645 & \textbf{1.00} & \textbf{0.98} & \textbf{0.96} & 0.95 & 0.75 & 0.54 & 0.86 & 0.85 & 0.94 & 0.91 & 0.87\\
RoSteALS (48)~\cite{bui2023rosteals} & 29.30/0.92 & 0.05/0.07 & 144 &\textbf{1.00} & 0.51 & 0.50 & \textbf{1.00} & 0.50 & \textbf{1.00} & 0.88 & 0.85 & \textbf{1.00} & 0.49 & 0.78\\
SS$^*$ (48)~\cite{fernandez2023stable} & 32.02/0.84 & 0.05/0.09 & 0 & 0.99 & 0.95 & 0.93 & 0.96 & 0.81 & 0.78 & 0.97 & 0.96 & 0.92 & 0.92 & 0.92\\
\textbf{LaWa}$^*$ (48) & 33.52/0.86 & \textbf{0.04/0.02} & 1 & \textbf{1.00} & 0.95 & 0.91 & 0.99 & 0.96 & 0.99 & \textbf{1.00} & \textbf{1.00} & \textbf{1.00} & 0.94 & \textbf{0.97} \\
\textbf{LaWa-post-gen} (48) & \textbf{33.45/0.87} & \textbf{0.04/0.02} & 34 & \textbf{1.00} & 0.94 & 0.90 & 0.96 & 0.97 & 0.99 & \textbf{1.00} & \textbf{1.00} & \textbf{1.00} & \textbf{0.95} & \textbf{0.97} \\

\midrule
\multicolumn{15}{c}{\textbf{CLIC}} \\
\midrule

DCT-DWT (32)~\cite{cox2007digital} & 38.94/0.97 & \textbf{0.02/0.02} & 230 & 0.87 & 0.51 & 0.51 & 0.51 & 0.51 & 0.52 & 0.50 & 0.50 & 0.52 & 0.50 & 0.55 \\
Hidden (30)~\cite{hiddenpretrained} & 33.33/0.96 & 0.04/0.04 & 28 & 0.89 & 0.89 & 0.88 & 0.86 & 0.76 & 0.64 & 0.76 & 0.74 & 0.54 & 0.58 & 0.75 \\
SSL (32)~\cite{fernandez2022watermarking} & 35.03/0.92 & 0.16/0.19 & 1261 & \textbf{1.00} & 0.86 & 0.80 & \textbf{1.00} & \textbf{1.00} & \textbf{1.00} & 0.96 & 0.93 & 0.98 & 0.90 & 0.94 \\
RoSteALS (32)~\cite{bui2023rosteals} & 28.22/0.91 & 0.08/0.04 & 240 & \textbf{1.00} & 0.50 & 0.50 & \textbf{1.00} & 0.51 & \textbf{1.00} & 0.90 & 0.87 & 1.00 & 0.50 & 0.78 \\
RivaGan (32)~\cite{zhang2019robust} & \textbf{42.04/0.98} & 0.04/0.06 & 84 & \textbf{1.00} & 0.98 & 0.96 & 0.99 & 0.95 & 0.99 & 0.89 & 0.87 & 0.99 & 0.96 & 0.96 \\
\textbf{LaWa} (32) & 36.19/0.94 & 0.05/0.03 & 32 & \textbf{1.00} & \textbf{1.00} & \textbf{0.97} & 0.98 & \textbf{1.00} & \textbf{1.00} & \textbf{1.00} & \textbf{0.99} & \textbf{1.00} & \textbf{0.99} & \textbf{0.99} \\

\midrule 
SSL (48)~\cite{fernandez2022watermarking} & 35.03/0.92 & 0.16/0.19  & 1261 & \textbf{1.00} & 0.83 & 0.78 & \textbf{1.00} & \textbf{1.00} & \textbf{1.00} & 0.96 & 0.92 & 0.98 & 0.88 & 0.93 \\
FNNS (48)~\cite{kishore2021fixed} & 35.54/\textbf{0.96} & 0.07/0.20 & 1116 & \textbf{1.00} & 0.99 & 0.95 & 0.94 & 0.75 & 0.51 & 0.85 & 0.82 & 0.91 & 0.89 & 0.86 \\
RoSteALS (48)~\cite{bui2023rosteals} & 27.83/0.91 & 0.08/\textbf{0.04} & 240 & \textbf{1.00} & 0.50 & 0.50 & \textbf{1.00} & 0.48 & \textbf{1.00} & 0.82 & 0.80 & \textbf{1.00} & 0.50 & 0.76\\
\textbf{LaWa} (48) & \textbf{35.58}/0.92 & \textbf{0.05}/0.05 &  34 & \textbf{1.00} & \textbf{1.00} & \textbf{0.97} & 0.98 & \textbf{0.98} & 0.98 & \textbf{1.00} & \textbf{0.99} & 0.98 & \textbf{0.90} & \textbf{0.98} \\
\bottomrule
\end{tabular}}
\vspace{-15pt}
\end{table*}

The upper portion of \cref{tab:tab1} summarizes LaWa's results with 32- and 48-bit watermark messages compared to the baselines in terms of the watermarked image quality as well as the bit extraction accuracy after different attacks. The results are obtained by watermarking 1K generated images (with 1K text prompts obtained from MS-COCO~\cite{lin2014microsoft} and MagicPrompt-SD~\cite{GustavostaPrompts} datasets). Resolution of generated images is set to $512 \times 512$. Each generated image is watermarked with 10 different messages and the average bit accuracy is reported. Combined attack includes a series of $40\%$ area center cropping, brightness 2.0, and JPEG 80. LaWa$^*$ and LaWa-post-gen refer to the cases where the proposed method is used as in-generation and post-generation watermarking, respectively.

As shown in \cref{tab:tab1}, LaWa achieves the best perceptual quality for AI-generated images with 32- and 48-bit versions in LPIPS and SIFID scores. For PSNR and SSIM, LaWa has comparable performance with other methods, while RivaGAN and FNNS outperform the others for 32- and 48-bit watermarks, respectively. For robustness to attacks, LaWa has the best bit accuracy for all attacks for 32-bit watermarks. For 48-bit watermarks, except for $10\%$ random crop attack where FNNS works better ($90\%$ bit accuracy of LaWa compared to $96\%$ of FNNS), LaWa has better or comparable bit accuracy. Overall, LaWa outperforms other methods in more number of attacks, specifically for the combined attack. Moreover, LaWa outperforms SS (as an in-generation method) in both image quality and robustness to a variety of attacks. More detailed set of results with different attack parameters is given in the supplementary material. When LaWa is used as a post-generation watermarking method (i.e, decoupled from the generation), the extraction accuracy is nearly unchanged compared to in-generation watermarking. 

\begin{figure}[t]
    \centering
    \includegraphics[width=0.98\linewidth]{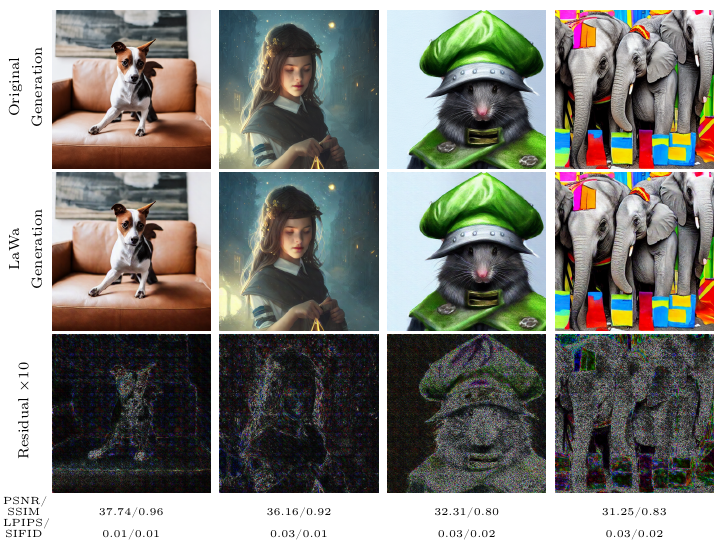}
    \vspace{-3pt}
    \caption{Qualitative examples of generated images using LaWa, original generated images, and residual images (multiplied by 10). Extensive added noise by LaWa in high frequency textures reduces the PSNR/SSIM without damaging the perceptual quality (LPIPS/SIFID).}
    \label{fig:fig5}
    \vspace{-15pt}
\end{figure}

The computational complexity of all methods compared to ours in terms of averaged embedding time is also given in \cref{tab:tab1}. Overall, it is seen that post-generation methods result in a significant overhead for watermarking the generated images, especially the iteration-based ones such as SSL and FNNS. On the other hand, the in-generation methods such as SS and ours provide near zero overhead as no post-processing is needed to add the watermark.

We also compare the performance of LaWa in general image watermarking against baseline methods using a natural (i.e, not AI-generated) dataset. Following earlier works in~\cite{bui2023rosteals, fernandez2022watermarking}, CLIC dataset~\cite{toderici2020workshop} is used, which has 428 test images with different resolutions. For all images, while maintaining the aspect ratio, larger dimension of each image is resized to 1024 to avoid memory limitation issue for some of the baselines. The results with CLIC dataset are given in the bottom part of \cref{tab:tab1}. LaWa-post-gen achieves the highest bit accuracy for almost all the attacks and comparable or better visual quality compared to the baselines. Qualitative results for CLIC are in the supplementary material.

\subsection{Qualitative Analysis}
\cref{fig:fig5} shows visual examples of 4 images generated by the original and modified decoders (i.e., original generated vs. watermarked images). The residual images (multiplied by 10) and the numerical quality metrics including PSNR, SSIM, LPIPS, and SIFID are also given for each pair.

Comparing the watermarked and original images, although the calculated PSNR is not very high, it is visually not possible to differentiate them as no visual artifact is evident. This is expected as we trained LaWa focusing on the perceptual quality of the watermarked images while giving lower weight to the MSE loss, which directly affects the PSNR. From the examples in \cref{fig:fig5}, it is clear that our method has learned to add more noise to the high frequency textures, especially for the generated images with more texture (e.g., the elephants). In practice, when using in-generation watermarking, there is no original image available for comparison as all generated images conceal the watermark. This relaxes the requirement for watermarked images to have a high pixel-wise similarity to the original image. Therefore, we argue that if generated images have no visual artifact, watermark addition is successful even if PSNR is not very high. More qualitative results compared to the previous works are presented in the supplementary material.

\subsection{Capacity, Quality, and Robustness Trade-off}
\def\rot{\rotatebox}
\newcolumntype{M}[1]{>{\centering\arraybackslash}m{#1}}
\setlength\intextsep{-9pt}
\begin{wraptable}[9]{R}{0.65\textwidth}
    \caption{Effect of the number of bits on generation quality and bit extraction accuracy.}
    \resizebox{0.64\textwidth}{!}{
    \begin{tabular}{c|M{50pt}M{50pt}M{25pt}M{25pt}M{25pt}M{25pt}M{25pt}M{25pt}}
        \hline
        \thead{Bit\\length} & \thead{PSNR/\\SSIM} & \thead{LPIPS/\\SIFID} & \rot{45}{None} &\rot{45}{C. Cro 0.1} & \rot{45}{Contr. 2.0} & \rot{45}{Rot. 15} & \rot{45}{JPEG 70} & \rot{45}{Comb.}  \\
        \hline
        32 & 34.25/0.89 & 0.03/0.01 & 1.00 & 1.00 & 1.00 & 1.00 & 1.00 & 0.97\\
        48 & 33.52/0.86 & 0.04/0.02 & 1.00 & 0.95 & 1.00 & 0.96 & 1.00 & 0.94\\
        64 & 33.11/0.83 & 0.04/0.02 & 1.00 & 0.93 & 0.99 & 0.92 & 0.96 & 0.95\\
        96 & 32.82/0.83 & 0.04/0.02 & 0.99 & 0.87 & 0.98 & 0.77 & 0.92 & 0.95\\
        128 & 31.87/0.82 & 0.08/0.09 & 0.95 & 0.93 & 0.95 & 0.84 & 0.92 & 0.87 \\
        \hline
    \end{tabular}}
    \label{tab:tab2}
\end{wraptable}


\textbf{Effect of Number of Bits.} We study the effect of the number of bits on generation quality and robustness in \cref{tab:tab2}. LaWa can hide 64- and 96-bit watermark messages with minor drop in image quality, while achieving near $100\%$ bit accuracy with no attack. With 128-bit messages, minor drop in the quality and bit accuracy against attack is observed compared to embedding shorter messages. Using higher number of bits can further improve the FPR of LaWa for its application in image generation services, which are designed to support a large number of users.

\textbf{Quality-Robustness Trade-off.} We can trade the generation quality for higher bit accuracy by changing $\lambda$ in \cref{eq:total_loss}. \Cref{tab:tab3} presents the performance of LaWa when trained with different $\lambda$s. The corresponding PSNR and bit accuracy (after applying combined attack) on 1K images is reported. As observed, by increasing $\lambda$, the extraction robustness is improved, but the quality of the watermarked images drops.

\subsection{LaWa's Performance for Different Tasks}
In this sub-section, the generality of LaWa for different image generation tasks~\cite{rombach2022high} including text-to-image, inpainting, super-resolution, and image editing is studied. The corresponding performance results are given in \cref{tab:tab5}. The results are obtained for 1K images generated/watermarked with a separate random watermark message of 48 bits (see the supplementary material for more details). As summarized in the \cref{tab:tab5}, the achieved bit accuracy averaged over all the tasks is higher than $97\%$, while the perceptual quality is preserved. This shows the applicability and generality of our proposed method to different generation tasks.
\begin{table}[t]
    \begin{minipage}[t]{.46\linewidth}
        \centering
        \caption{Trade-off between image quality and bit accuracy. $\lambda$: trade-off weight in \cref{eq:total_loss}.}
        \resizebox{1\textwidth}{!}{
        \begin{tabular}{c|ccccccc}
            \hline
            $\lambda$ & 1.0 & 1.7  & 2.0 & 4.0 & 7.0 & 10.0\\
            \hline
            PSNR $\uparrow$ & 34.76 & 33.84  & 33.52 & 33.06 & 30.56 & 28.47 \\
            Bit acc. on comb. $\uparrow$ & 0.65 & 0.75 & 0.94 & 0.96 & 0.97 & 0.99\\
            \hline
        \end{tabular}}
        \label{tab:tab3}
        \vspace{-5pt}
    \end{minipage}\hfill
    \begin{minipage}[t]{.51\linewidth}
        \centering
        \caption{Ablation study on the effect of image quality loss weights.}
        \resizebox{1.0\textwidth}{!}{
        \begin{tabular}{c|ccc}
            \hline
            $\lambda_I$ & 0.1 & 0.5  & 1.0 \\
            \hline
            PSNR/SSIM $\uparrow$ & 33.52/0.86 & 33.60/0.88 & \textbf{33.94/0.90} \\
            LPIPS/SIFID $\downarrow$ & \textbf{0.04/0.02} & 0.05/0.04 & 0.06/0.05 \\
            Bit acc. on None/comb. $\uparrow$ & \textbf{1.00/0.94} &1.00/0.88 & 0.99/0.84 \\
            \hline
        \end{tabular}}
        \label{tab:tab4}
        \vspace{-5pt}
    \end{minipage}
\end{table}


\subsection{False Positive Analysis}
The false positive analysis of LaWa's performance for both detection and attribution problems is given in this sub-section. As discussed in \cref{watermark_matching}, in the detection problem, we only aim to detect if an image is watermarked while having a low FPR (detecting vanilla images as watermarked). On the other hand, in the attribution problem, we aim to find the user who created the image without making a false attribution to the wrong user (false attribution).
\begin{figure*}[b]
\vspace{-12pt}
\centering
\includegraphics[width=0.98\linewidth]{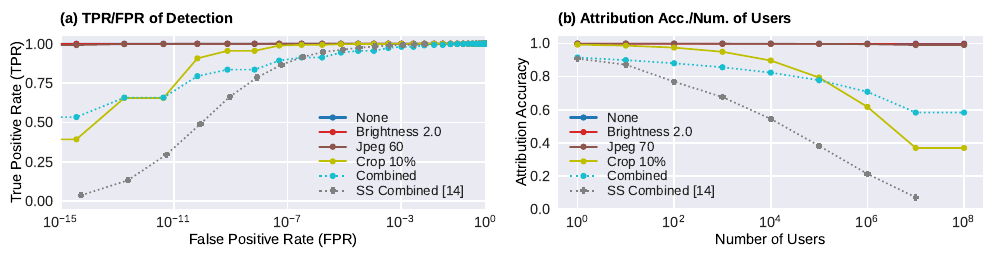}
\vspace{-5pt}
\caption{Watermark matching performance in detection and attribution. (a) Detection TPR/FPR for different number of matching bits under different attacks. (b) Accuracy of attributing the watermarked image to the right user under different attacks and number of users assuming the detection FPR of $10^{-6}$.}
\label{fig:fig6}
\end{figure*}

\textbf{Detection.} For detection, we use 1K watermarked images with a single 48-bit message to study the effect of the threshold $n$ used for soft watermark matching on $FPR_{det}$. For each $n \in \{0,...,48\}$, we use \cref{eq:fpr_det} to calculate $FPR_{det}$, while $TPR_{det}$ is experimentally obtained for the 1K images under various attacks. {We experimentally validate the theoretical $FPR_{det}$ in the supplementary material}. 

\Cref{fig:fig6}a shows LaWa performance at different $FPR_{det}$. Under no attack, brightness 2.0, and JPEG 60, LaWa successfully detects 99\% of the images while only flagging 1 out of $10^{15}$ vanilla images as watermarked. For more severe attacks such as crop 10\% and combined attacks, there is a clear trade-off between detection accuracy and $FPR_{det}$. LaWa still detects more than 95\% of the cropped and 83\% of the combined-attacked images while only flagging 1 out of $10^9$ vanilla images. We also compare LaWa with Stable Signature (SS)~\cite{fernandez2023stable} under combined attack and show that LaWa has a much higher TPR.
\newcolumntype{M}[1]{>{\centering\arraybackslash}m{#1}}
\begin{table}[t]
    \begin{minipage}[t]{.52\linewidth}
        \centering
        \caption{Performance of LaWa in different image generation tasks in terms of quality and robustness.}
        \resizebox{1.0\textwidth}{!}{
        \begin{tabular}{M{70pt}|M{50pt}M{50pt}M{25pt}M{25pt}M{25pt}M{25pt}M{25pt}}
            \hline
            Task & \thead{PSNR/\\SSIM} & \thead{LPIPS/\\SIFID} &\rot{45}{C. Crop 0.1} & \rot{45}{Contr. 2.0} & \rot{45}{Rot. 15} & \rot{45}{JPEG 70} \\
            \hline
            Text-to-Image & 33.52/0.86 & 0.04/0.02  & 0.95 & 1.00 & 0.96 & 1.00 \\
            Inpainting & 33.68/0.86 & 0.04/0.02 & 0.94 & 1.00 & 0.97 & 0.99 \\
            Super-Resolution & 35.40/0.89 & 0.04/0.01 & 0.96 & 0.99 & 0.96 & 0.98 \\
            Image Editing & 34.22/0.88 & 0.04/0.01  & 0.95 & 1.00 & 0.96 & 0.99 \\
            \hline
        \end{tabular}}
        \label{tab:tab5}
        \vspace{-15pt}
    \end{minipage}\hfill
    \begin{minipage}[t]{.46\linewidth}
        \centering
        \caption{Ablation study on the effect of each loss term on image quality and bit extraction robustness.}
        \resizebox{1.0\textwidth}{!}{
        \begin{tabular}{M{27pt}M{27pt}M{27pt}|M{60pt}M{60pt}M{50pt}}
            \hline
            MSE & LPIPS & Critic  & PSNR/SSIM & LPIPS/SIFID  & Bit Acc. Comb. \\
            \hline
            \checkmark & & & 31.26/0.82 & 0.22/0.52 & 0.96\\
             &  \checkmark  &  & 28.83/0.36 & 0.05/0.41 & 0.93\\
               \checkmark &  \checkmark &  &  29.47/0.47 &  0.04/0.31 &   0.85\\
             \checkmark &  &  \checkmark &  32.29/0.80 &   0.10/0.05 &   \textbf{0.99} \\
             &  \checkmark &  \checkmark  &  33.05/0.85 &  0.05/0.02 &  0.73\\
              \checkmark &  \checkmark &  \checkmark &  \textbf{33.52/0.86} &  \textbf{0.04/0.02}  &  0.94\\
            \hline
        \end{tabular}}
        \label{tab:tab6}
        \vspace{-15pt}
    \end{minipage}
\end{table}

\textbf{Attribution.} For the attribution problem, we can use \cref{eq:fpr_det,eq:fpr_att} reversely to find the number of possible users for any $FPR_{att}$. For example, for $FPR_{att}=10^{-6}$ in \cref{eq:fpr_att}, we can calculate the required $FPR_{det}$ for any number of users $N$. We further use the computed $FPR_{det}$ and \cref{eq:fpr_det} to find the necessary watermark matching threshold $n$. For example, for $FPR_{att}=10^{-6}$ and $N=10^{5}$, we should use $n=45$ as the threshold of soft watermark matching for 48-bit watermark messages. 

We study LaWa's attribution accuracy for each $N$ with 50K different 48-bit messages. We generate 2 images per message resulting in a total of 100K images. For $FPR_{att}=10^{-6}$ and each $N$, the required threshold $n$ is computed. Using this $n$, we perform watermark matching between the extracted message from each image and all {the other} messages. The match with the highest score above $n$ is considered true positive (TP) if it corresponds to the correct user. Accuracy is calculated as $TP/100k$. For $N$s larger than 50k, we assume some users have not generated any image. 

\Cref{fig:fig6}b shows the attribution accuracy of our method for different attacks. Under no attack, brightness 2.0, and JPEG 70, LaWa attributes 99\% of the generated images to the correct user among $10^6$ total users. Accuracy decreases to less than 70\% and 60\% for more severe combined and 10\% crop attacks, respectively. Compared to Stable Signature (SS)~\cite{fernandez2023stable}, LaWa achieves a better performance with 20\% accuracy for the combined attack. Moreover, LaWa has zero false attribution when $N=10^6$, which means no image was attributed to the wrong user. This is expected as the watermark matching threshold is really high at $n=47$. Overall, as we increase the number of users, the probability of false accusations also increases.

\subsection{Learning-based Attacks}
\begin{wrapfigure}[13]{R}{0.5\textwidth}
    \centering
    \includegraphics[width=1.0\linewidth]{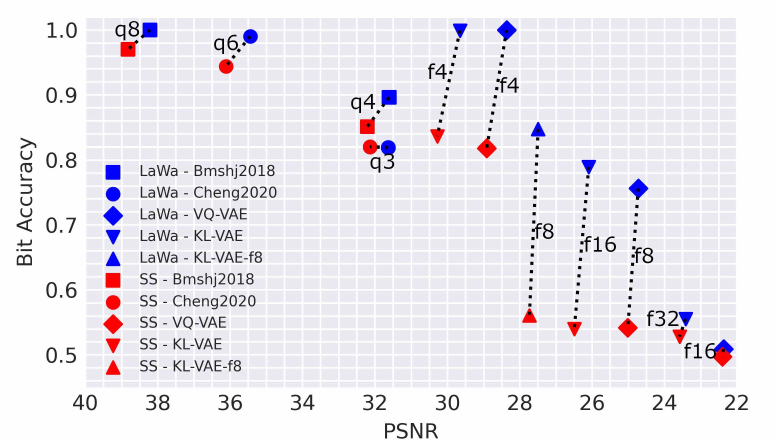}
    \caption{Effect of the learning-based removal attacks. Results are reported for (SS)~\cite{fernandez2023stable} and LaWa. $q,f$: quality/downsampling factors for each attack.}
    \label{fig:fig7}
\end{wrapfigure}
We study the impact of learning-based watermark removal attacks, where the watermarked image is passed through an auto-encoder model ~\cite{akbari2021learned,balle2018variational,cheng2020learned,esser2021taming,rombach2022high} as an attempt to remove the watermark. \Cref{fig:fig7} shows the robustness of LaWa compared to Stable Signature (SS)~\cite{fernandez2023stable} against several auto-encoder models including Bmshj2018~\cite{balle2018variational}, Cheng2020~\cite{cheng2020learned}, KL-VAE~\cite{rombach2022high}, and VQ-VAE~\cite{esser2021taming}. 

Each type of attack is applied using different quality/downsampling factors used in the auto-encoders. We observe that LaWa is quite robust against the attacks performed by Bmshj2018 with q8, Cheng2020 with q6, KL-VAE with f4, and VQ-VAE with f4. As the attack gets stronger (i.e., images are more compressed with larger quality or downsampling factors), both the bit accuracy and the PSNR reduce. Meaning if the attack is strong enough to remove the watermark, the quality of the watermarked image will inevitably experience a significant degradation (i.e., PSNR$<$24). It is also observed that even when the auto-encoder used in the attack is the same as the one we used to generate/watermark the images (i.e., KL-VAE f8), the attack will not erase the watermark and the bit accuracy remains as high as 85\%. In contrast, the bit accuracy of SS~\cite{fernandez2023stable} drops to 55\%. It should be noted that unlike ~\cite{fernandez2023stable}, we calculate the PSNR between the watermarked and the attacked images which is a more accurate representation of the attack's effect.

\subsection{Ablation Study}
\textbf{Loss Terms.} \Cref{tab:tab6} shows the effect of each loss used in \cref{eq:image_rec_loss}. We observe that the Critic network helps to improve the visual quality of the images. Using MSE and Critic losses achieves the best bit accuracy, but this setting leads to low perceptual quality and high visual artifacts in the image. By adding LPIPS loss, such visual artifacts are removed from the generated images. 

\textbf{Loss Weights.} \Cref{tab:tab4} shows the effect of changing loss weights in \cref{eq:image_rec_loss}. We did this study by fixing both $\lambda_{LPIPS}$ and $\lambda_{adv}$ to 1.0 and changing $\lambda_{I}$. We observe that $\lambda_{I}$ > 0.1 provides minor improvement in PSNR/SSIM, but damages LPIPS/SIFID and LaWa’s robustness.
{More ablation studies over other components of LaWa along with qualitative results are given in the supplementary material}.

\section{Conclusion}
In this paper, we proposed an in-generation image watermarking solution, which is applicable to LDMs and any image generation task. We showed that by injecting the watermark embedding modules into the intermediate layers of the frozen decoder, these modules can learn the decoder-specific mapping of latent space to RGB images. Thus, the watermark message is concealed in the generated images in a way that is imperceptible and robust to image modifications. Our solution is scalable to support a large number of users only with one single decoder model. With the fast evolution of generative AI models and its corresponding risks, model developers are expected to actively help with the safe adaptation of these models. Our work is a step towards responsible AI and we hope it can help model developers to implement their generative AI service while they have countermeasures in place to limit any improper use.

%
%
\bibliographystyle{splncs04}
\bibliography{references}

\newpage
\appendix

\section{Additional Experiments}
\subsection{Robustness Under Different Attacks}
In addition to the results in the main body of the paper, \cref{tab:supp_tab1} provides a comprehensive list of different attacks and the robustness of LaWa and other post- and in-generation methods on the AI-Generated dataset. LaWa outperforms other methods in the majority of attacks and has comparable robustness for the other attacks. Furthermore, \cref{tab:supp_tab2} presents more detailed performance results of LaWa (48-bit version) in different generative tasks under different attack parameters. It is evident that LaWa has great robustness to common image transformation attacks for all generative tasks.

\def\rot{\rotatebox}
\newcolumntype{M}[1]{>{\centering\arraybackslash}m{#1}}
\begin{table*}[t]
\centering
\caption{
Comparison results of LaWa with existing post- and in-generation image
watermarking methods in robustness against an extended set of attacks. \textbf{*}: in-generation methods.
}
\vspace{-3pt}
\label{tab:supp_tab1}
\resizebox{0.98\textwidth}{!}{
\begin{tabular}{l|M{27pt}M{27pt}M{27pt}M{27pt}M{27pt}M{27pt}M{27pt}M{27pt}M{27pt}M{27pt}M{27pt}M{27pt}M{27pt}M{27pt}M{27pt}M{27pt}M{27pt}}
\toprule  
\multirow{2}{*}[-15pt]{\thead{Method (bit$\#$)}} &\multicolumn{17}{c}{Bit accuracy $\uparrow$} \\
& \rot{90}{None} & \rot{90}{Rot. $5$} & \rot{90}{Rot. $15$} & \rot{90}{Rot. $25$} & \rot{90}{C. Crop 0.1} & \rot{90}{C. Crop 0.3} & \rot{90}{C. Crop 0.5} & \rot{90}{C. Crop 0.7} & \rot{90}{C. Crop 0.9} & \rot{90}{R. Crop 0.1} & \rot{90}{R. Crop 0.3} & \rot{90}{R. Crop 0.5} & \rot{90}{R. Crop 0.7} & \rot{90}{R. Crop 0.9} & \rot{90}{Resize 0.6} & \rot{90}{Resize 0.7} & \rot{90}{Resize 0.8} \\
\midrule
\midrule
DCT-DWT (32)~\cite{cox2007digital} & 0.52 & 0.51 & 0.51 & 0.51 & 0.50 & 0.51 & 0.51 & 0.51 & 0.51 & 0.51 & 0.51 & 0.51 & 0.51 & 0.51 & 0.50 & 0.51 & 0.51\\
Hidden (30)~\cite{hiddenpretrained} & 0.91 & 0.89 & 0.79 & 0.62 & 0.91 & 0.91 & 0.91 & 0.91 & 0.91 & 0.91 & 0.91 & 0.91 & 0.91 & 0.91 & 0.87 & 0.89 & 0.89\\
SSL (32)~\cite{fernandez2022watermarking} & \textbf{1.00} & \textbf{1.00} & \textbf{0.99} & \textbf{0.99} & 0.74 & 0.92 & 0.97 & 0.99 & \textbf{1.00} & 0.72 & 0.92 & 0.98 & 0.99 & \textbf{1.00} & 0.98 & 0.99 & \textbf{1.00}\\
RoSteALS (32)~\cite{bui2023rosteals} & 0.98 & 0.82 & 0.47 & 0.49 & 0.52 & 0.50 & 0.52 & 0.68 & 0.96 & 0.52 & 0.52 & 0.51 & 0.64 & 0.90 & 0.98 & 0.98 & 0.98 \\
RivaGan (32)~\cite{zhang2019robust} & 0.99 & 0.98 & 0.91 & 0.72 & 0.93 & 0.98 & 0.99 & 0.99 & 0.99 & 0.93 & 0.98 & \textbf{0.99} & \textbf{0.99} & 0.99 & 0.97 & 0.98 & 0.99 \\
\textbf{LaWa}$^*$ (32) & \textbf{1.00} & \textbf{1.00} & \textbf{1.00} & \textbf{0.99} & \textbf{1.00} & \textbf{1.00} & \textbf{1.00} & \textbf{1.00} & \textbf{1.00} & \textbf{0.98} & \textbf{0.99} & \textbf{0.99} & \textbf{0.99} & 0.99 & \textbf{0.99} & \textbf{1.00} & \textbf{1.00}\\
\textbf{LaWa-post-gen} (32) & \textbf{1.00} & \textbf{1.00} & \textbf{1.00} & 0.98 & \textbf{1.00} & \textbf{1.00} & \textbf{1.00} & \textbf{1.00} & \textbf{1.00} & 0.96 & 0.98 & \textbf{0.99} & \textbf{0.99} & 0.99 & 0.97 & \textbf{1.00} & \textbf{1.00}\\

\midrule 
SSL (48)~\cite{fernandez2022watermarking} & \textbf{1.00} & \textbf{0.99} & \textbf{0.99} & \textbf{0.98} & 0.72 & 0.88 & 0.96 & 0.97 & 0.99 & 0.70 & 0.89 & 0.96 & \textbf{0.99} & \textbf{1.00} & 0.96 & 0.99 & 0.99 \\
FNNS (48)~\cite{kishore2021fixed} & \textbf{1.00} & \textbf{0.99} & 0.76 & 0.58 & \textbf{0.98} & 0.99 & \textbf{0.99} & \textbf{0.99} & \textbf{0.99} & \textbf{0.96} & \textbf{0.98} & \textbf{0.99} & \textbf{0.99} & \textbf{0.99} & 0.86 & 0.95 & 0.99\\
RoSteALS (48)~\cite{bui2023rosteals} & \textbf{1.00} & 0.56 & 0.53 & 0.49 & 0.51 & 0.51 & 0.50 & 0.49 & 0.83 & 0.50 & 0.50 & 0.50 & 0.50 & 0.76 & \textbf{1.00} & \textbf{1.00} & \textbf{1.00}\\
SS$^*$ (48)~\cite{fernandez2023stable} & 0.99 & 0.96 & 0.81 & 0.65 & 0.95 & 0.97 & 0.98 & \textbf{0.99} & \textbf{0.99} & 0.91 & 0.95 & 0.97 & 0.98 & \textbf{0.99} & 0.80 & 0.87 & 0.94\\
\textbf{LaWa}$^*$ (48) & \textbf{1.00} & \textbf{0.99} & \textbf{0.99} & 0.89 & 0.95 & 0.98 & \textbf{0.99} & \textbf{0.99} & \textbf{0.99} & 0.85 & 0.96 & 0.98 & \textbf{0.99} & 0.99 &0.89 & 0.98 & \textbf{1.00} \\
\textbf{LaWa-post-gen} (48) & \textbf{1.00} & \textbf{0.99} & 0.97 & 0.83 & 0.93 & \textbf{1.00} & \textbf{0.99} & \textbf{0.99} & \textbf{0.99} & 0.84 & 0.90 & \textbf{0.99} & \textbf{0.99} & \textbf{0.99} & 0.95 & 0.95 & 0.98 \\

\toprule  
\multirow{2}{*}[-15pt]{\thead{Method (bit$\#$)}} &\multicolumn{17}{c}{Bit accuracy $\uparrow$} \\
& \rot{90}{Resize 0.6} & \rot{90}{Blur 3} & \rot{90}{Blur 7} & \rot{90}{Blur 11} & \rot{90}{Blur 15} & \rot{90}{Blur 19} & \rot{90}{Contr. 0.5} & \rot{90}{Contr. 1.5} & \rot{90}{Contr. 2.0} & \rot{90}{Bright. 0.5} & \rot{90}{Bright. 1.5} & \rot{90}{Bright. 2.0} & \rot{90}{JPEG 10} & \rot{90}{JPEG 30} & \rot{90}{JPEG 50} & \rot{90}{JPEG 70} & \rot{90}{JPEG 90}\\
\midrule
\midrule

DCT-DWT (32)~\cite{cox2007digital} & 0.51 & 0.78 & 0.56 & 0.53 & 0.51 & 0.51 & 0.50 & 0.52 & 0.50 & 0.50 & 0.51 & 0.50 & 0.50 & 0.50 & 0.51 & 0.52 & 0.55\\
Hidden (30)~\cite{hiddenpretrained} & 0.89 & 0.84 & 0.76 & 0.66 & 0.56 & 0.51 & 0.75 & 0.78 & 0.75 & 0.79 & 0.77 & 0.74 & 0.46 & 0.48 & 0.50 & 0.53 & 0.64\\
SSL (32)~\cite{fernandez2022watermarking} & \textbf{1.00} & \textbf{1.00} & \textbf{1.00} & \textbf{1.00} & 0.97 & 0.84 & \textbf{1.00} & 0.99 & 0.96 & \textbf{1.00} & 0.99 & 0.95 & 0.57 & 0.83 & 0.95 & 0.99 & \textbf{1.00} \\
RoSteALS (32)~\cite{bui2023rosteals} & 0.98 & 0.98 & 0.98 & 0.98 & 0.98 & \textbf{0.98} & 0.77 & 0.77 & 0.74 & 0.80 & 0.75 & 0.73 & \textbf{0.84} & 0.95 & 0.97 & 0.98 & 0.98\\
RivaGan (32)~\cite{zhang2019robust} & 0.99 & 0.99 & 0.99 & 0.98 & 0.96 & 0.91 & 0.74 & 0.84 & 0.81 & 0.77 & 0.83 & 0.80 & 0.63 & 0.85 & 0.94 & 0.98 & 0.99 \\
\textbf{LaWa}$^*$ (32) & \textbf{1.00} & \textbf{1.00} & \textbf{1.00} & 0.99 & 0.98 & \textbf{0.98} & \textbf{1.00} & \textbf{1.00} & \textbf{1.00} & \textbf{1.00} & \textbf{1.00} & \textbf{1.00} & 0.72 & \textbf{0.97} & \textbf{1.00} & \textbf{1.00} & \textbf{1.00}\\
\textbf{LaWa-post-gen} (32) & \textbf{1.00} & \textbf{1.00} & \textbf{1.00} & \textbf{1.00} & \textbf{0.99} & \textbf{0.98} & \textbf{1.00} & \textbf{1.00} & \textbf{1.00} & \textbf{1.00} & \textbf{1.00} & \textbf{1.00} & 0.71 & \textbf{0.97} & \textbf{1.00} & \textbf{1.00} & \textbf{1.00} \\

\midrule 
SSL (48)~\cite{fernandez2022watermarking} & \textbf{1.00} & \textbf{1.00} & \textbf{1.00} & 0.99 & 0.97 & 0.83 & \textbf{1.00} & 0.98 & 0.94 & \textbf{1.00} & 0.99 & 0.94 & 0.59 & 0.82 & 0.94 & 0.99 & \textbf{1.00}\\
FNNS (48)~\cite{kishore2021fixed} & \textbf{1.00} & 0.82 & 0.54 & 0.51 & 0.50 & 0.50 & 0.53 & 0.90 & 0.88 & 0.54 & 0.88 & 0.85 & 0.59 & 0.81 & 0.89 & 0.94 & 0.98\\
RoSteALS (48)~\cite{bui2023rosteals} & \textbf{1.00} & \textbf{1.00} & \textbf{1.00} & \textbf{1.00} & \textbf{1.00} & \textbf{1.00} & 0.88 & 0.89 & 0.88 & 0.92 & 0.88 & 0.86 & \textbf{0.97} & \textbf{1.00} & \textbf{1.00} & \textbf{1.00} & \textbf{1.00}\\
SS$^*$ (48)~\cite{fernandez2023stable} & 0.98 & 0.96 & 0.78 & 0.63 & 0.55 & 0.50 & 0.97 & 0.98 & 0.97 & 0.97 & 0.98 & 0.96 & 0.63 & 0.84 & 0.89 & 0.92 & 0.95\\
\textbf{LaWa}$^*$ (48) & 0.99 & \textbf{1.00} & 0.99 & 0.99 & 0.96 & 0.93 & \textbf{1.00} & \textbf{1.00} & \textbf{1.00} & \textbf{1.00} & \textbf{1.00} & \textbf{1.00} & 0.82 & 0.98 & \textbf{1.00} & \textbf{1.00} & \textbf{1.00}\\
\textbf{LaWa-post-gen} (48) & 0.98 & \textbf{1.00} & 0.99 & 0.98 & 0.95 & 0.92 & \textbf{1.00} & \textbf{1.00} & \textbf{1.00} & \textbf{1.00} & \textbf{1.00} & \textbf{1.00} & 0.80 & 0.98 & \textbf{1.00} & \textbf{1.00} & \textbf{1.00}\\
\bottomrule
\end{tabular}}
\vspace{-15pt}
\end{table*}

\subsection{General Image Watermarking}
In the main body of the paper, the results of LaWa for general image watermarking are reported using CLIC dataset. We also use DIV2K~\cite{agustsson2017ntire} dataset, which has larger variety of scenes compared to CLIC dataset to evaluate the performance of LaWa against baselines. All the 900 images in the training and validation set of Div2K are used in the experiments. For all images, while maintaining the aspect ratio, larger dimension of each image is resized to 1024 to avoid memory limitation issue for some of the baselines. The quality and the robustness for LaWa as well as the baselines are shown in \cref{tab:supp_tab3}. LaWa outperforms baseline methods in the majority of the attacks. For image quality, it has comparable perceptual quality to the baseline methods. In particular, for 48-bit scenario, FNNS has higher PSNR, but it suffers from significantly high computational complexity as well as low robustness against many attacks such as blur, contrast, and brightness.

\newcommand{\setWidth}{48pt}
\begin{table}[t]
\centering
\caption{
Performance of LaWa for different generative tasks and different image modifications attacks.
}
\label{tab:supp_tab2}
\resizebox{0.98\textwidth}{!}{
\begin{tabular}{l|M{20pt}M{\setWidth}M{\setWidth}M{\setWidth}M{\setWidth}M{\setWidth}M{35pt}M{\setWidth}M{\setWidth}M{35pt}M{30pt}}
\toprule  
\multirow{2}{*}[2pt]{\thead{Generative\\Task}} & \multicolumn{11}{c}{Bit accuracy $\uparrow$} \\
& None & C. Crop 0.4 & C. Crop 0.2 & R. Crop 0.4 & R. Crop 0.1 & Resize 0.7 & Blur 11 & Contr. 1.5 & Bright. 2.0 & JPEG 60 & Comb. \\
\midrule
Inpainting & 1.00 & \hspace{10pt} 0.99 & \hspace{10pt} 0.98 & \hspace{10pt} 0.91 & \hspace{10pt} 0.86 & \hspace{9pt} 0.97 & \hspace{5pt} 0.98 & \hspace{7pt} 1.00 & \hspace{10pt} 1.00 & \hspace{6pt} 0.98 & \hspace{3pt} 0.96 \\
Super-Resolution & 1.00 & \hspace{10pt} 0.98 & \hspace{10pt} 0.97 & \hspace{10pt} 0.91 & \hspace{10pt} 0.90 & \hspace{9pt} 0.97 & \hspace{5pt} 0.95 & \hspace{7pt} 1.00 & \hspace{10pt} 0.99 & \hspace{6pt} 0.98 & \hspace{3pt} 0.92 \\
Image Editing & 1.00 & \hspace{10pt} 0.99 & \hspace{10pt} 0.98 & \hspace{10pt} 0.97 & \hspace{10pt} 0.87 & \hspace{9pt} 0.97 & \hspace{5pt} 0.98 & \hspace{7pt} 1.00 & \hspace{10pt} 0.98 & \hspace{6pt} 1.00 & \hspace{3pt} 0.94\\
Text-to-Image & 1.00 & \hspace{10pt} 0.99  & \hspace{10pt} 0.98 & \hspace{10pt} 0.90 & \hspace{10pt} 0.85 & \hspace{9pt} 0.98 & \hspace{5pt} 0.99 & \hspace{7pt} 0.99 & \hspace{10pt} 1.00 & \hspace{6pt} 1.00 & \hspace{3pt} 0.94 \\
\bottomrule
\end{tabular}}
\end{table}

\subsection{Ablation Studies}
In this section, more ablation analysis over different components of the proposed method is provided.

\textbf{Different Loss Terms.}
We study the effect of different loss terms on the perceptual quality of generated images in the main body of the paper and show that using $\mathcal{L}_{MSE}$, $\mathcal{L}_{LPIPS}$, and $\mathcal{L}_{Critic}$ together achieves the best performance. In \cref{fig:supp_fig1}, the effect of each of these loss terms on the watermarked images is qualitatively illustrated. When $\mathcal{L}_{Critic}$ is not used, visual artifacts are present in the image. Adding $\mathcal{L}_{Critic}$ improves the quality of the image. Furthermore, using $\mathcal{L}_{LPIPS}+\mathcal{L}_{Critic}$ is more effective than $\mathcal{L}_{MSE}+\mathcal{L}_{Critic}$, while best perceptual quality is achieved when three losses are used together.

\def\rot{\rotatebox}
\newcolumntype{M}[1]{>{\centering\arraybackslash}m{#1}}
\begin{table*}[t]
\centering
\caption{
Comparison results of LaWa with existing post-generation image watermarking methods in terms of quality, embedding time, and robustness against various attacks on DIV2K dataset.}
\vspace{-3pt}
\label{tab:supp_tab3}
\resizebox{0.99\textwidth}{!}{
\begin{tabular}{l|cc|M{21pt}|M{23pt}M{27pt}M{27pt}M{27pt}M{27pt}M{27pt}M{27pt}M{27pt}M{27pt}M{27pt}|M{16pt}}
\toprule  
\multirow{2}{*}[-15pt]{\thead{Method (bit$\#$)}} & \multicolumn{2}{c|}{Image quality} & &\multicolumn{10}{c}{Bit accuracy $\uparrow$} & \\
& PSNR/SSIM $\uparrow$ & LPIPS/SIFID $\downarrow$ & Emb. time (ms) &\rot{45}{None} & \rot{45}{C. Crop 0.1} & \rot{45}{R. Crop 0.1} & \rot{45}{Resize 0.7} & \rot{45}{Rot. 15} & \rot{45}{Blur} & \rot{45}{Contr. 2.0} & \rot{45}{Bright. 2.0} & \rot{45}{JPEG 70} & \rot{45}{Comb.} & \rot{45}{Ave.} \\

\midrule
\multicolumn{15}{c}{\textbf{DIV2K}} \\
\midrule

DCT-DWT (32)~\cite{cox2007digital} & 36.73/0.97 & \textbf{0.01}/\textbf{0.01} & 224 & 0.89 & 0.50 & 0.50 & 0.50 & 0.51 & 0.51 & 0.50 & 0.50 & 0.51 & 0.50 & 0.54 \\
Hidden (30)~\cite{hiddenpretrained} & 32.34/0.95 & 0.04/0.06 & 25 & 0.89 & 0.89 & 0.89 & 0.86 & 0.76 & 0.66 & 0.75 & 0.73 & 0.56 & 0.62 & 0.76 \\
SSL (32)~\cite{fernandez2022watermarking} & 35.04/0.94 & 0.10/0.11 & 1181 & \textbf{1.00} & 0.83 & 0.78 & \textbf{1.00} & \textbf{1.00} & \textbf{1.00} & 0.97 & 0.95 & 0.99 & 0.90 & 0.94 \\
RoSteALS (32)~\cite{bui2023rosteals} & 25.91/0.88 & 0.09/0.07 & 236 & \textbf{1.00} & 0.50 & 0.50 & \textbf{1.00} & 0.50 & \textbf{1.00} & 0.90 & 0.87 & \textbf{1.00} & 0.50 & 0.78 \\
RivaGan (32)~\cite{zhang2019robust} & \textbf{42.16}/\textbf{0.98} & 0.03/0.04 & 81 & \textbf{1.00} & 0.97 & 0.95 & 0.99 & 0.94 & 0.99 & 0.86 & 0.84 & 0.98 & 0.96 & 0.95 \\
\textbf{LaWa} (32) & 35.19/0.92 & 0.05/0.04 & 34 & \textbf{1.00} &  \textbf{1.00} & \textbf{0.96} & 0.97 & 0.99 & \textbf{1.00} & \textbf{1.00} & \textbf{1.00} & \textbf{1.00} & \textbf{1.00} & \textbf{0.99} \\

\midrule 
SSL (48)~\cite{fernandez2022watermarking} & 35.04/0.94 & 0.10/0.11 & 1181 & \textbf{1.00} & 0.81 & 0.76 & \textbf{0.99} & \textbf{0.99} & \textbf{1.00} & 0.96 & 0.95 & 0.98 & 0.88 & 0.93 \\
FNNS (48)~\cite{kishore2021fixed} & \textbf{37.41}/\textbf{0.97} & \textbf{0.03}/0.08 & 1003 & \textbf{1.00} & \textbf{1.00} & \textbf{0.97} & 0.93 & 0.75 & 0.51 & 0.85 & 0.81 & 0.90 & 0.90 & 0.86 \\
RoSteALS ($48$)~\cite{bui2023rosteals} & 25.62/0.88 & 0.09/\textbf{0.07} & 236 & 0.99 & 0.50 & 0.50 & \textbf{0.99} & 0.47 & 0.99 & 0.82 & 0.80 & \textbf{0.99} & 0.50 & 0.76\\
\textbf{LaWa} ($48$) & 34.57/0.91 & 0.07/0.08 & 34 & \textbf{1.00} & \textbf{1.00} & 0.95 & 0.97 & \textbf{0.99} & 0.99 & \textbf{1.00} & \textbf{1.00} & \textbf{0.99} & \textbf{0.91} & \textbf{0.98} \\
\bottomrule
\end{tabular}}
\end{table*}

\newcommand{\setwidth}{0.4\textwidth}
\begin{table*}[t]
     \centering
    \resizebox{0.95\textwidth}{!}{ 
    \begin{tabular}{ c c c c c  c c}
    \toprule
    {\LARGE $Original$} & {\LARGE $MSE+LPIPS+Critic$} & {\LARGE $MSE$} & {\LARGE $LPIPS$} & {\LARGE $MSE+LPIPS$} & {\LARGE $MSE+Critic$} & {\LARGE $LPIPS+Critic$} \\
    
     \includegraphics[width=\setwidth]{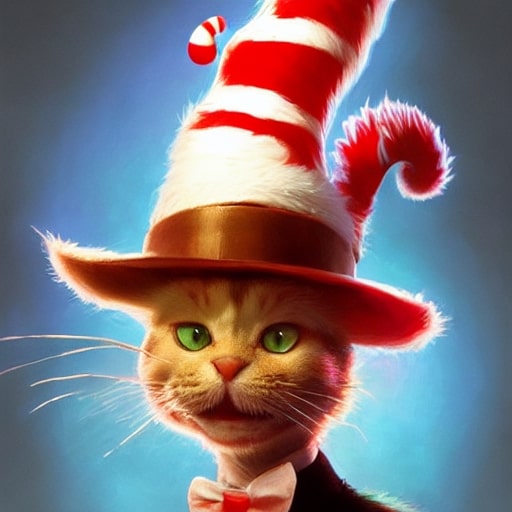}
      &
      \includegraphics[width=\setwidth]{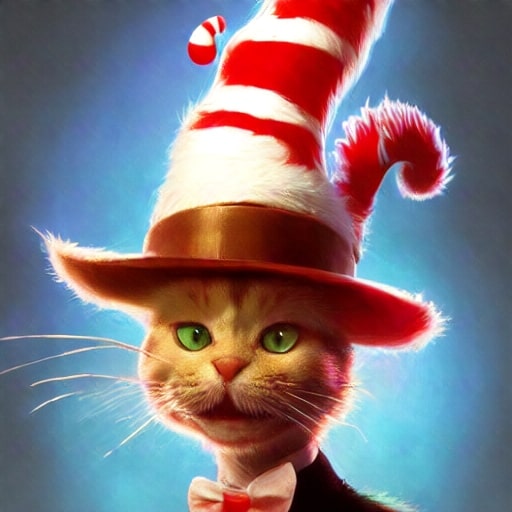}
      &
       \includegraphics[width=\setwidth]{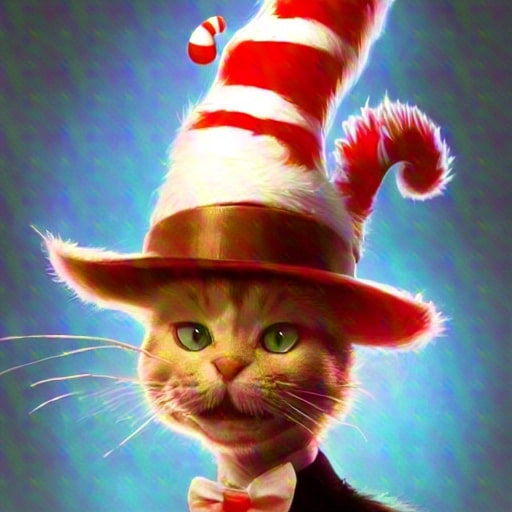}
            &
        \includegraphics[width=\setwidth]{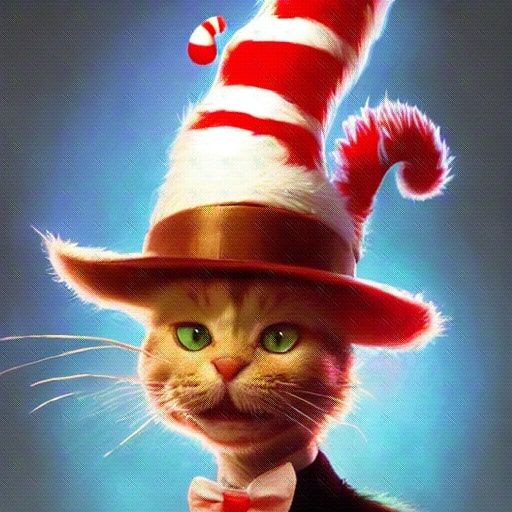}
      &
        \includegraphics[width=\setwidth]{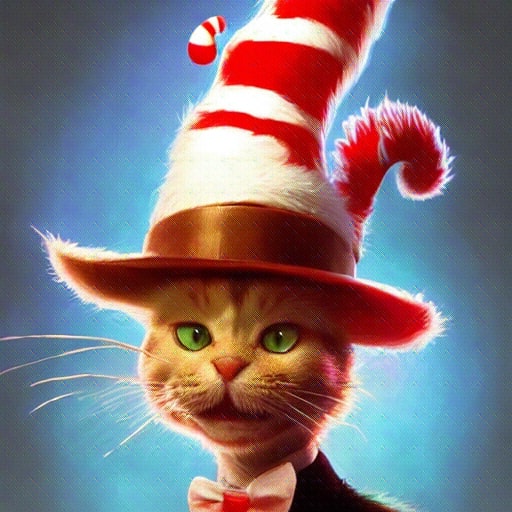}
      &
        \includegraphics[width=\setwidth]{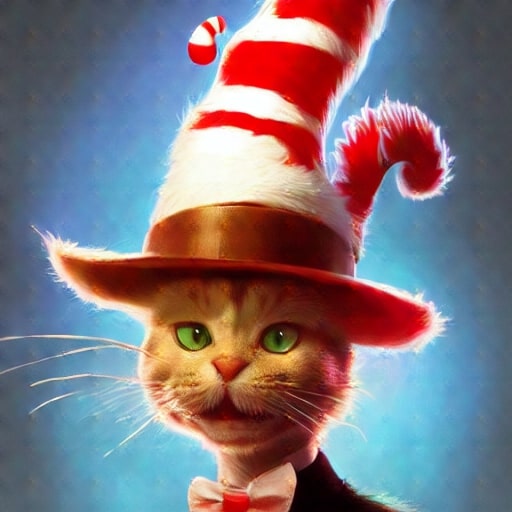}
      &
        \includegraphics[width=\setwidth]{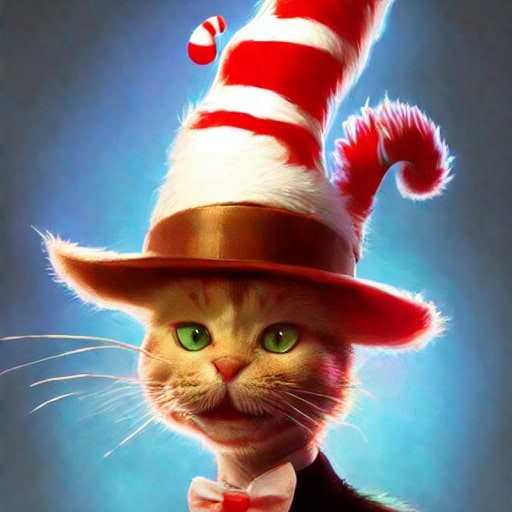}
      \\
      \includegraphics[width=\setwidth]{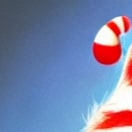}
      &
      \includegraphics[width=\setwidth]{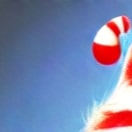}
      &
       \includegraphics[width=\setwidth]{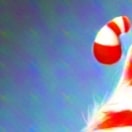}
            &
        \includegraphics[width=\setwidth]{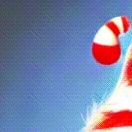}
      &
        \includegraphics[width=\setwidth]{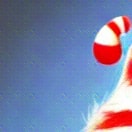}
      &
        \includegraphics[width=\setwidth]{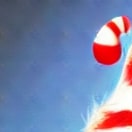}
      &
        \includegraphics[width=\setwidth]{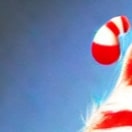}
      \\
      \bottomrule
      \end{tabular}}
      \captionof{figure}{Qualitative effect of using different loss terms for training LaWa.
      } \label{fig:supp_fig1}
\end{table*}

\textbf{Watermark Block Size.}
\Cref{tab:supp_tab4} shows the effect of changing the size $B$ of noise block $b_i \in \mathbb{R}^{B \times B \times C_i}$ using 10 messages (48 bits) and 1k images per message. As we increase $B$, the perceptual quality of the generated images improves. However, when $B$ is set to 16 or 32, LaWa is not robust to rotation and crop attacks anymore. Thus, we use $B=8$ in all our experiments.

\begin{table}[t]
    \centering
    \caption{Effect of the watermark block size.}
    \resizebox{0.8\textwidth}{!}{
    \begin{tabular}{c|M{70pt}M{70pt}M{30pt}M{50pt}M{50pt}M{30pt}M{50pt}}
        \hline
        Block size (B) & PSNR/SSIM $\uparrow$ & LPIPS/SIFID $\downarrow$ & None & C. Cro 0.1 & Contr. 2.0 & Rot. 15 & JPEG 70 \\
        \hline
        2 & 30.88/0.66 & 0.07/0.21 & 0.99 & 0.85 & 0.98 & 0.82 & 0.97 \\
        4 & 32.36/0.81 & 0.06/0.07 & 0.99 & 0.90 & 0.99 & 0.82 & 0.95 \\
        8 & 33.52/0.86 & 0.04/0.02 & \textbf{1.00} & \textbf{0.95} & \textbf{1.00} & \textbf{0.96} & \textbf{1.00} \\
        16 & 34.43/0.88 & 0.03/0.02 & 0.99 & 0.38 & 0.98 & 0.48 & 0.96 \\
        32 & \textbf{34.51/0.91} & \textbf{0.02/0.02} & 0.90 & 0.56 & 0.86 & 0.45 & 0.83 \\ 
        \hline
    \end{tabular}}
    \label{tab:supp_tab4}
\end{table}

\textbf{Number of Embedding Modules.}
\Cref{tab:supp_tab5} illustrates the effect of increasing the number of watermarking modules from 1 to 4, showing an improvement in both robustness and perceptual quality. Specifically, adding more modules significantly improves the robustness to crop and rotation attacks.

\begin{table}[b]
    \centering
    \scriptsize
    \caption{Effect of number of embedding modules.}
    \resizebox{0.9\textwidth}{!}{
    \begin{tabular}{M{27pt}M{27pt}M{27pt}M{27pt}|M{60pt}M{60pt}M{20pt}M{40pt}M{40pt}M{30pt}M{33pt}}
        \hline
        $W_{Emb_0}$ & $W_{Emb_1}$ & $W_{Emb_2}$ & $W_{Emb_3}$  & PSNR/SSIM $\uparrow$ & LPIPS/SIFID $\downarrow$ & None & C. Cro 0.1 & Contr. 2.0 & Rot. 15 & JPEG 70 \\
        \hline
        \checkmark & & & & 30.75/0.79 & 0.07/0.08 & \textbf{1.00} & 0.90 & 0.99 & 0.90 & 0.99\\
        \checkmark & \checkmark & & &  32.83/0.80 &  0.04/0.03 & \textbf{1.00} & 0.94 & 1.00 & 0.94 & 0.99\\
         \checkmark &  \checkmark &  \checkmark  &  &  32.6/0.82 &  0.04/0.02 &  0.99 & 0.94 & \textbf{1.00} & \textbf{0.96} & \textbf{1.00}\\
         \checkmark &  \checkmark &  \checkmark &  \checkmark &  \textbf{33.52/0.86}  &  \textbf{0.04/0.02} &  \textbf{1.00} & \textbf{0.95} & \textbf{1.00} & \textbf{0.96} & \textbf{1.00}\\
        \hline
    \end{tabular}}
    \label{tab:supp_tab5}
\end{table}

\subsection{Other Auto-Encoder Networks}
As discussed in the main text, LDM’s image decoder $\mathcal{D}$ upsamples the generated latent to image by a factor $f$, where $f \in \{4,8,16,32\}$. Furthermore, different decoder networks such as KL-VAE~\cite{rombach2022high} and VQ-VAE~\cite{esser2021taming} can be used for LDMs. To show the generality of LaWa to different networks and downsampling factors, we train LaWa using different auto-encoders and factors. The corresponding results are shown in \cref{tab:supp_tab6}. For KL-f8 model, we use 10 messages and generate 1K images per message using LDM model. For other models, we use the frozen encoder of each model to map 1K images to the latent space. We then reconstruct one image using the original decoder and one image using the modified decoder. Each image is watermarked using a random 48-bit message.

As shown in \cref{tab:supp_tab6}, LaWa achieves high perceptual quality and high robustness against attacks in all scenarios, which shows the applicability of LaWa to different auto-encoders and downsampling factors. It should be noted that as the downsampling factor increases, the perceptual quality of the generated images as well as the bit extraction accuracy experience minor degradation. This is due to more compression of the latent providing less information for generating and watermarking the images.

\subsection{Theoretical vs. Empirical False Positive Rate}
Assuming that extracted bits from vanilla images are independent and identically distributed (i.i.d) Bernoulli random variables with parameter 0.5, a theoretical upper bound for the FPR of detection and attribution can be defined.

\textbf{Detection.}
If we use LaWa only for the detection problem, all images can be generated with the same message $m$. To calculate the FPR in this problem, we test the hypothesis $H_1:M(\hat{m} , m) \ge n$ against the null hypothesis $H_0$: "any random message, $\Tilde{m}$, match with $m$ in more than $n$ bits". Under the null hypothesis, the probability of success follows a binomial distribution where total number of trials is $k$, number of successes is $n$, and the probability of a match for each bit is $0.5$. The calculated probability from the binomial distribution corresponds to the upper bound of the FPR, which happens if $W_{Ext}$ extracts random bits. Therefore, for each $n$, we can obtain the FPR of detection, $FPR_{det}(n)$, using the CDF of the binomial distribution:
\begin{equation}
\vspace{-2pt}
    \footnotesize
    FPR_{det}(n) = Pr\big(M(\Tilde{m}, m) > n\ | H_0\big) = \sum_{i=n+1}^{k}\binom{k}{i}0.5^k.
    \label{eq:binomial_dist}
    \vspace{-2pt}
\end{equation}

By setting the $FPR_{det}$ at a desired level, we can use \cref{eq:binomial_dist} to obtain the minimum required $n$ to be used for watermark matching.

\textbf{Attribution.}
In the attribution problem, we attribute the image to a specific user by matching $\hat{m}$ with one of the messages in $\{m_1, ..., m_N\}$, where $N$ is the total number of users. If $M(\hat{m}, m_i) \ge n$, then the $i$th user is flagged as the creator of the image. The watermark matching is repeated for all $N$ users, which corresponds to having $N$ hypothesises. The attribution FPR, $FPR_{att}(N,n)$, or the probability of a random message, $\Tilde{m}$, to match with at least one message in $\{m_1, ..., m_N\}$ is the global $p$-value~\cite{zhang2013combined}:
\begin{equation}
\vspace{-2pt}
    \small
    FPR_{att}(N,n) = 1 - (1-FPR_{det}(n))^N.
    \label{eq:global_p_value}
    \vspace{-2pt}
\end{equation}

\begin{figure*}[t]
  \centering
    \includegraphics[width=0.5\textwidth]{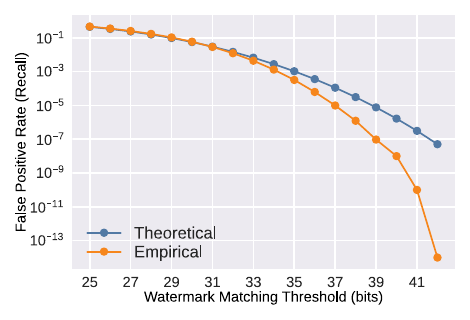}
  \caption{Comparison of the theoretical and empirical false positive rate for the detection task using 48-bit watermarks and different watermark matching thresholds ($n$).
  }
  \label{fig:supp_fig2}
\end{figure*}

The detection and attribution FPRs are calculated with the assumption that the extracted bits from vanilla unwatermarked images are i.i.d Bernoulli random variables. However, the i.i.d assumption may not hold for our trained watermark extractor, meaning that extracted watermark bits may not be independent. Therefore, we empirically calculate the detection false positive rate. For 1 million vanilla images selected from the training set of ImageNet dataset~\cite{deng2009imagenet}, we resize the images to $512\times512$ and extract a watermark from each vanilla image $m_{vanilla}$. We then randomly generate 100 different ground truth watermark messages $m_{gt}$. The watermark matching threshold $n$ is selected from $\{25,...,42\}$. For each pair of watermark message and watermark matching threshold, we perform watermark matching between the 1 million extracted messages and the ground truth watermark message and calculate the FPR. False positive corresponds to vanilla images that their message matches the ground truth message ($M(m_{vanilla},m_{gt}) \ge n$). \Cref{fig:supp_fig2} compares the FPR for both theoretical and empirical cases. The empirical rate is the average over the false positive rate of all 100 messages. For thresholds above 33, the empirical false positive rate is below the theoretical one. This is good result and shows that LaWa achieves lower rate of falsely flagging vanilla images as watermarked compared to the theoretical rate. For example for $n=41$, LaWa flags only 1 of 10 million vanilla image as watermarked ($FPR_{empirical} = 10^{-10}$ vs $FPR_{theoretical} = 10^{-7}$). 

\def\rot{\rotatebox}
\newcolumntype{M}[1]{>{\centering\arraybackslash}m{#1}}
\begin{table}[t]
    \centering
    \caption{Performance of LaWa with different auto-encoder networks and downsampling factors.}
    \vspace{-3pt}
    \resizebox{0.8\textwidth}{!}{
    \begin{tabular}{c|M{70pt}M{70pt}M{30pt}M{50pt}M{50pt}M{30pt}M{50pt}}
        \hline
        Model & PSNR/SSIM $\uparrow$ & LPIPS/SIFID $\downarrow$ & None & C. Cro 0.1 & Contr. 2.0 & Rot. 15 & JPEG 70 \\
        \hline
        KL-f4 & 35.21/0.93 & 0.03/0.03 & 1.00 & 0.96 & 1.00 & 0.94 & 0.97 \\
        KL-f8 & 33.52/0.86 & 0.04/0.02 & 1.00 & 0.95 & 1.00 & 0.96 & 1.00 \\
        KL-f16 & 32.87/0.85 & 0.06/0.03 & 1.00 & 0.93 & 0.98 & 0.91 & 0.95 \\
        VQ-f4 & 36.06/0.94 & 0.02/0.01 & 1.00 & 0.91 & 1.00 & 0.90 & 0.98 \\
        VQ-f8 & 34.21/0.92 & 0.04/0.02 & 1.00 & 0.90 & 0.99 & 0.89 & 0.98 \\
        VQ-f16 & 33.15/0.88 & 0.07/0.02 & 0.99 & 0.88 & 0.97 & 0.85 & 0.95 \\
        \hline
    \end{tabular}}
    \label{tab:supp_tab6}
\end{table}

\subsection{Additional Qualitative Analysis}
In this section, we provide more qualitative results of LaWa's performance.

\textbf{Qualitative Comparison to Other Methods.}
\Cref{fig:supp_fig3,fig:supp_fig4} compare the performance of LaWa to other image watermarking methods for 48-bit and 32-bit watermark messages, respectively. Images are generated with resolution $512\times512$. For the examples provided in both fig-ures, the original, residual (multiplied by 10), and cropped (for better visibility) images are shown.

As shown in the residual images in \cref{fig:supp_fig3} (the 48-bit version), although FNNS seems to provide the lowest amount of noise in the solid backgrounds, it still suffers from the large artifacts on the edges of the objects (e.g., the girl’s face). Such large artifact blocks are also seen for RoSteALS. SSL, on the other hand, gives the worst quality compared to the others, while it tends to provide uniformly distributed small noises all around the images. From the watermarked images, the best quality is presented by LaWa and Stable Signature, but still LaWa adds less noise, especially in the solid backgrounds.

From the cropped images in \cref{fig:supp_fig4} (the 32-bit version), SSL’s low quality is still clear. From the residual images, RivaGAN and DCT-DWT seem to have the lowest noise added to the generated images. However, some red stripped noise lines on the images watermarked by RivaGAN are evident (e.g., on the girl’s face as well as the background). Despite the best quality given by DCT-DWT, it has a very low robustness against attacks as described in the main text. Hidden’s quality is also shown to be bad from both the residual and cropped images as it adds too much noise during the watermarking. In general, LaWa and RoSteALS show the best perceptual quality compared to the others.

\Cref{fig:supp_fig5} provides a qualitative comparison between watermarked images from the CLIC~\cite{toderici2020workshop} dataset. Furthermore, more qualitative results from LaWa are illustrated in \cref{fig:supp_fig6} including the original and watermarked images as well as their residuals. Images are generated using prompts from MS-COCO and MagicPrompt-SD datasets with a resolution of $512 \times 512$.

\section{Implementation Details}
\subsection{Training Details}
At each training step, an image transformation attack $T$ is randomly selected from $\mathcal{T}$, which includes the following:
\begin{enumerate}
    \item Center area crop (C. Crop) with areas between 8\% to 95\%
    \item Random area crop (R. Crop) with areas between 8\% to 95\%
    \item Rotation (Rot.) with angles between 2 and 46 degrees
    \item Area resizing with areas between 0.5 and 1.5 
    \item Brightness (Bright.) with factors between 0.0 and 3.0
    \item Gaussian noise with standard deviations between 0 and 0.05
    \item Blur with kernels between 3 and 19
    \item Contrast (Contr.) with factors between 0.0 and 3.0
    \item JPEG compression with compression factors between 40 and 100
    \item Identity
\end{enumerate}

The parameter for each attack is randomly sampled from a uniform distribution between the lowest and highest parameter. At the training time, we start applying attacks after the bit accuracy reaches a certain threshold, set to 75\%. Furthermore, training for 40 epochs using 8 GPUs with 32GB of memory takes 2 days to finish.

\begin{table*}[tb]
     \centering
    \resizebox{0.95\textwidth}{!}{ 
    \begin{tabular}{ c c c  c  c c}
    \toprule
    Original & LaWa & Stable Signature~\cite{fernandez2023stable} & SSL~\cite{fernandez2022watermarking} & FNNS~\cite{kishore2021fixed} & RoSteALS~\cite{bui2023rosteals} \\
    \midrule
      \includegraphics[width=0.3\textwidth]{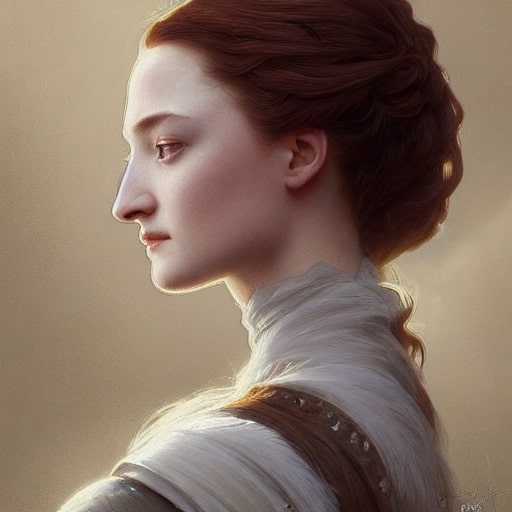}
      &
      \includegraphics[width=0.3\textwidth]{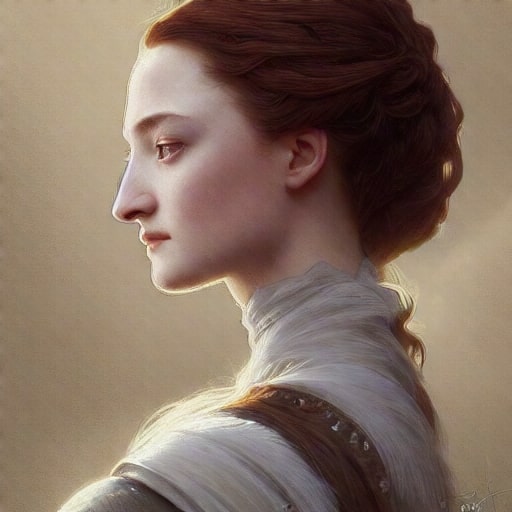}
      &
      \includegraphics[width=0.3\textwidth]{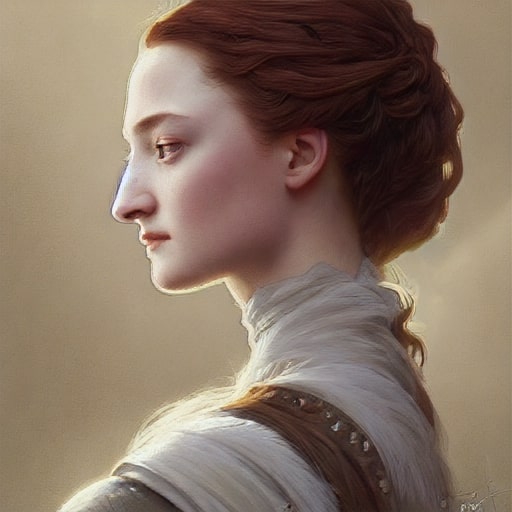}
      & 
      \includegraphics[width=0.3\textwidth]{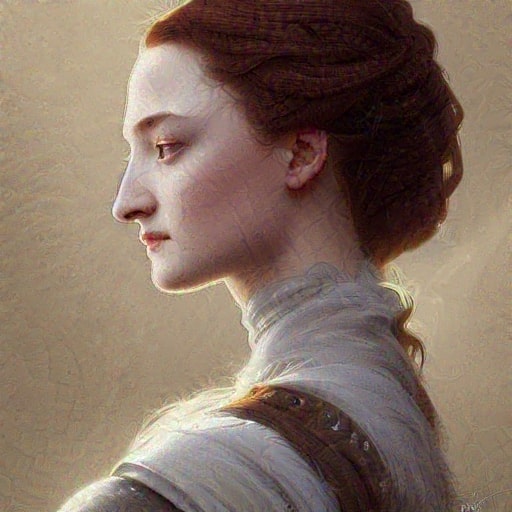}
      & 
      \includegraphics[width=0.3\textwidth]{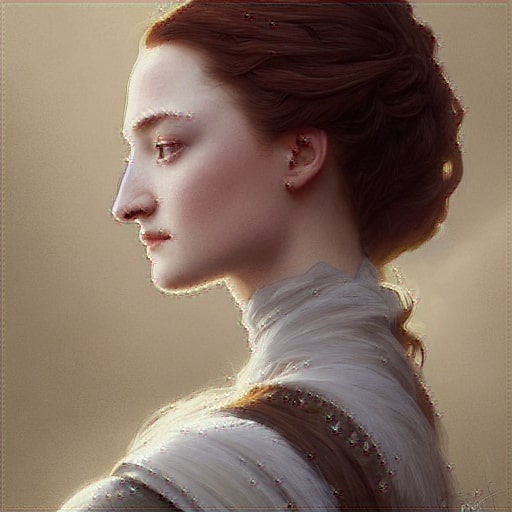}
      & 
      \includegraphics[width=0.3\textwidth]{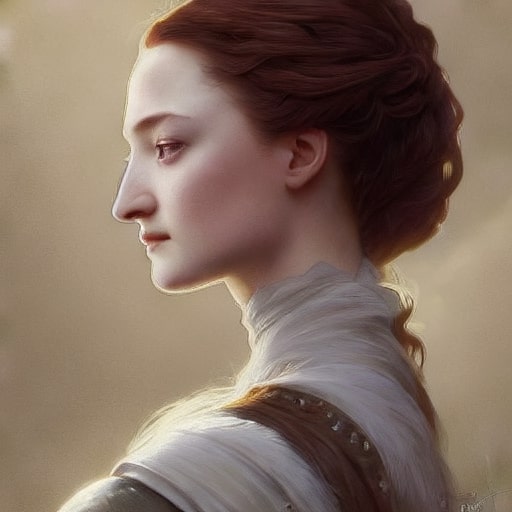}
      \\
      &
      \includegraphics[width=0.3\textwidth]{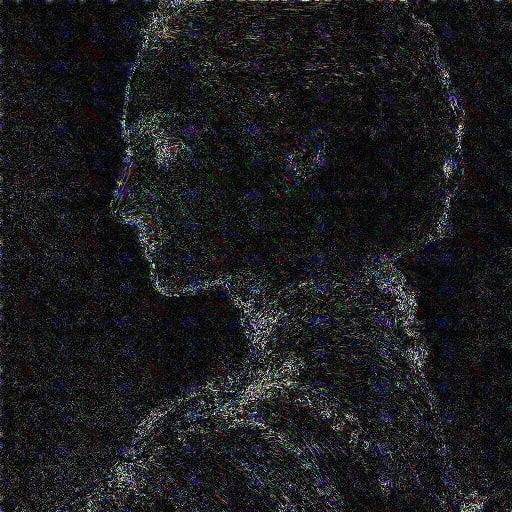}
      &
      \includegraphics[width=0.3\textwidth]{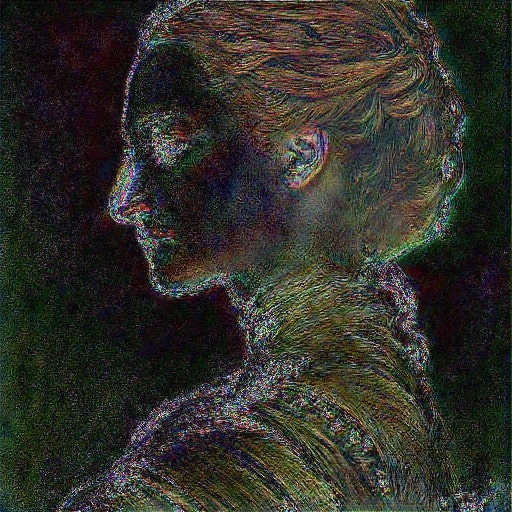}
      & 
      \includegraphics[width=0.3\textwidth]{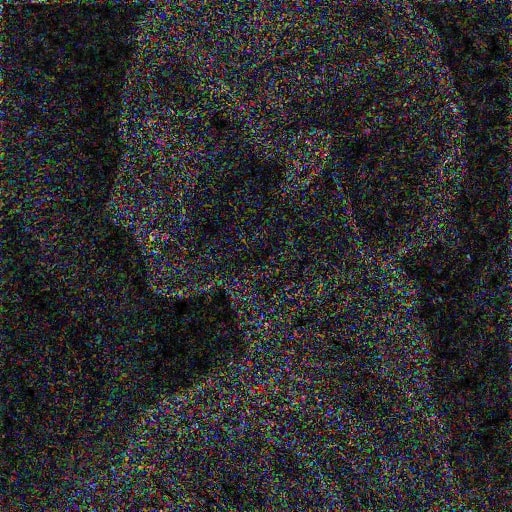}
      & 
      \includegraphics[width=0.3\textwidth]{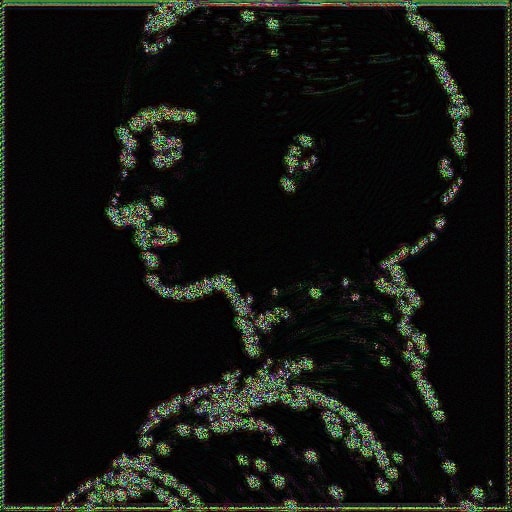}
      & 
      \includegraphics[width=0.3\textwidth]{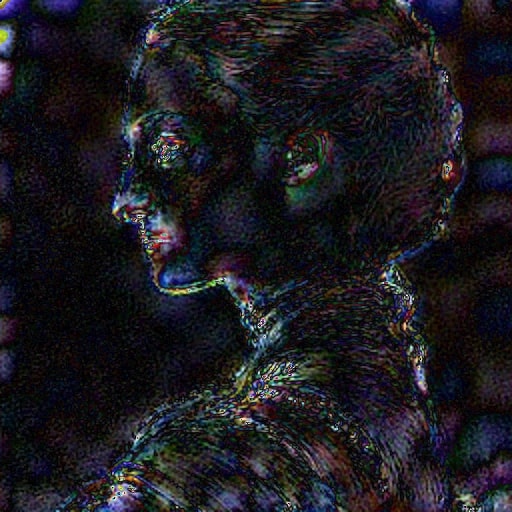}
      \\
      \includegraphics[width=0.3\textwidth]{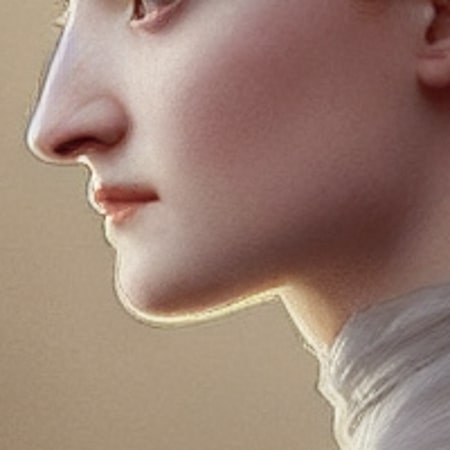}
      &
      \includegraphics[width=0.3\textwidth]{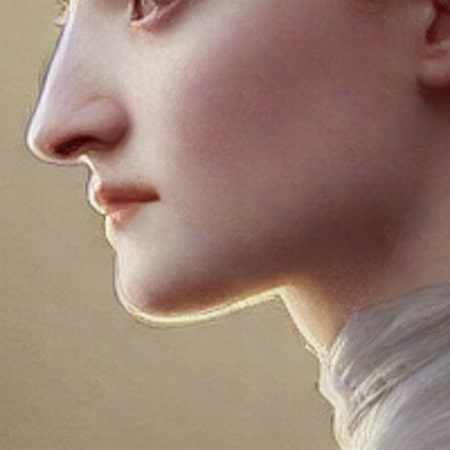}
      &
      \includegraphics[width=0.3\textwidth]{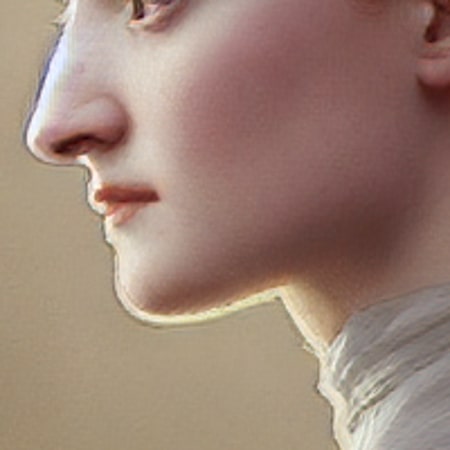}
      & 
      \includegraphics[width=0.3\textwidth]{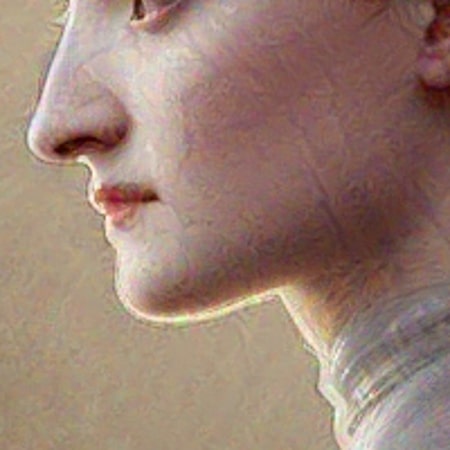}
      & 
      \includegraphics[width=0.3\textwidth]{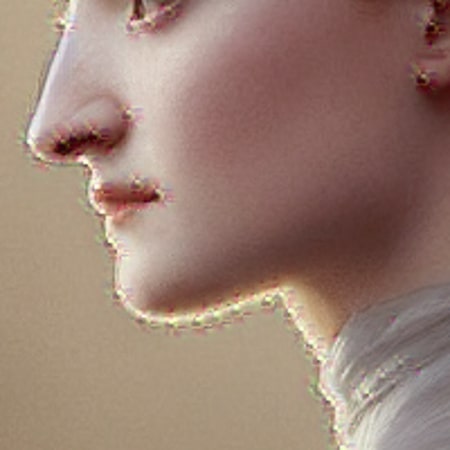}
      & 
      \includegraphics[width=0.3\textwidth]{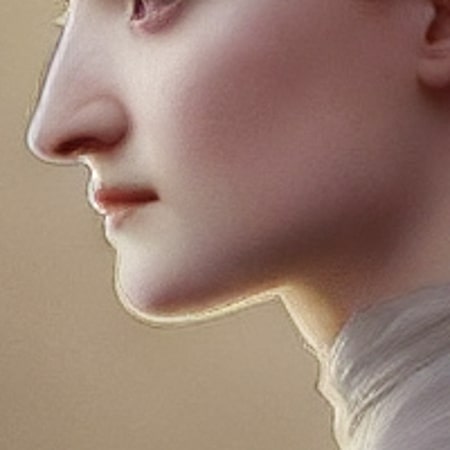}
      \\
      \includegraphics[width=0.3\textwidth]{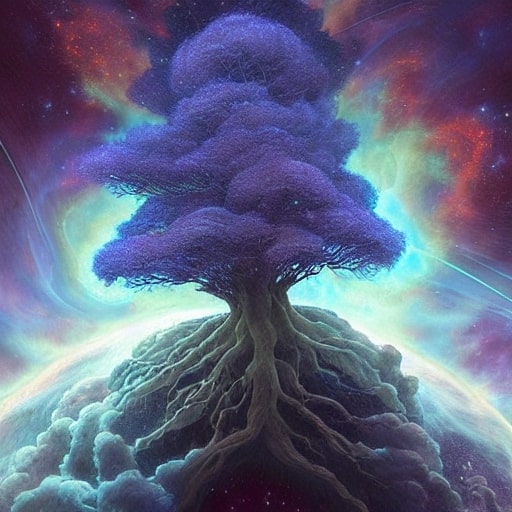}
      &
      \includegraphics[width=0.3\textwidth]{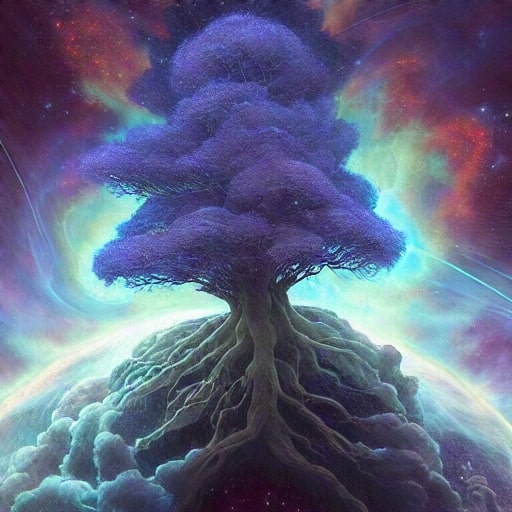}
      &
      \includegraphics[width=0.3\textwidth]{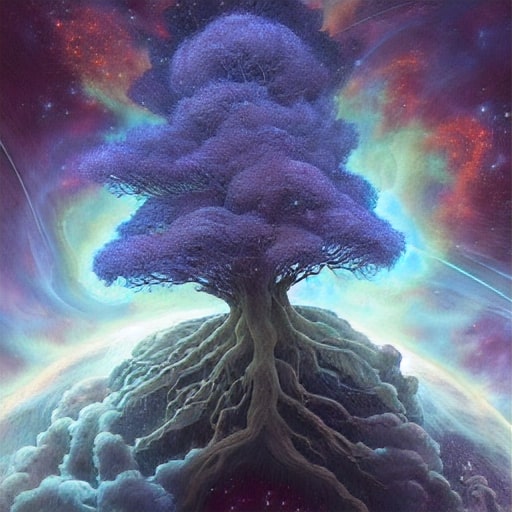}
      & 
      \includegraphics[width=0.3\textwidth]{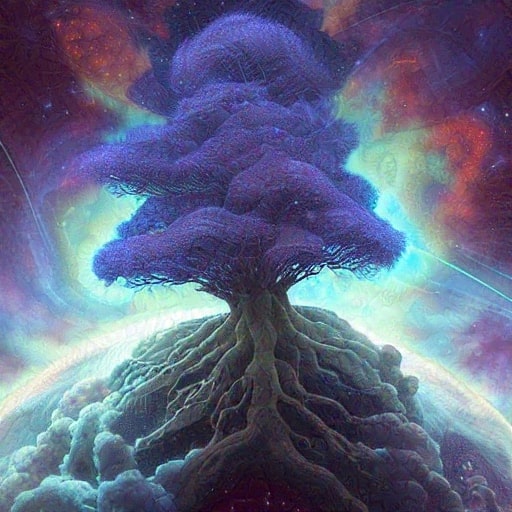}
      & 
      \includegraphics[width=0.3\textwidth]{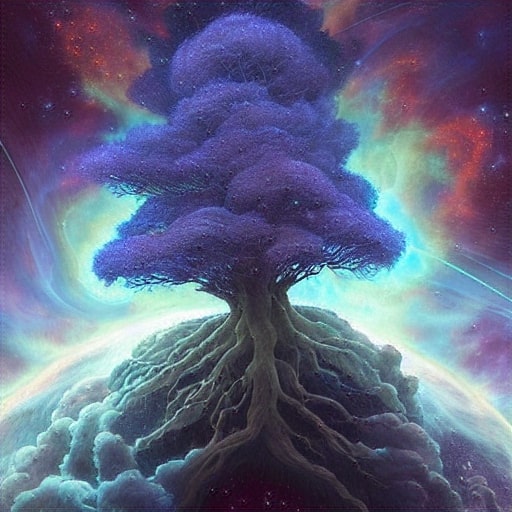}
      & 
      \includegraphics[width=0.3\textwidth]{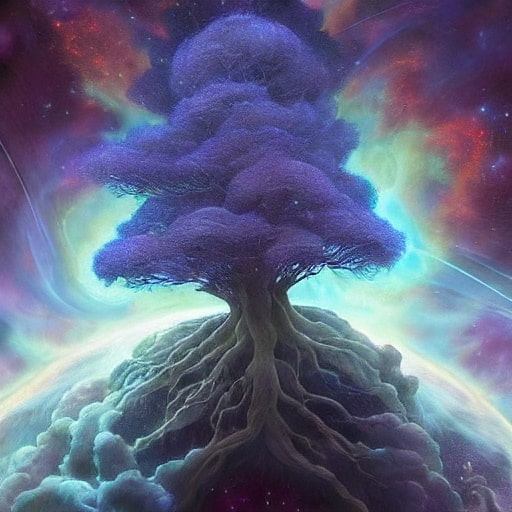}
      \\
      &
      \includegraphics[width=0.3\textwidth]{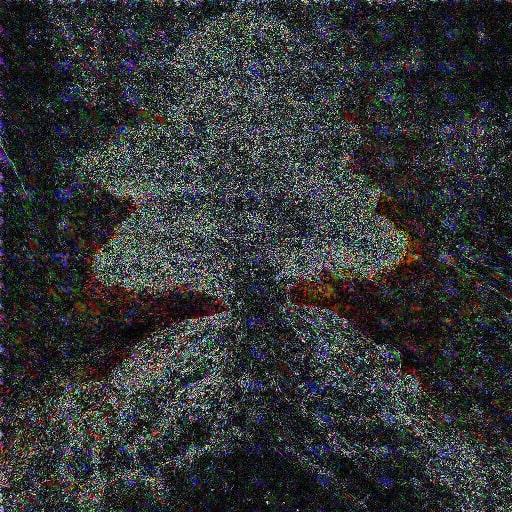}
      &
      \includegraphics[width=0.3\textwidth]{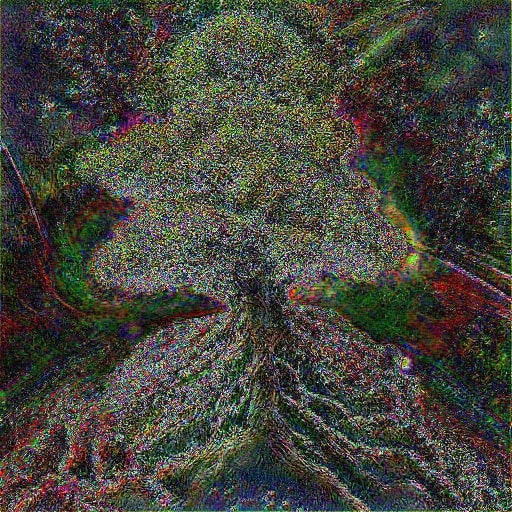}
      & 
      \includegraphics[width=0.3\textwidth]{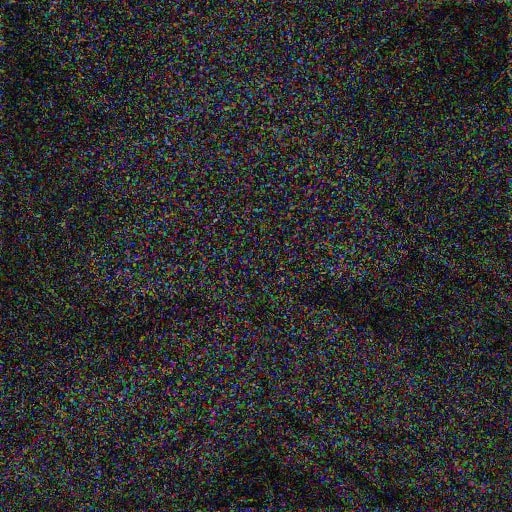}
      & 
      \includegraphics[width=0.3\textwidth]{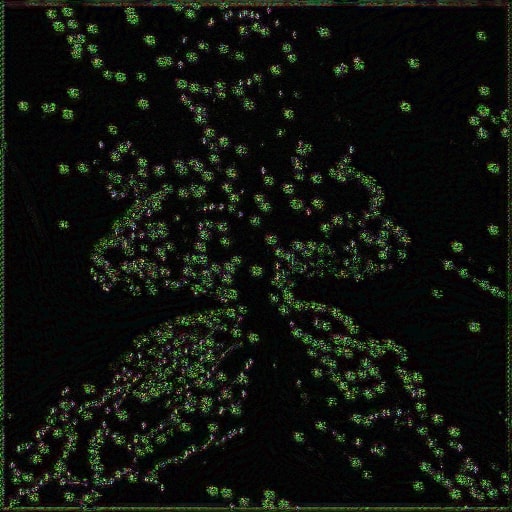}
      & 
      \includegraphics[width=0.3\textwidth]{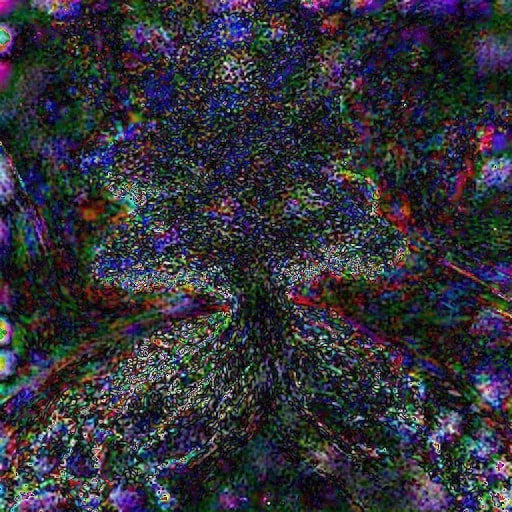}
      \\
      \includegraphics[width=0.3\textwidth]{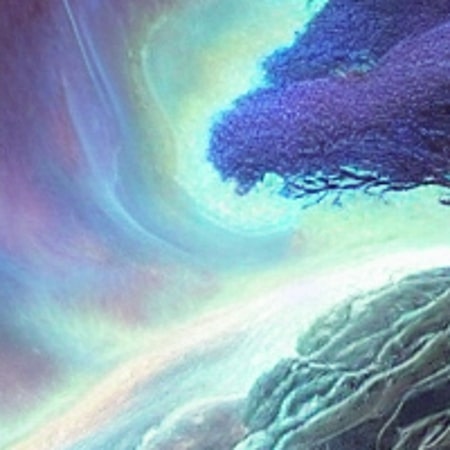}
      &
      \includegraphics[width=0.3\textwidth]{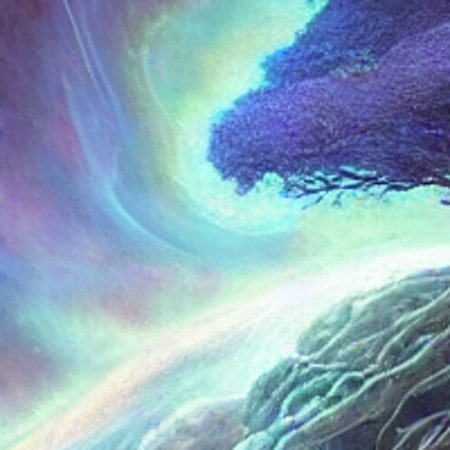}
      &
      \includegraphics[width=0.3\textwidth]{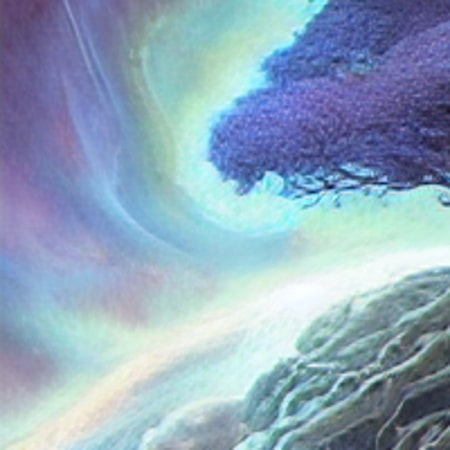}
      & 
      \includegraphics[width=0.3\textwidth]{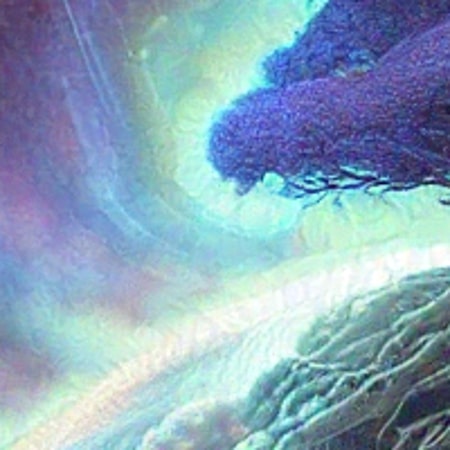}
      & 
      \includegraphics[width=0.3\textwidth]{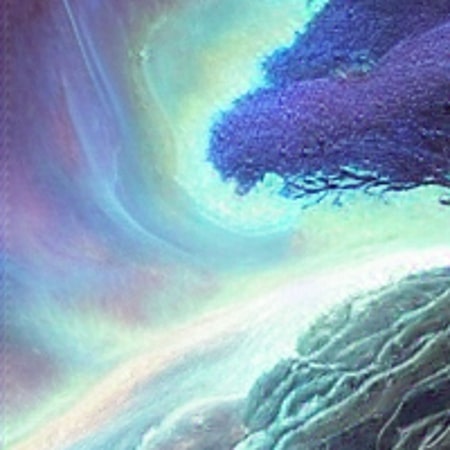}
      & 
      \includegraphics[width=0.3\textwidth]{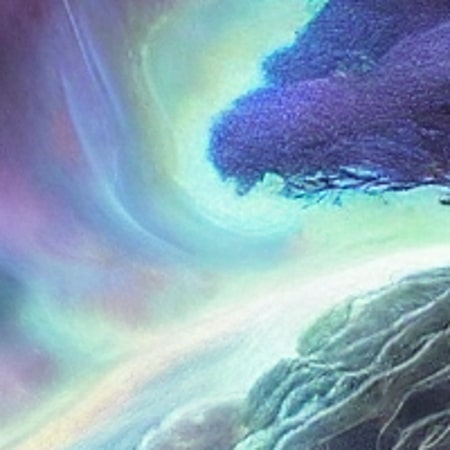}
      \\  
      \bottomrule
      \end{tabular}}
      \captionof{figure}{Qualitative comparison of watermarked images using LaWa and other methods for 48-bit watermarks. 
      } \label{fig:supp_fig3}
    \vspace{-15pt}
\end{table*}

\subsection{More Details on Experiments Settings}
For all the experiments related to the text-to-image task, we use Stable Diffusion version 1.4~\cite{stableDiffusion14Git}. However, for the other image generation tasks (i.e., editing, inpainting, and super-resolution), Stable Diffusion version 2.1~\cite{stableDiffusion21Git} is employed. For all the tasks, their corresponding default parameters of the Stable Diffusion models are used. In all experiments at the test/inference time, we use images with $512 \times 512$ resolution. For all tasks except text-to-image, we create a set of 1K random 48-bit messages and create one image per message. For text-to-image task, we use 10 different messages and create 1K images per message.

\subsection{Image Generation Tasks}
More details about each task is provided below.
\textbf{Image editing.} In this task, an existing image is modified by the diffusion process using a new text description. We follow~\cite{couairon2022diffedit} and use the validation set of MS-COCO~\cite{lin2014microsoft} to generate 1k pairs of image and novel descriptions that the new generated image should follow.

\begin{table*}[tb]
     \centering
    \resizebox{0.95\textwidth}{!}{ 
    \begin{tabular}{ c c c  c  c c c}
    \toprule
    Original & LaWa & SSL~\cite{fernandez2022watermarking} & RoSteALS~\cite{bui2023rosteals} & RivaGan~\cite{zhang2019robust} & Hidden~\cite{hiddenpretrained} & DCT-DWT~\cite{cox2007digital}\\
    \midrule
      \includegraphics[width=0.3\textwidth]{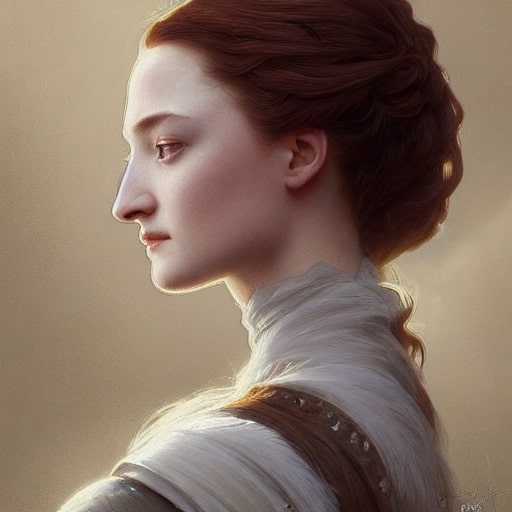}
      &
      \includegraphics[width=0.3\textwidth]{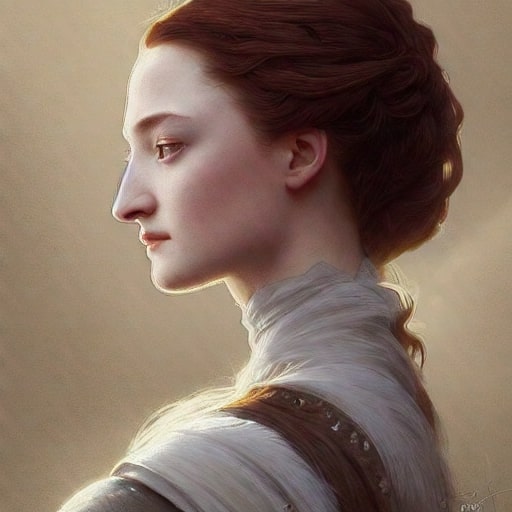}
      &
      \includegraphics[width=0.3\textwidth]{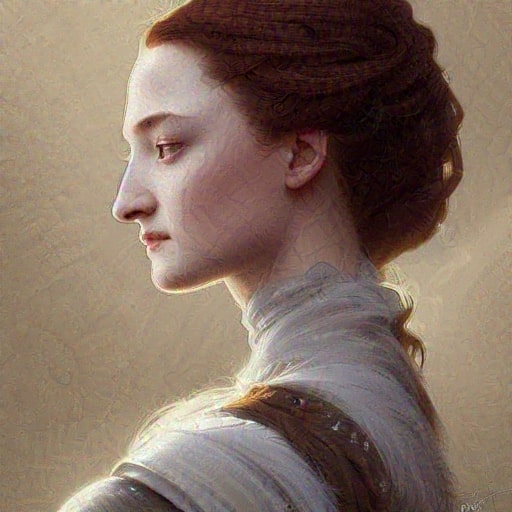}
      & 
      \includegraphics[width=0.3\textwidth]{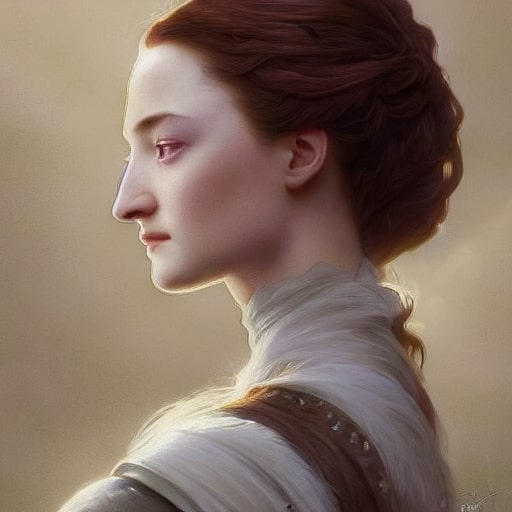}
      & 
      \includegraphics[width=0.3\textwidth]{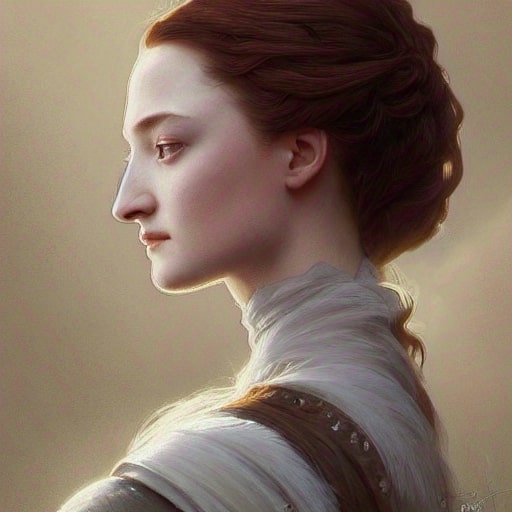}
      & 
      \includegraphics[width=0.3\textwidth]{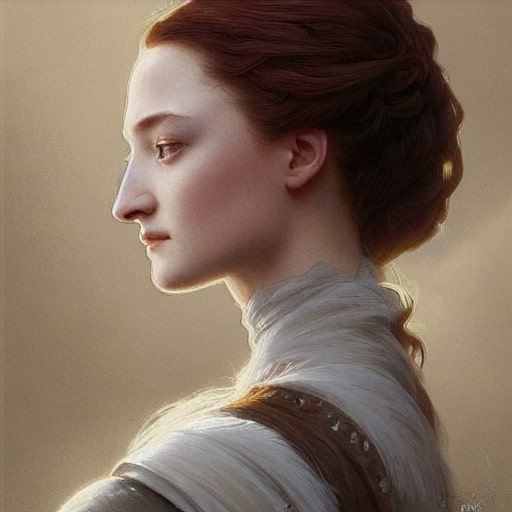}
      & 
      \includegraphics[width=0.3\textwidth]{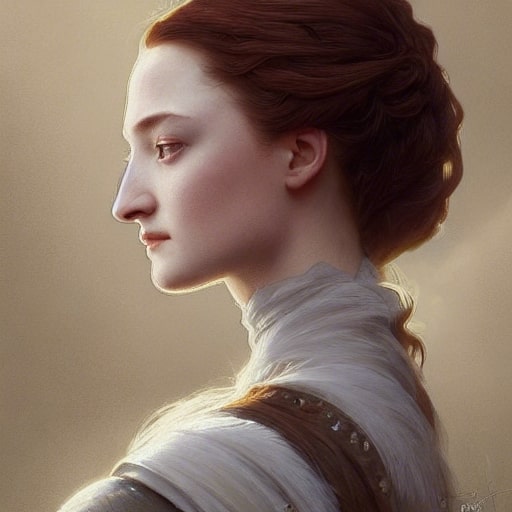}
      \\
      &
      \includegraphics[width=0.3\textwidth]{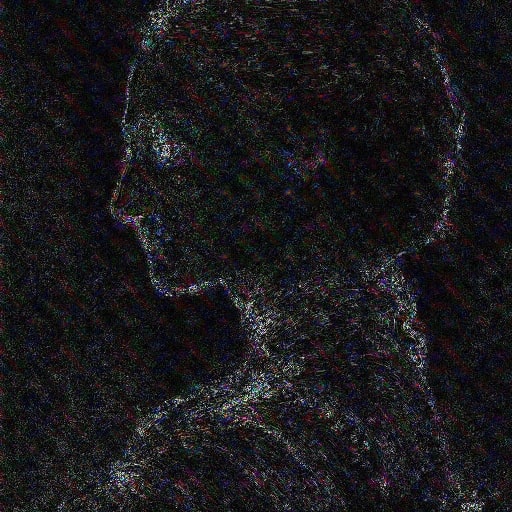}
      &
      \includegraphics[width=0.3\textwidth]{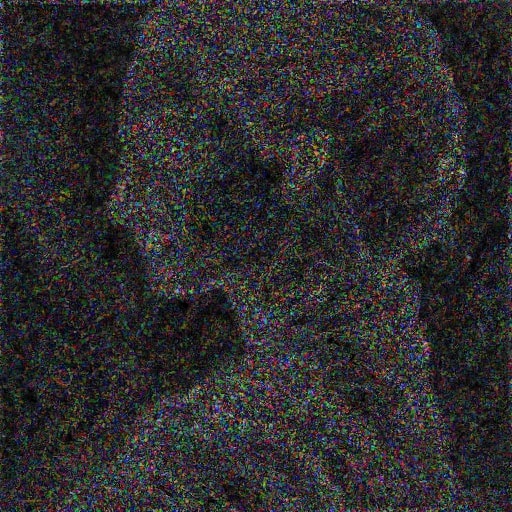}
      & 
      \includegraphics[width=0.3\textwidth]{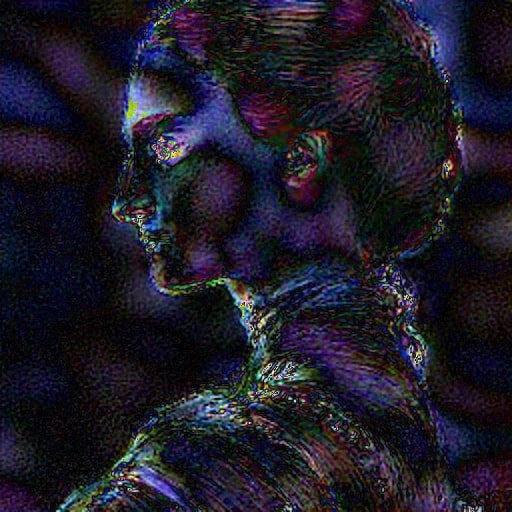}
      & 
      \includegraphics[width=0.3\textwidth]{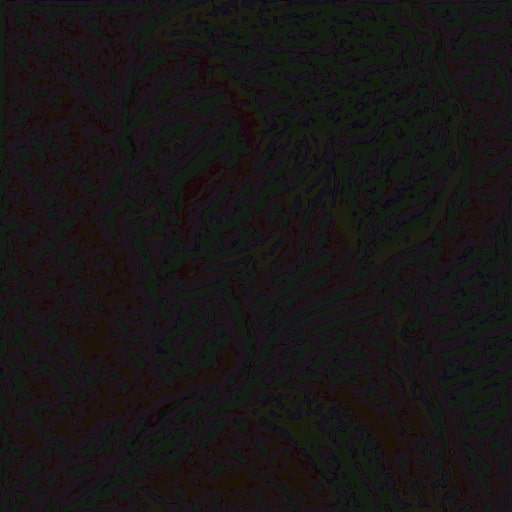}
      & 
      \includegraphics[width=0.3\textwidth]{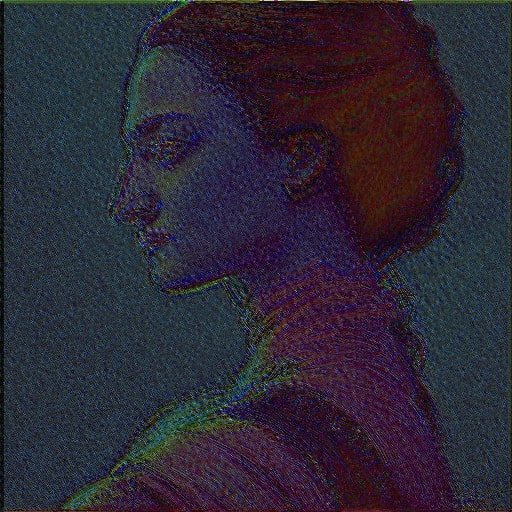}
      & 
      \includegraphics[width=0.3\textwidth]{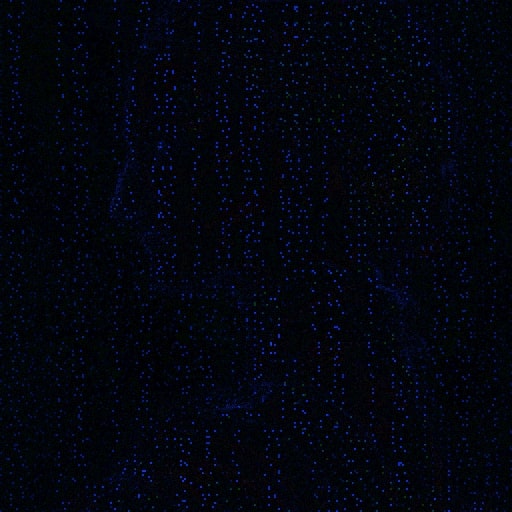}
      \\
      \includegraphics[width=0.3\textwidth]{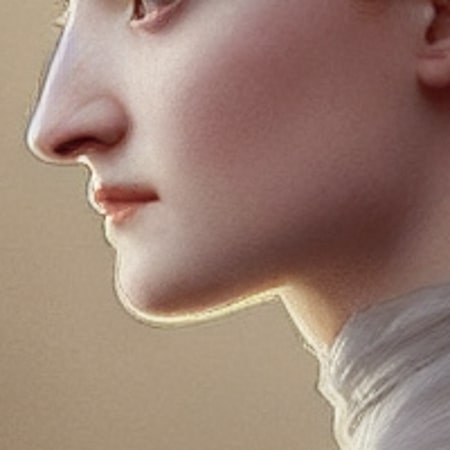}
      &
      \includegraphics[width=0.3\textwidth]{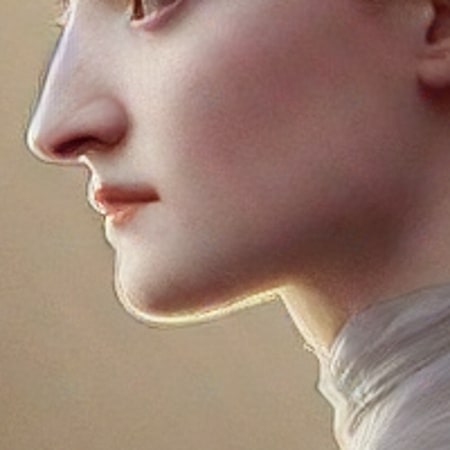}
      &
      \includegraphics[width=0.3\textwidth]{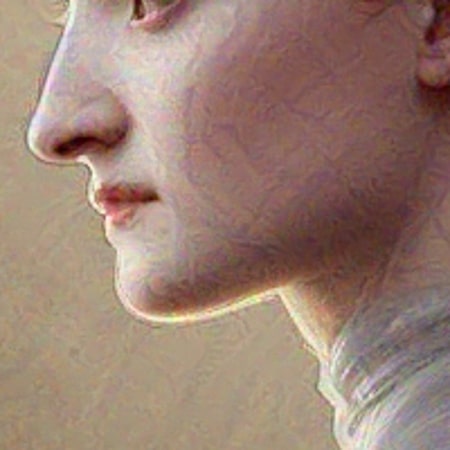}
      & 
      \includegraphics[width=0.3\textwidth]{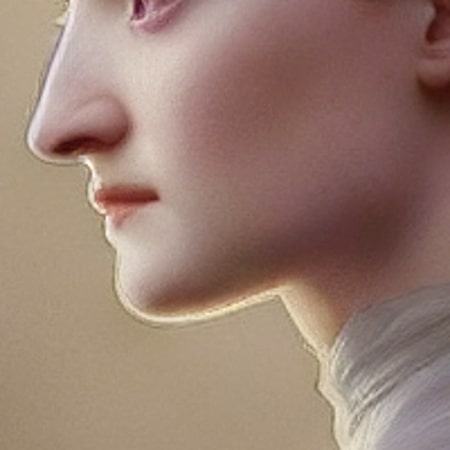}
      & 
      \includegraphics[width=0.3\textwidth]{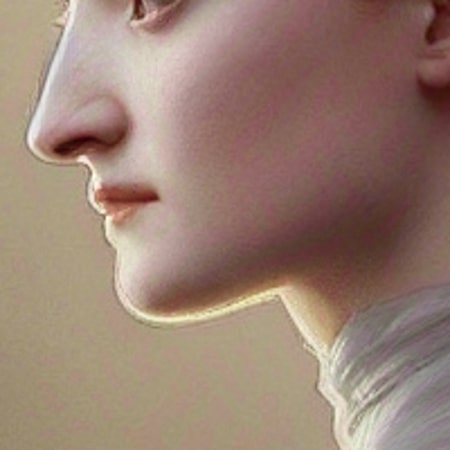}
      & 
      \includegraphics[width=0.3\textwidth]{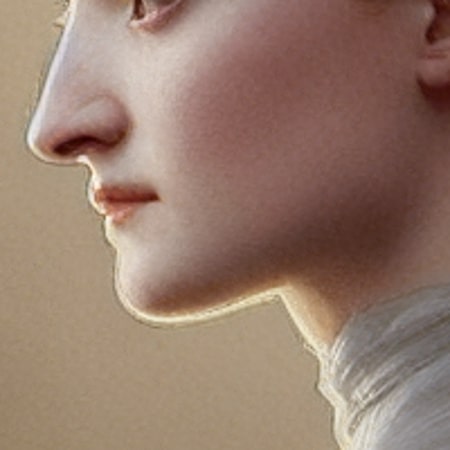}
      & 
      \includegraphics[width=0.3\textwidth]{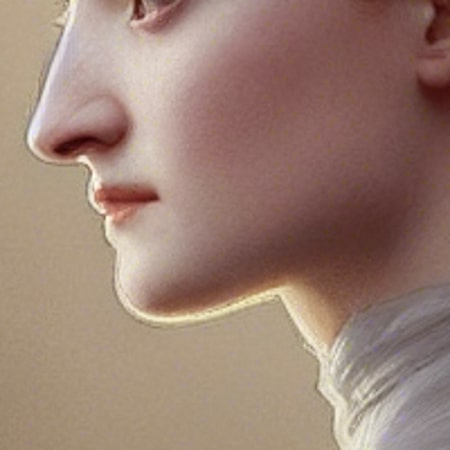}
      \\
      \includegraphics[width=0.3\textwidth]{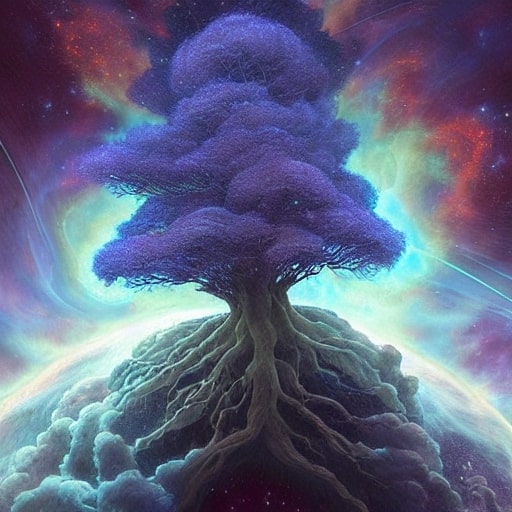}
      &
      \includegraphics[width=0.3\textwidth]{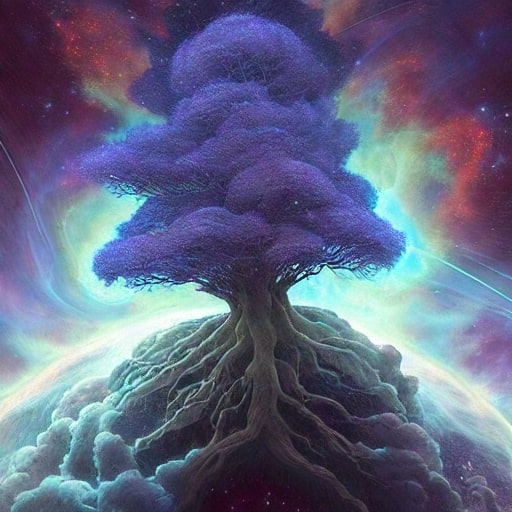}
      &
      \includegraphics[width=0.3\textwidth]{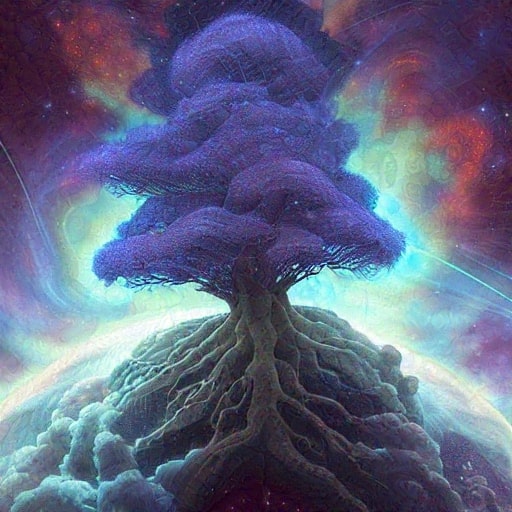}
      & 
      \includegraphics[width=0.3\textwidth]{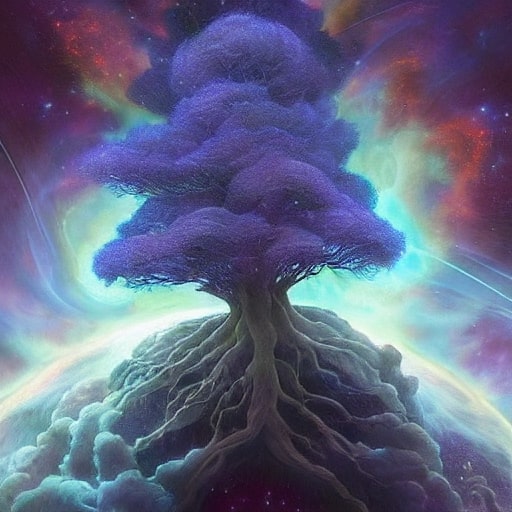}
      & 
      \includegraphics[width=0.3\textwidth]{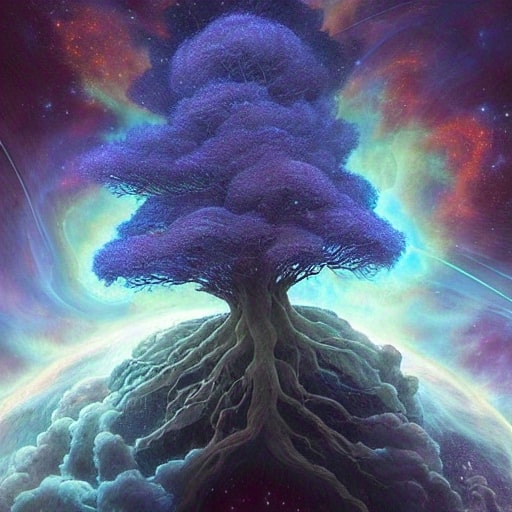}
      & 
      \includegraphics[width=0.3\textwidth]{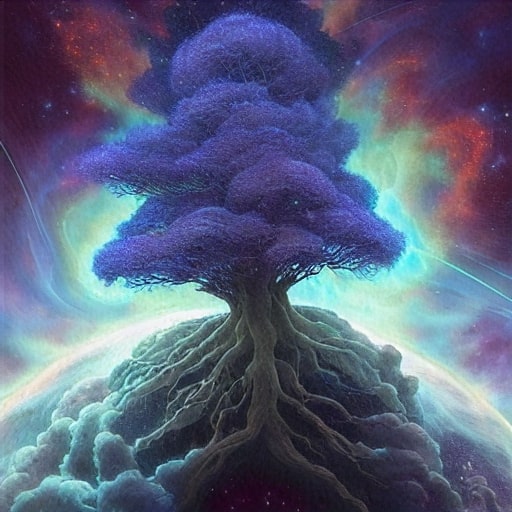}
      & 
      \includegraphics[width=0.3\textwidth]{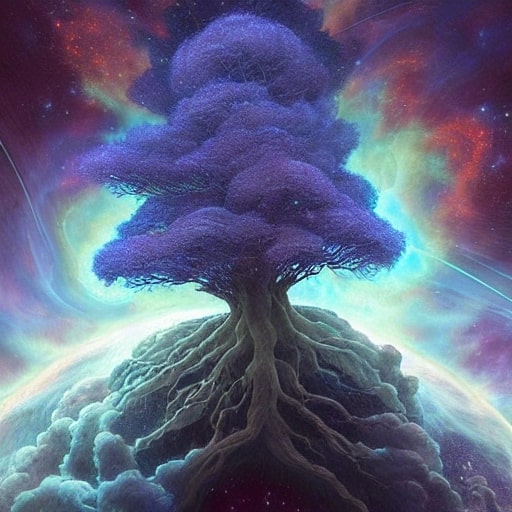}
      \\
      &
      \includegraphics[width=0.3\textwidth]{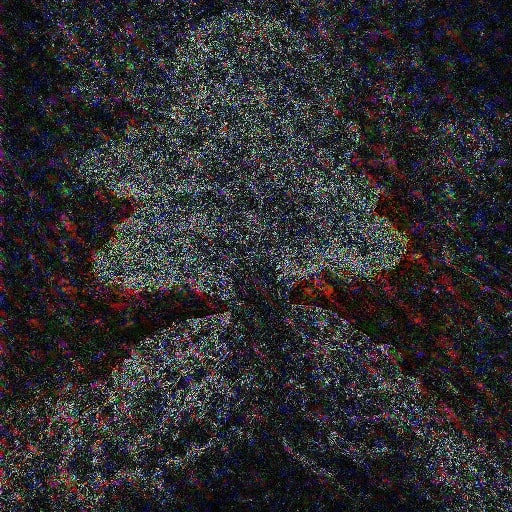}
      &
      \includegraphics[width=0.3\textwidth]{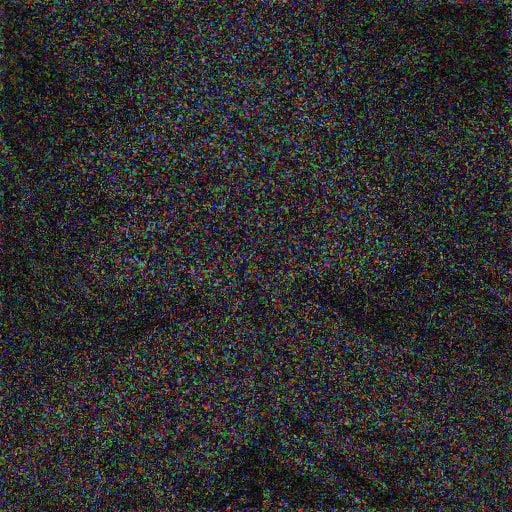}
      & 
      \includegraphics[width=0.3\textwidth]{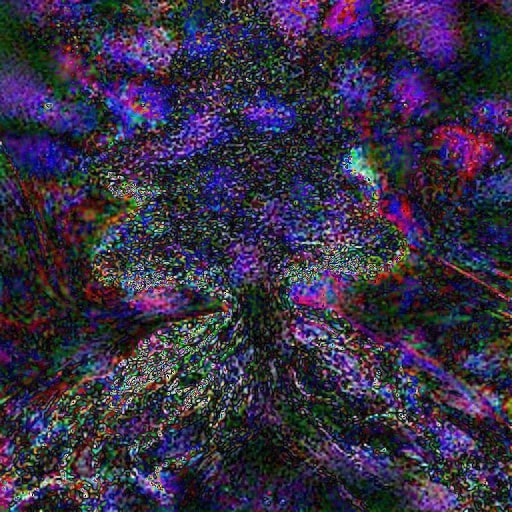}
      & 
      \includegraphics[width=0.3\textwidth]{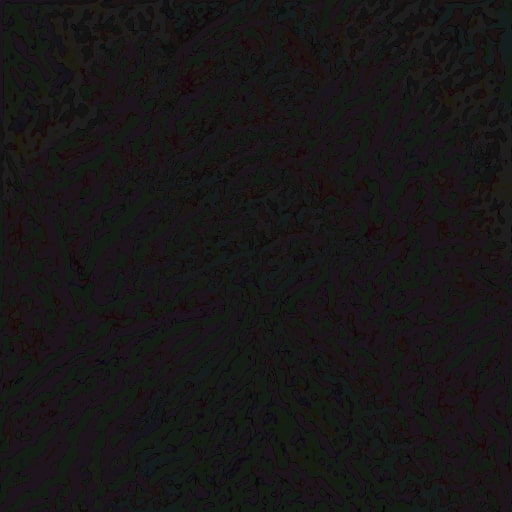}
      & 
      \includegraphics[width=0.3\textwidth]{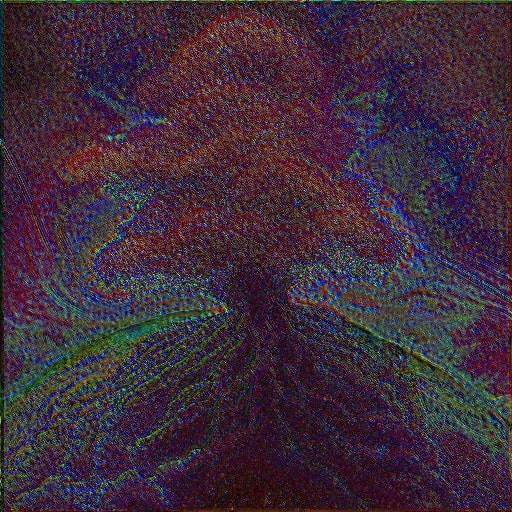}
      & 
      \includegraphics[width=0.3\textwidth]{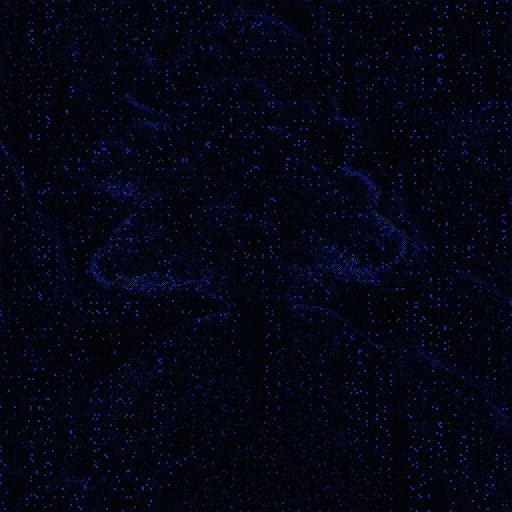}
      \\
      \includegraphics[width=0.3\textwidth]{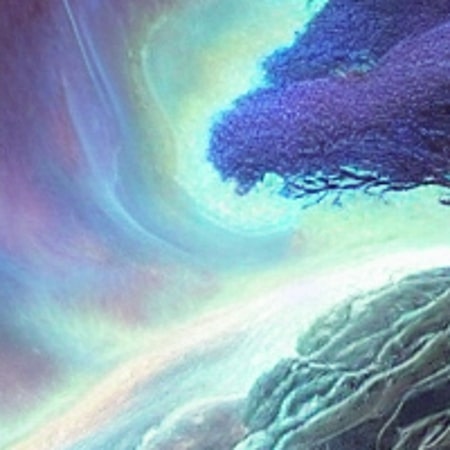}
      &
      \includegraphics[width=0.3\textwidth]{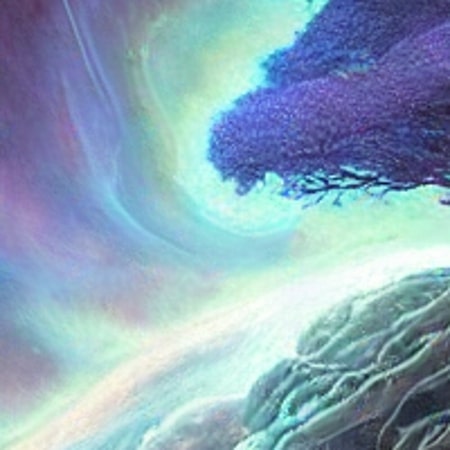}
      &
      \includegraphics[width=0.3\textwidth]{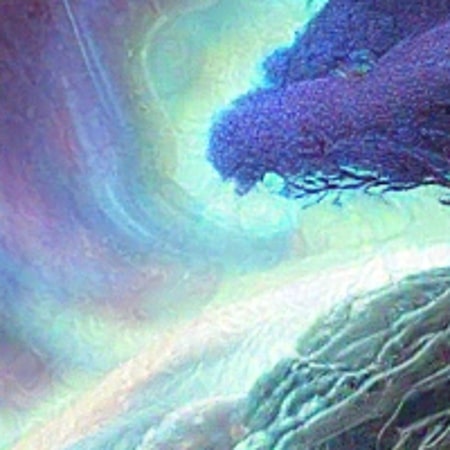}
      & 
      \includegraphics[width=0.3\textwidth]{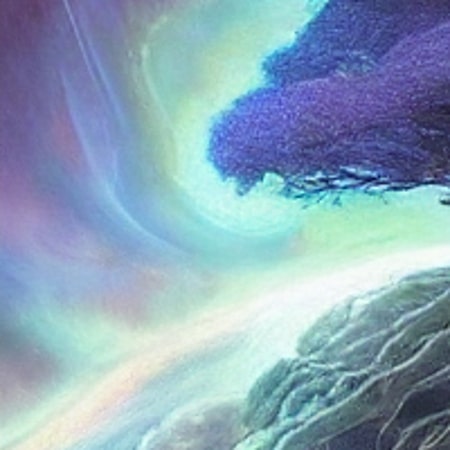}
      & 
      \includegraphics[width=0.3\textwidth]{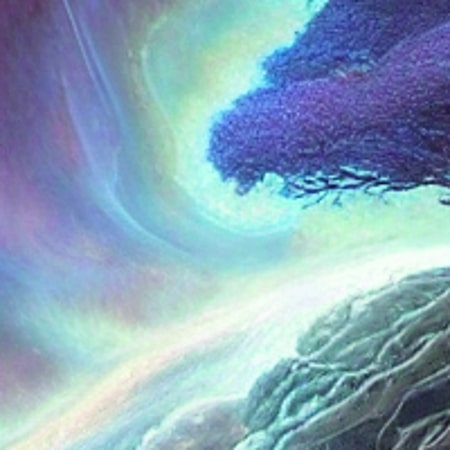}
      & 
      \includegraphics[width=0.3\textwidth]{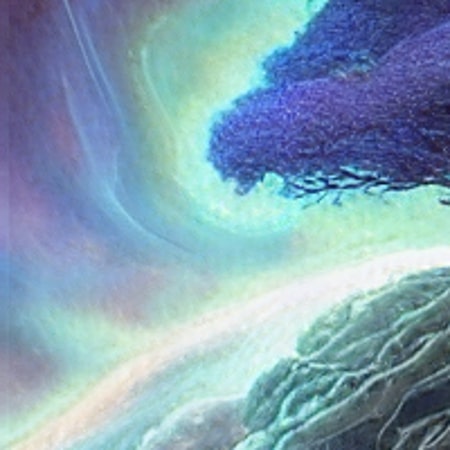}
      & 
      \includegraphics[width=0.3\textwidth]{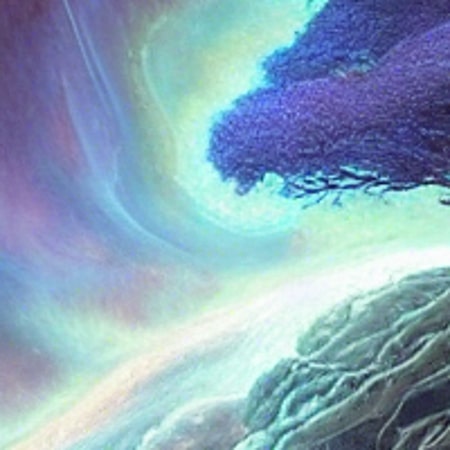}
      \\
      \bottomrule
      \end{tabular}}
      \captionof{figure}{Qualitative comparison of watermarked images using LaWa and other methods for 32-bit watermarks.
      } \label{fig:supp_fig4}
    \vspace{-15pt}
\end{table*}
\textbf{Inpainting.} In this task, some parts of an image is covered by a mask and the diffusion process inpaints the hidden parts behind the mask. For this task, we follow the protocol of LaMa~\cite{suvorov2022resolution} for 1k images selected from MS-COCO. We use "thick" setting of the protocol and create masks that randomly cover 1\% to 50\% of the image.

\textbf{Super-resolution.} In this task, diffusion process is used to increase the resolution of an existing image. We resize 1k images from MS-COCO to $128\times128$ resolution using bicubic interpolation. Diffusion process scales the resolution of the images back to $512\times512$.

\textbf{Text-to-image.} In this task, a text prompt guides the diffusion process. We use 500 prompts randomly selected from MS-COCO and 500 prompts randomly selected from MagicPrompt-SD~\cite{GustavostaPrompts}. Using 10 random messages, we generate a total of 10k images for this task.

\begin{table*}[tb]
     \centering
    \resizebox{0.95\textwidth}{!}{ 
    \begin{tabular}{ c c c  c  c c c}
    \toprule
    Original & LaWa & SSL (48)~\cite{fernandez2022watermarking} & FNNS (48)~\cite{kishore2021fixed} & RoSteALS (48)~\cite{bui2023rosteals} & RivaGan (32)~\cite{zhang2019robust} \\
    \midrule
      \includegraphics[width=0.3\textwidth]{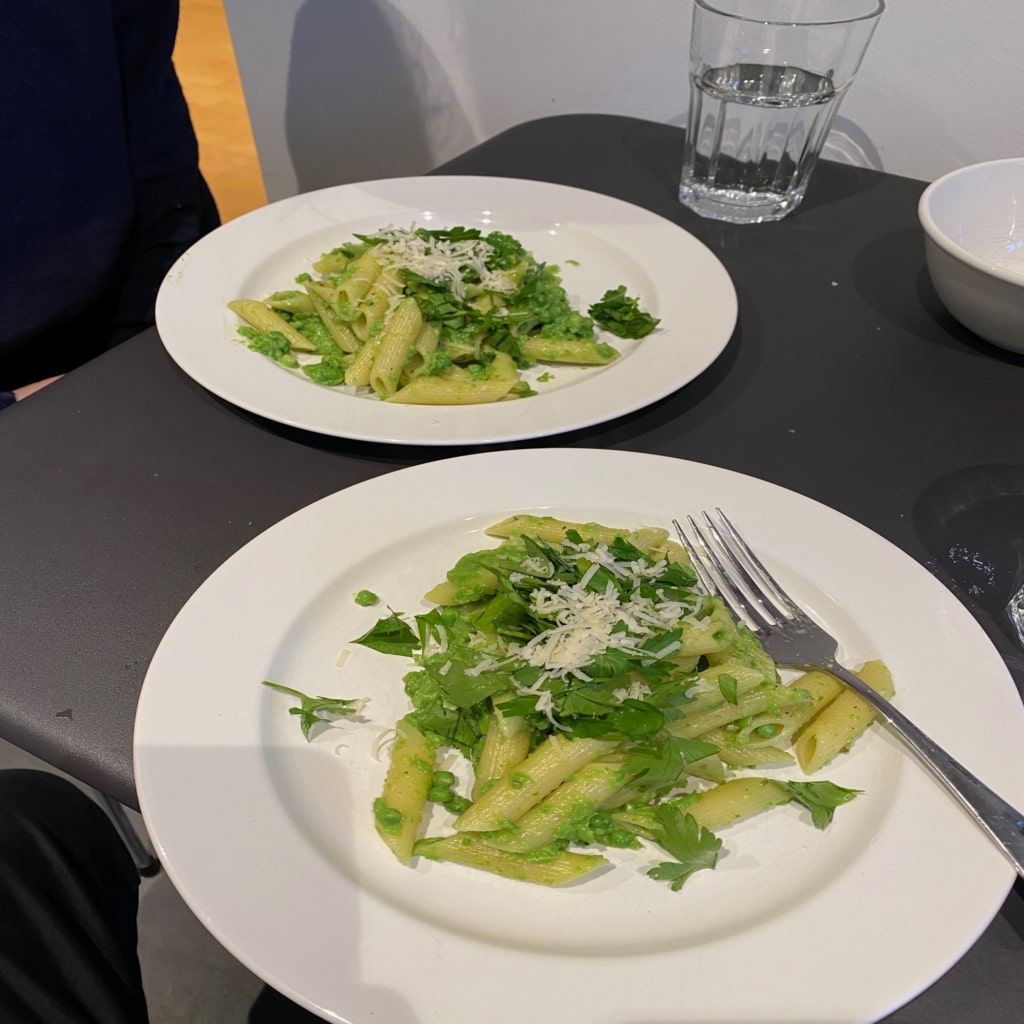}
      &
      \includegraphics[width=0.3\textwidth]{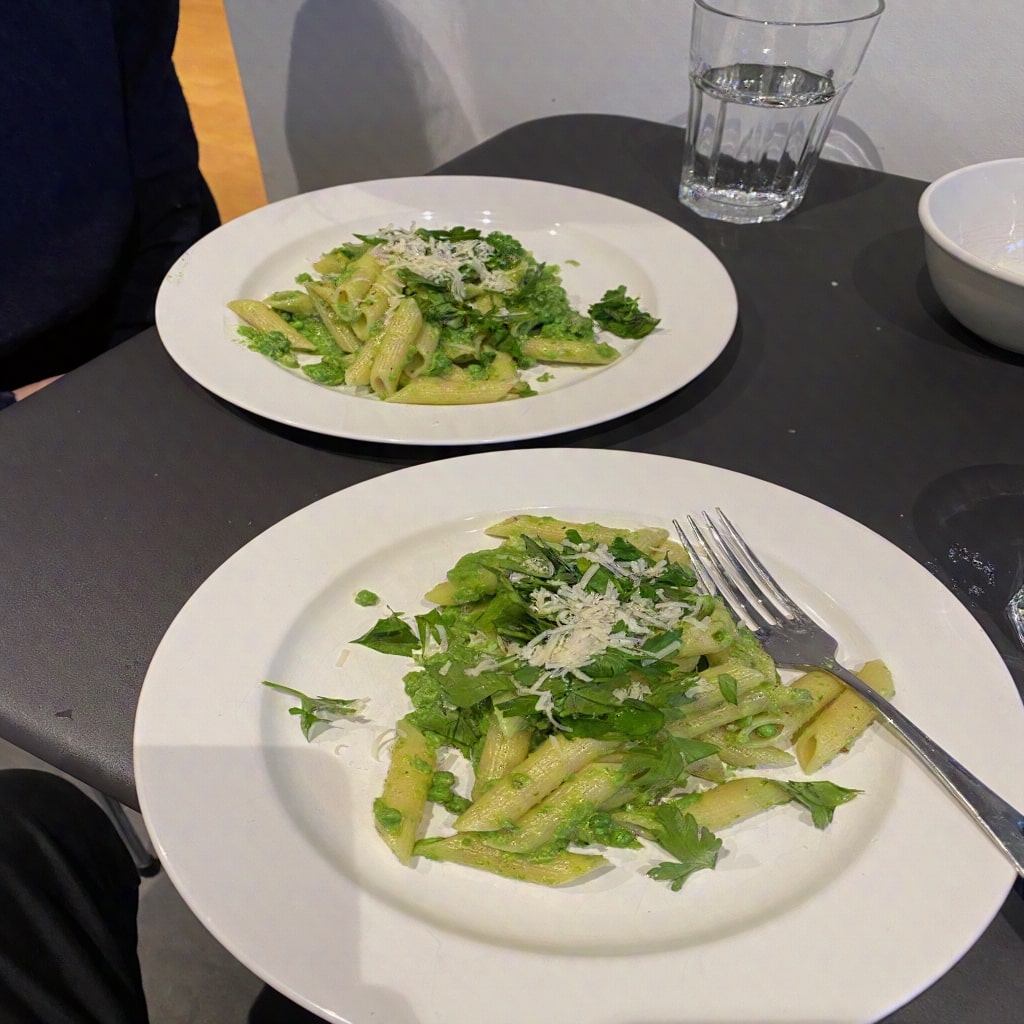}
      &
      \includegraphics[width=0.3\textwidth]{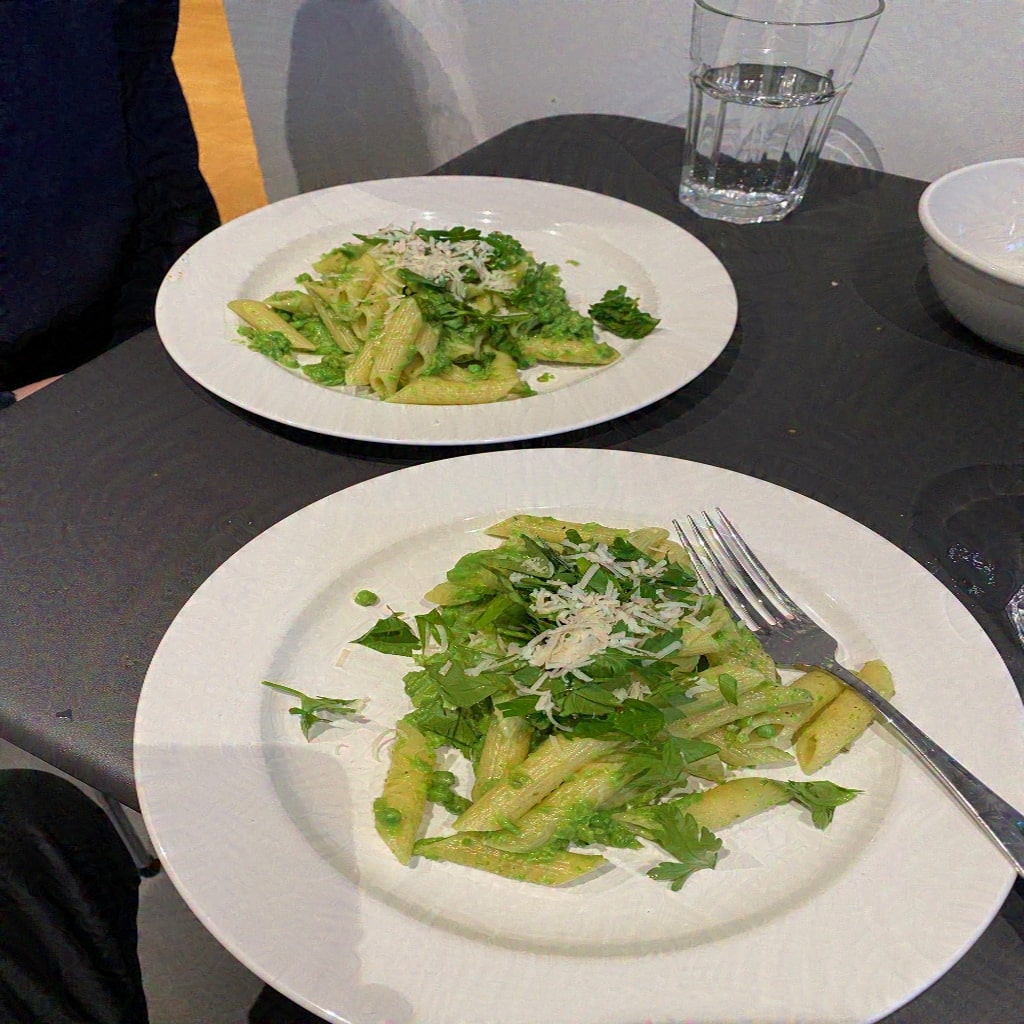}
      &
      \includegraphics[width=0.3\textwidth]{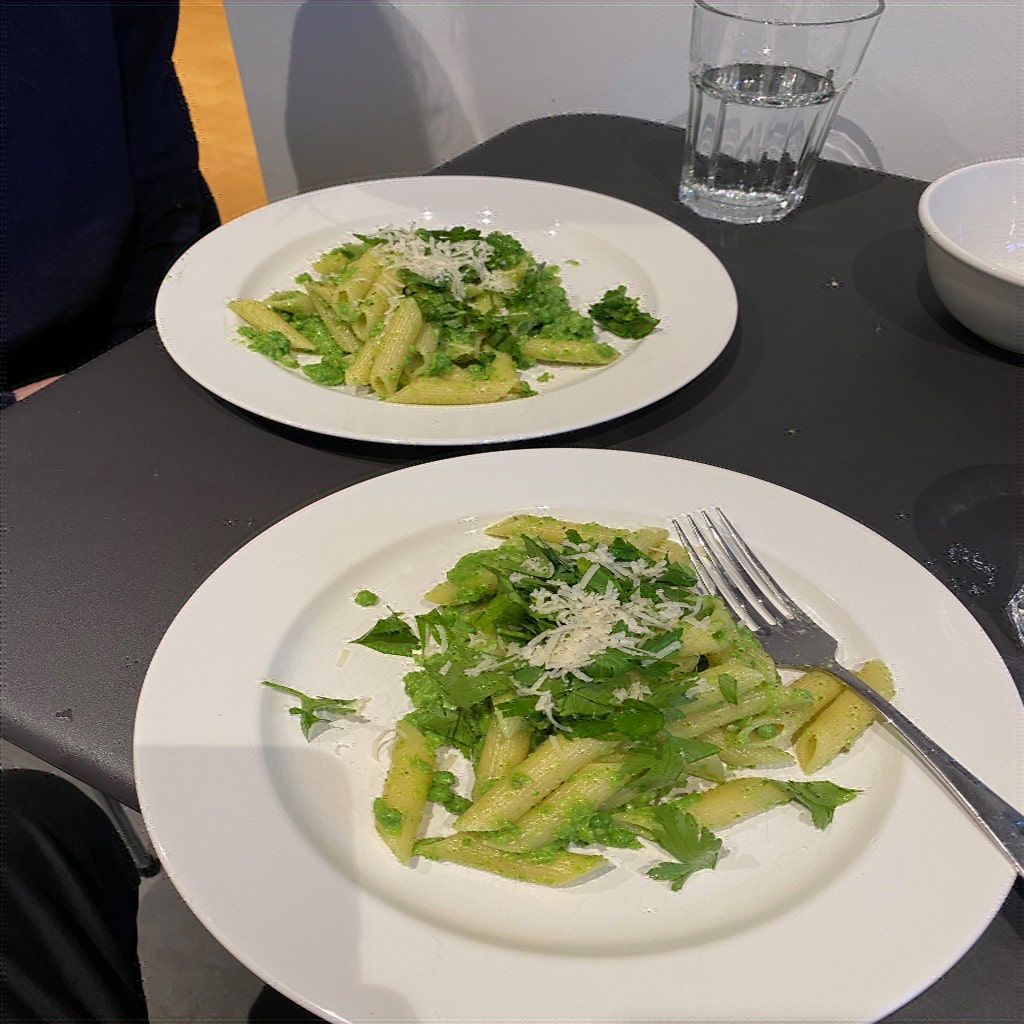}
      & 
      \includegraphics[width=0.3\textwidth]{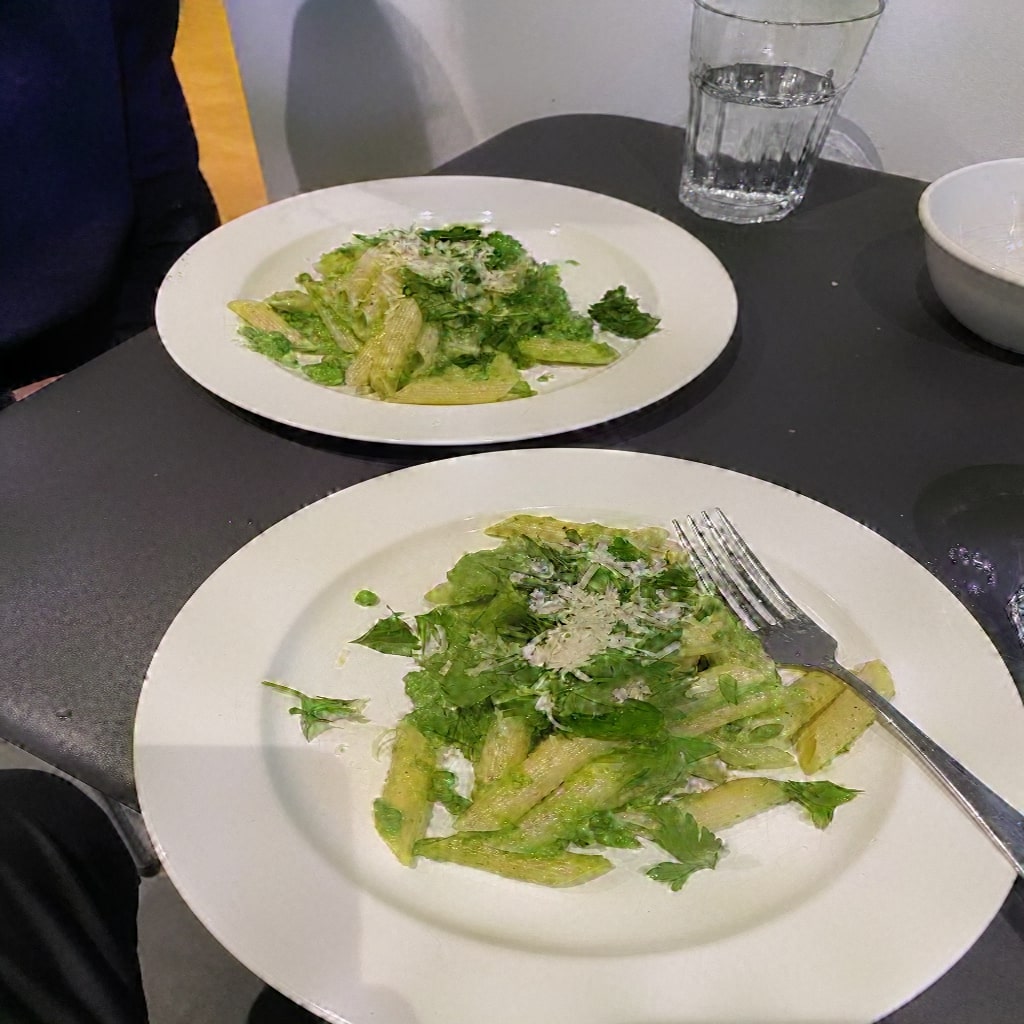}
      & 
      \includegraphics[width=0.3\textwidth]{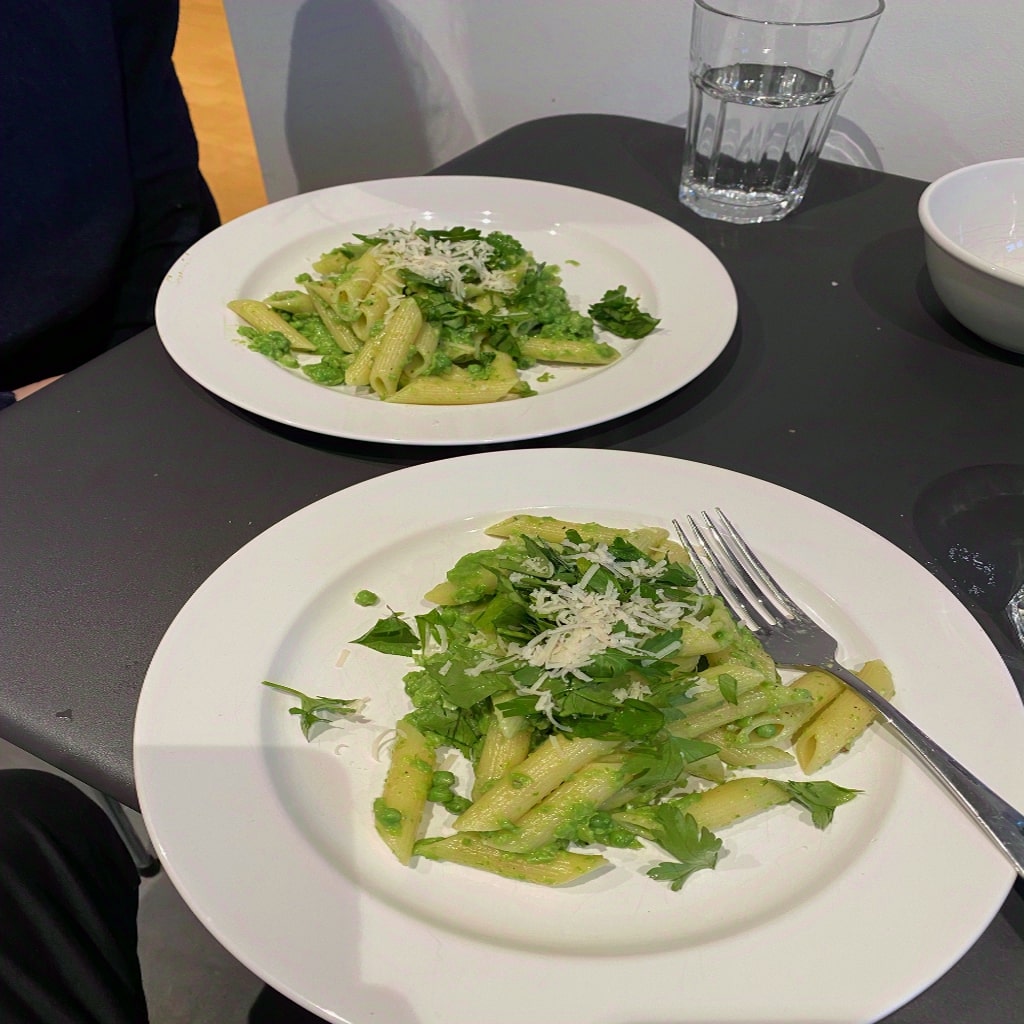}
      \\
      &
      \includegraphics[width=0.3\textwidth]{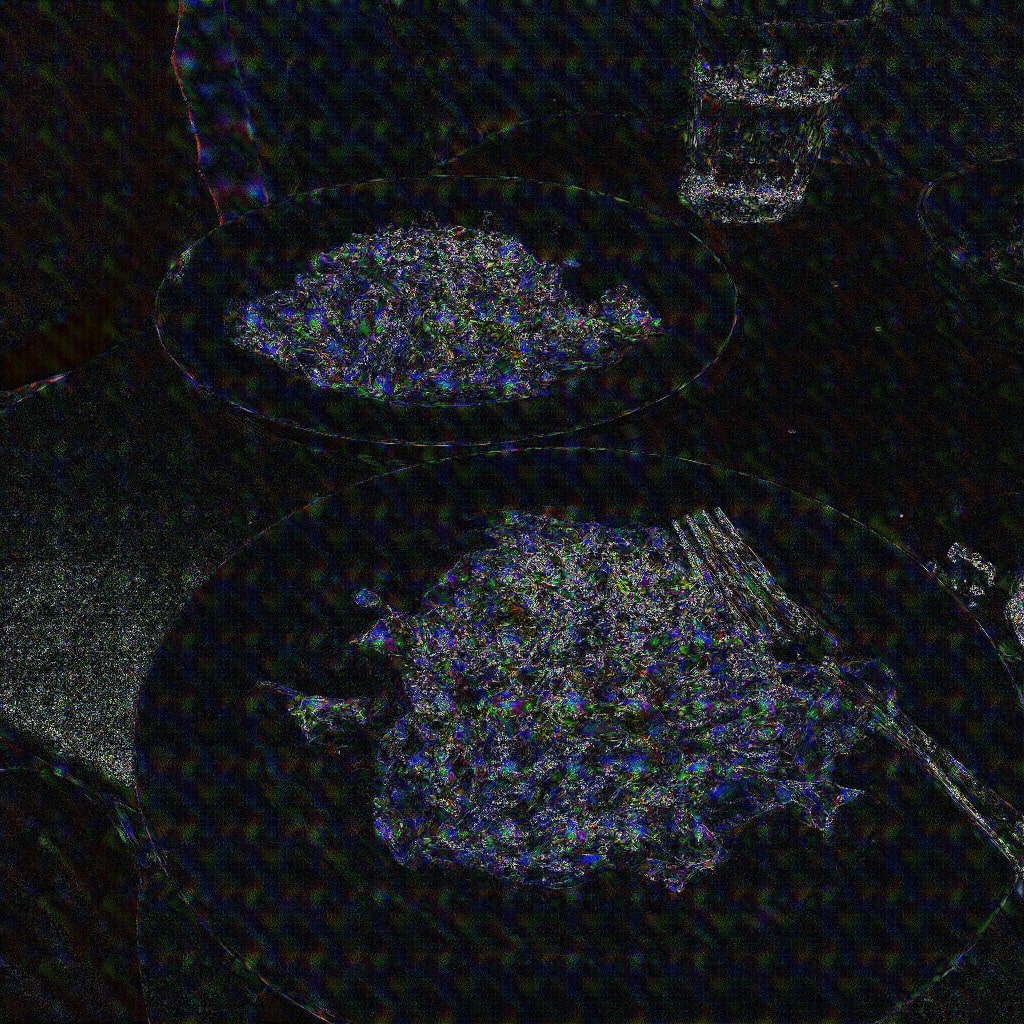}
      &
      \includegraphics[width=0.3\textwidth]{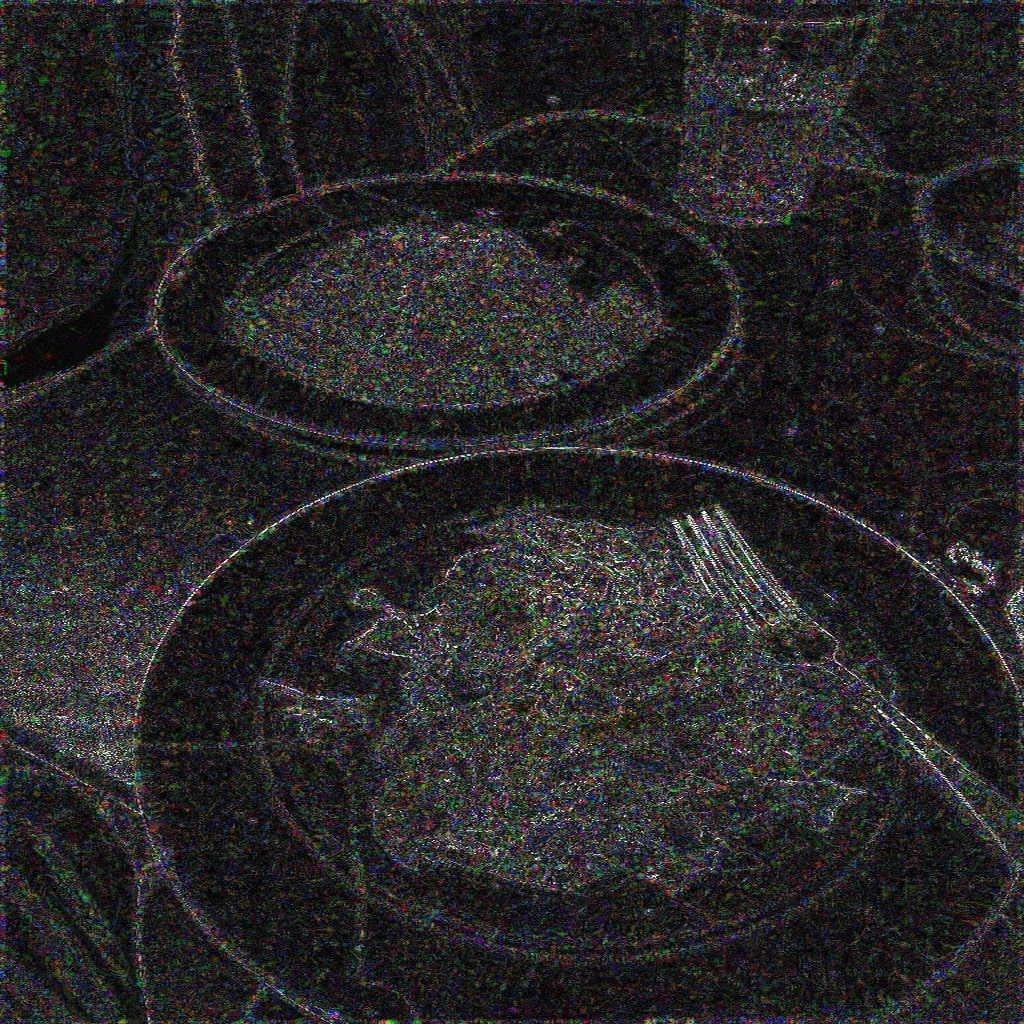}
      &
      \includegraphics[width=0.3\textwidth]{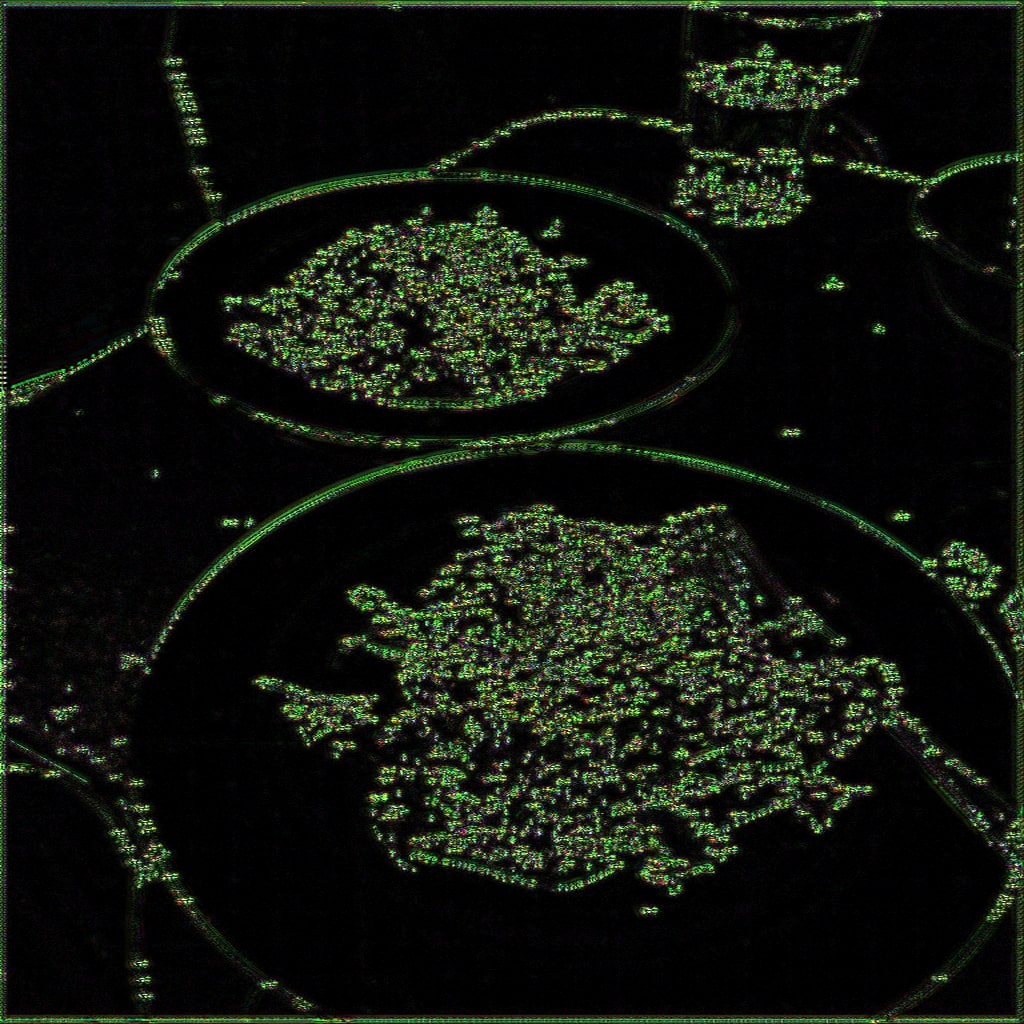}
      & 
      \includegraphics[width=0.3\textwidth]{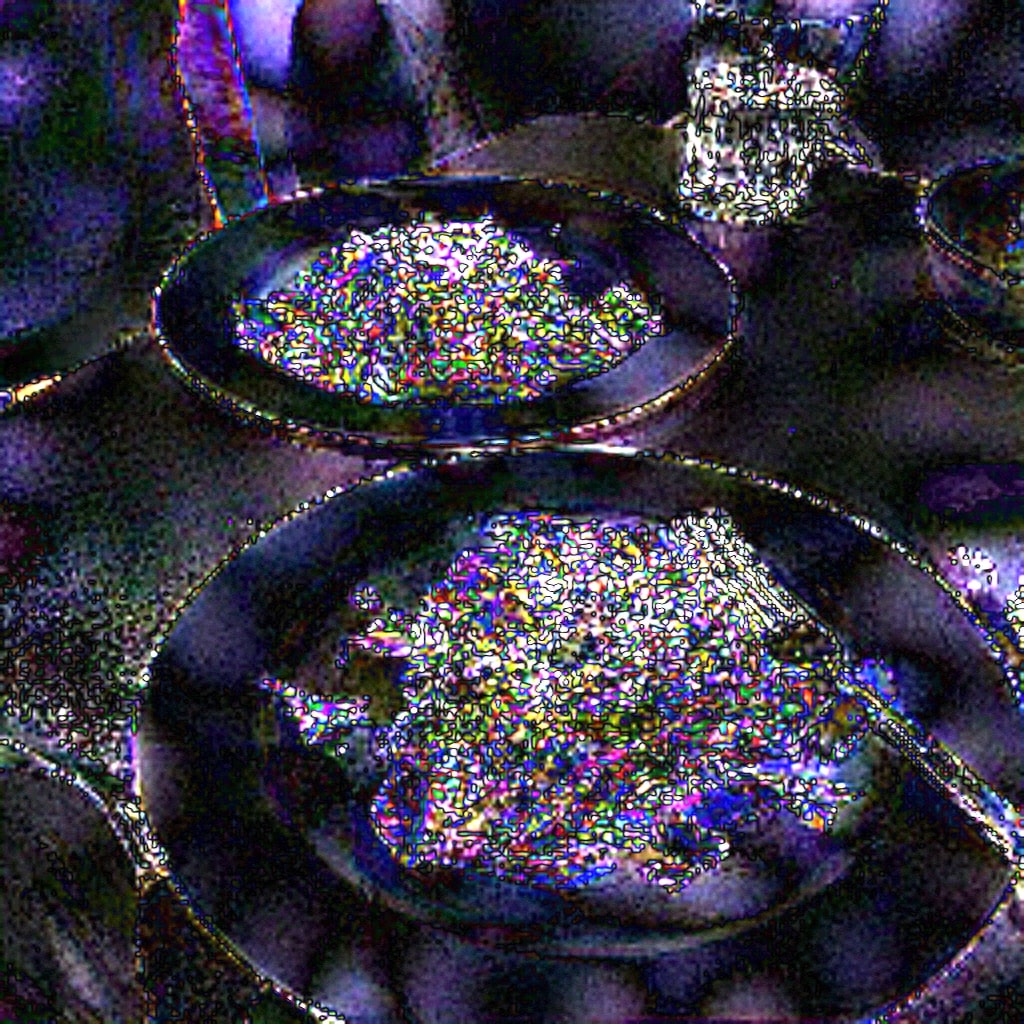}
      & 
      \includegraphics[width=0.3\textwidth]{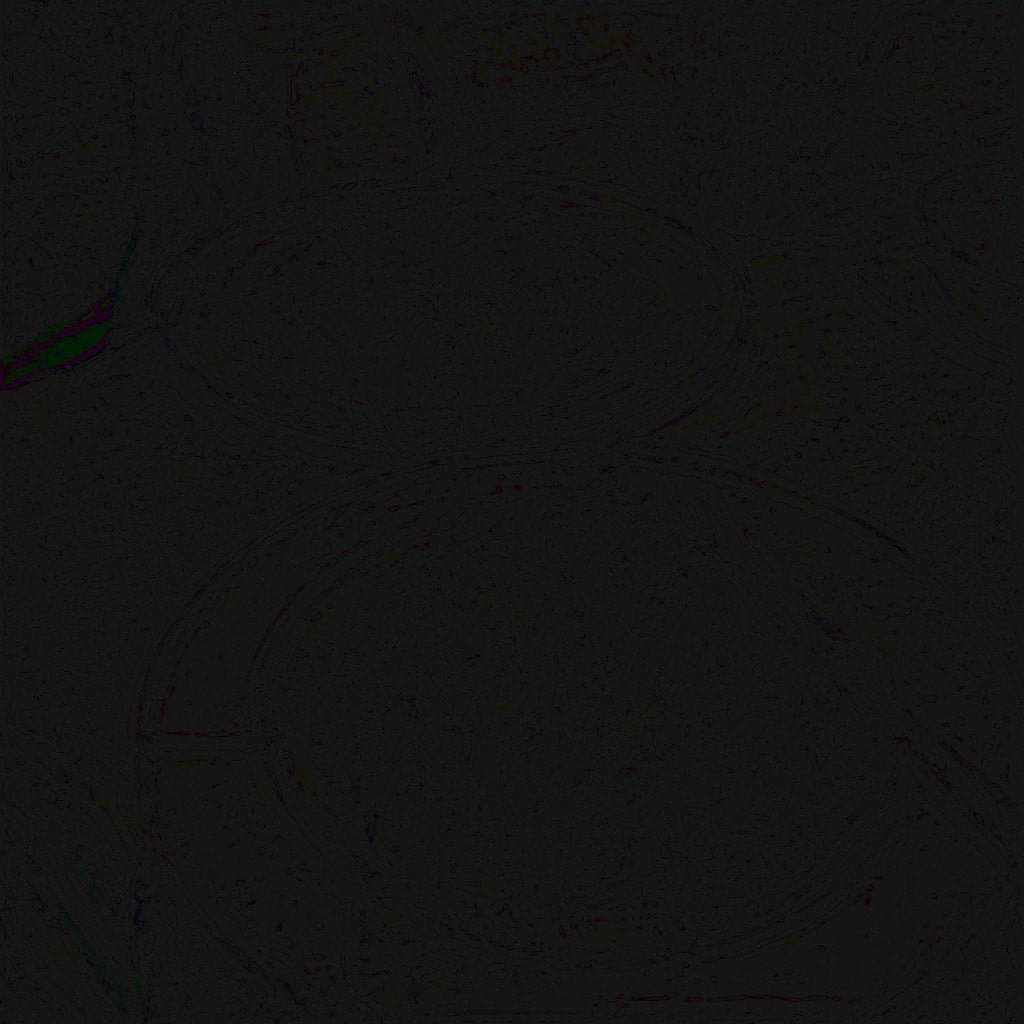}
      \\
      \includegraphics[width=0.3\textwidth]{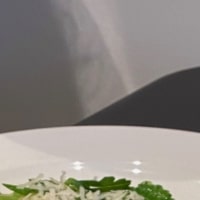}
      &
      \includegraphics[width=0.3\textwidth]{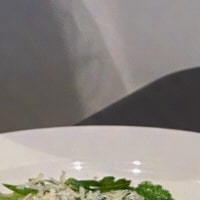}
      &
      \includegraphics[width=0.3\textwidth]{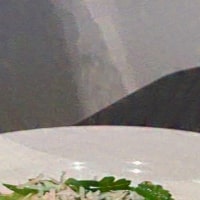}
      & 
      \includegraphics[width=0.3\textwidth]{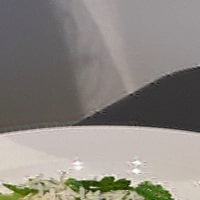}
      & 
      \includegraphics[width=0.3\textwidth]{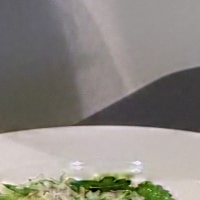}
      & 
      \includegraphics[width=0.3\textwidth]{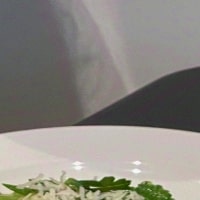}
      \\
      \includegraphics[width=0.3\textwidth]{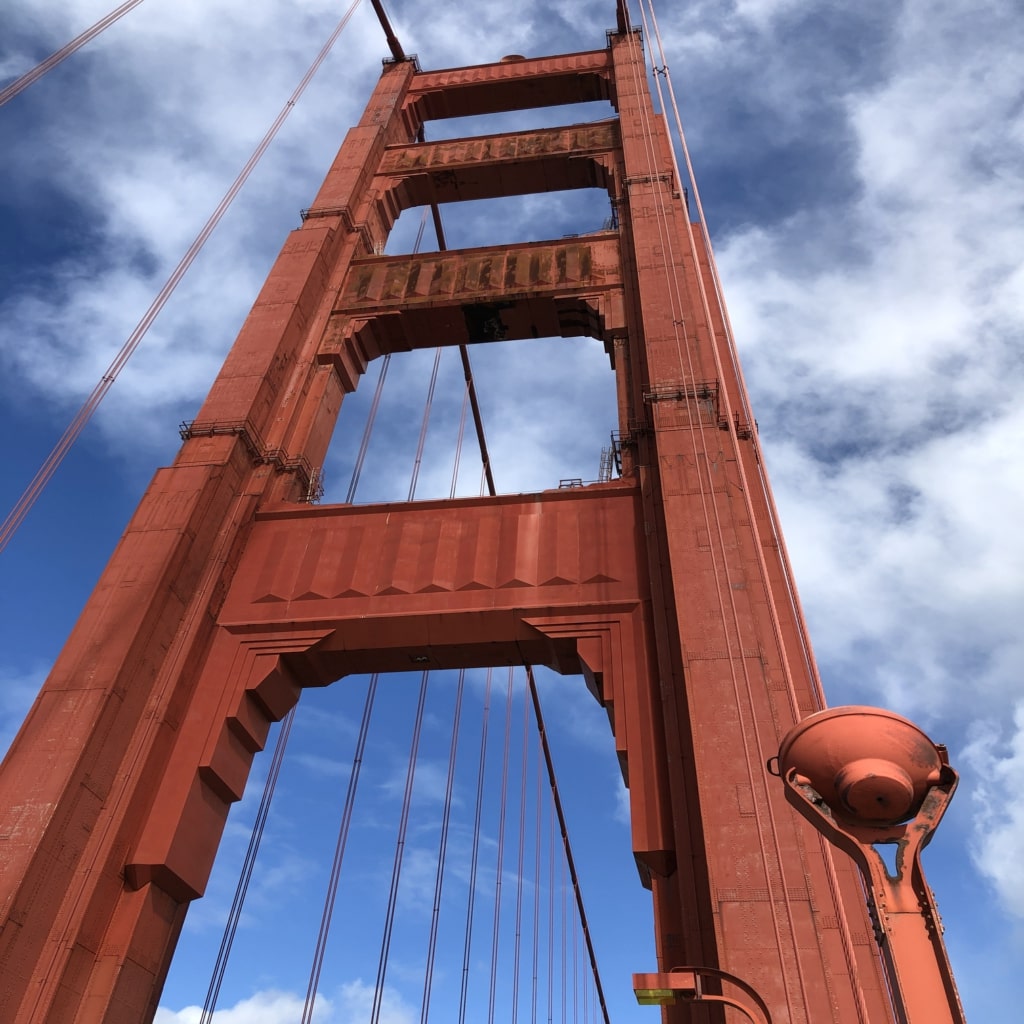}
      &
      \includegraphics[width=0.3\textwidth]{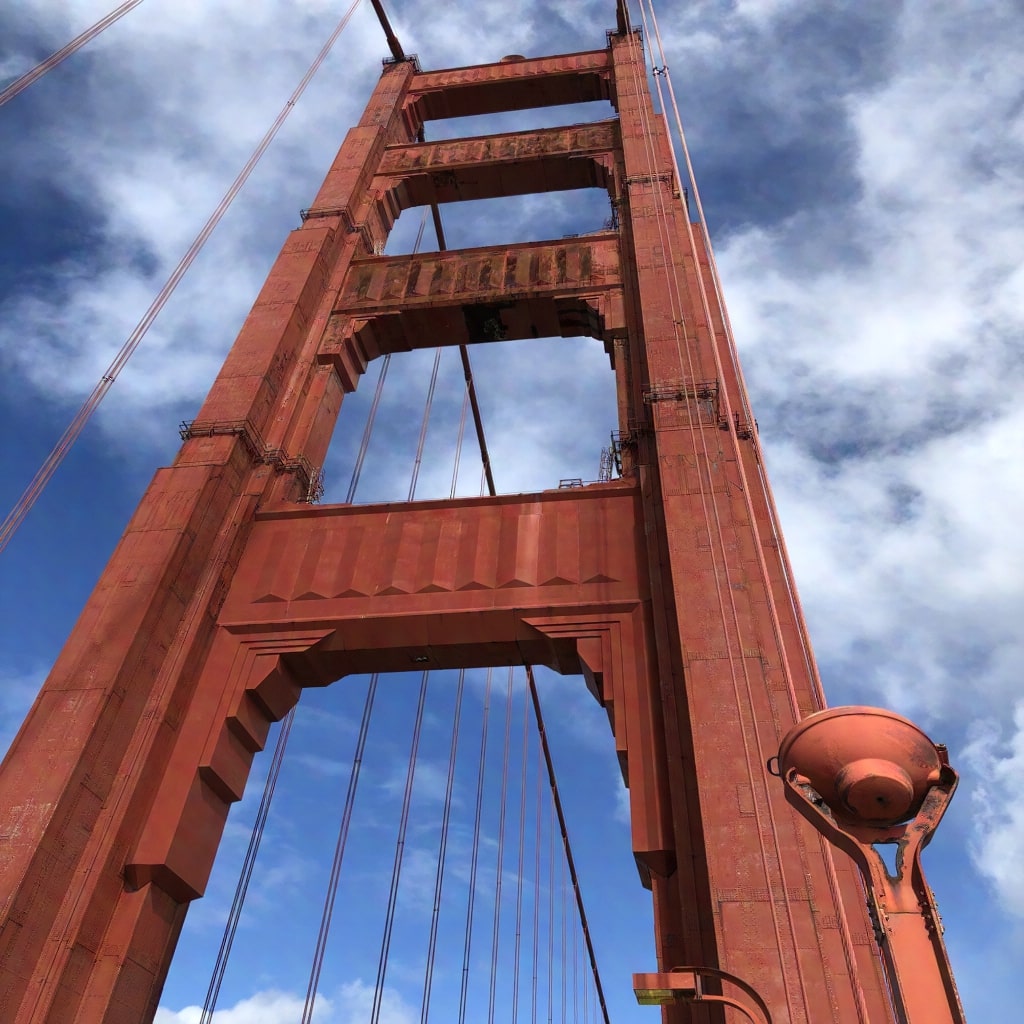}
      &
      \includegraphics[width=0.3\textwidth]{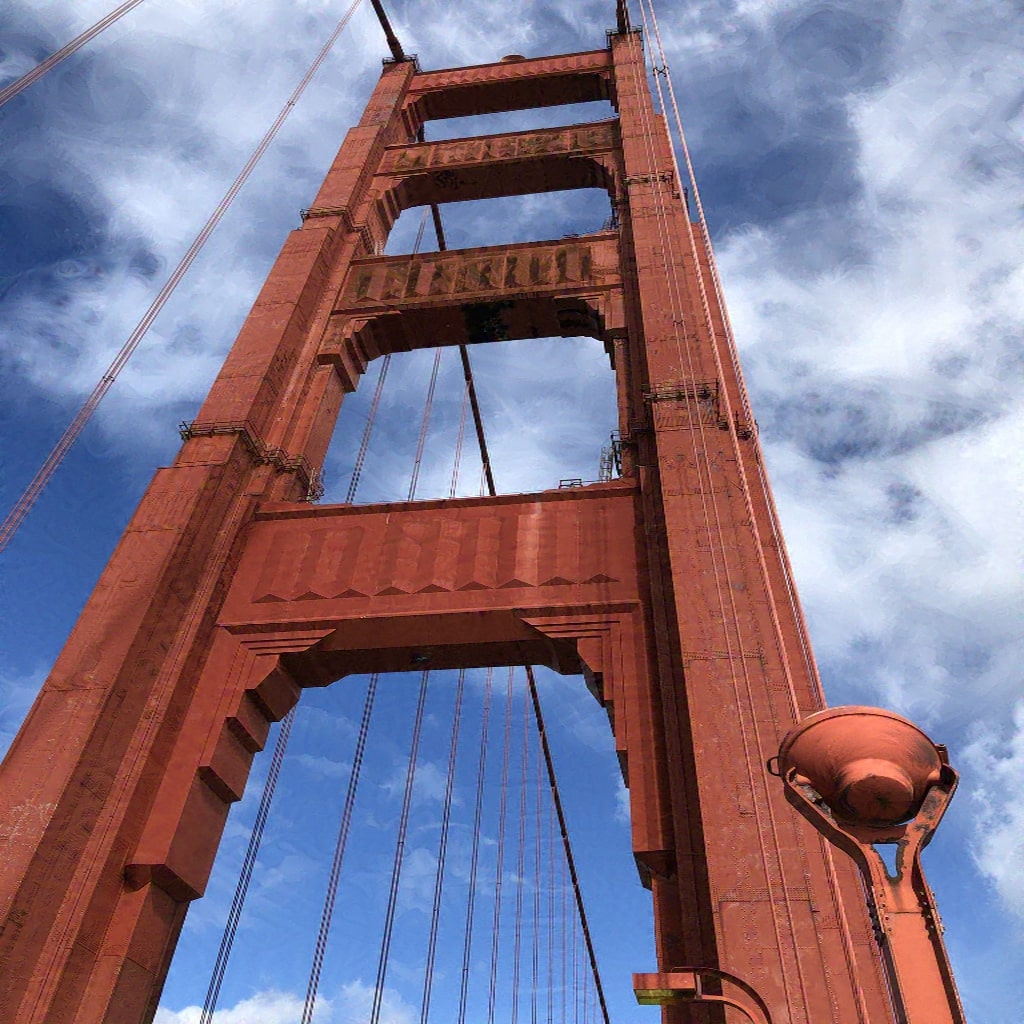}
      &
      \includegraphics[width=0.3\textwidth]{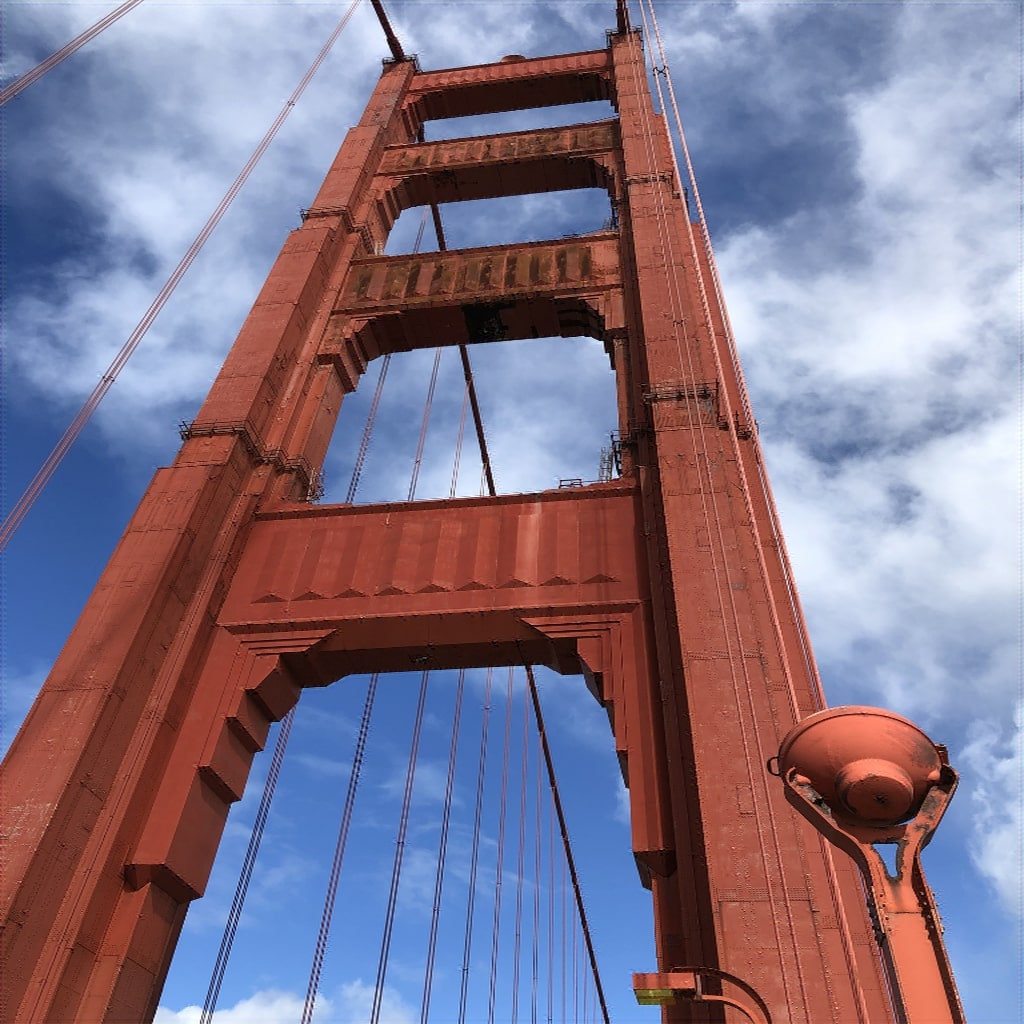}
      & 
      \includegraphics[width=0.3\textwidth]{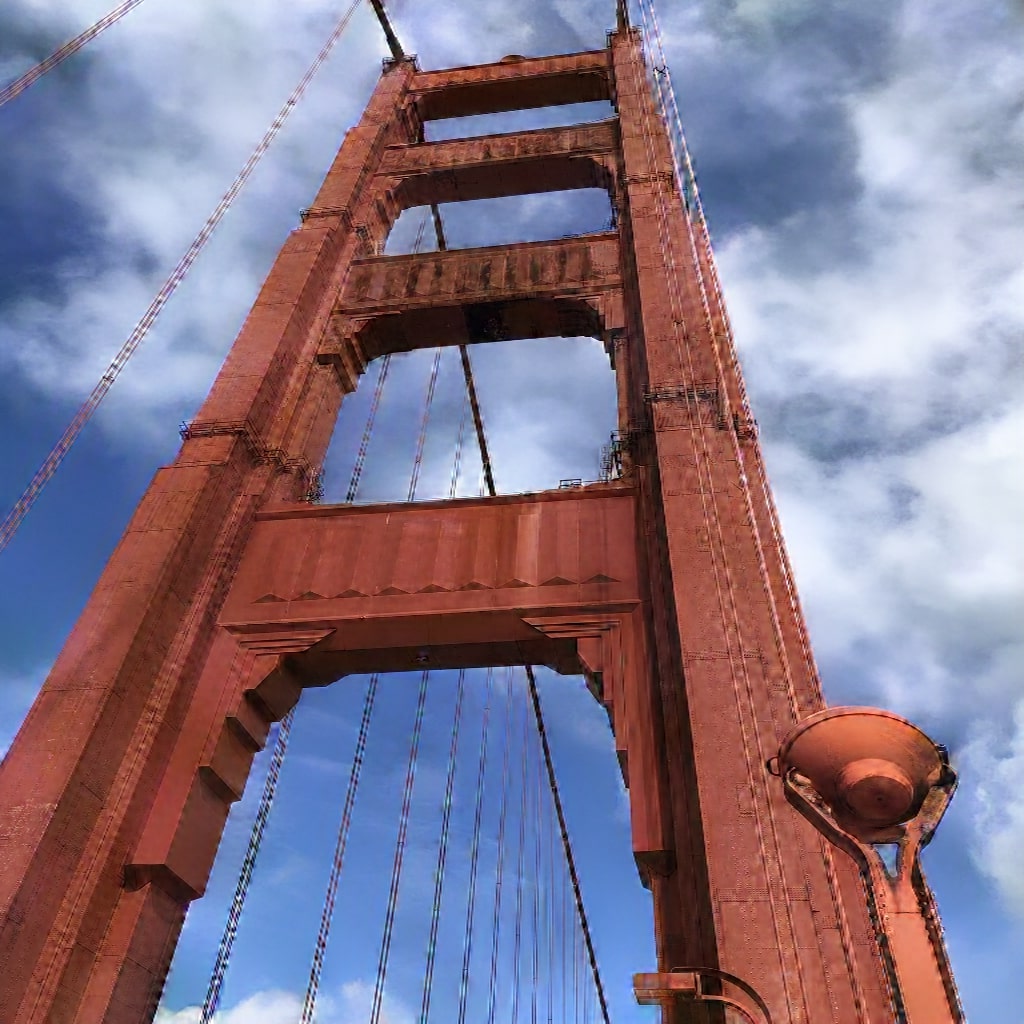}
      & 
      \includegraphics[width=0.3\textwidth]{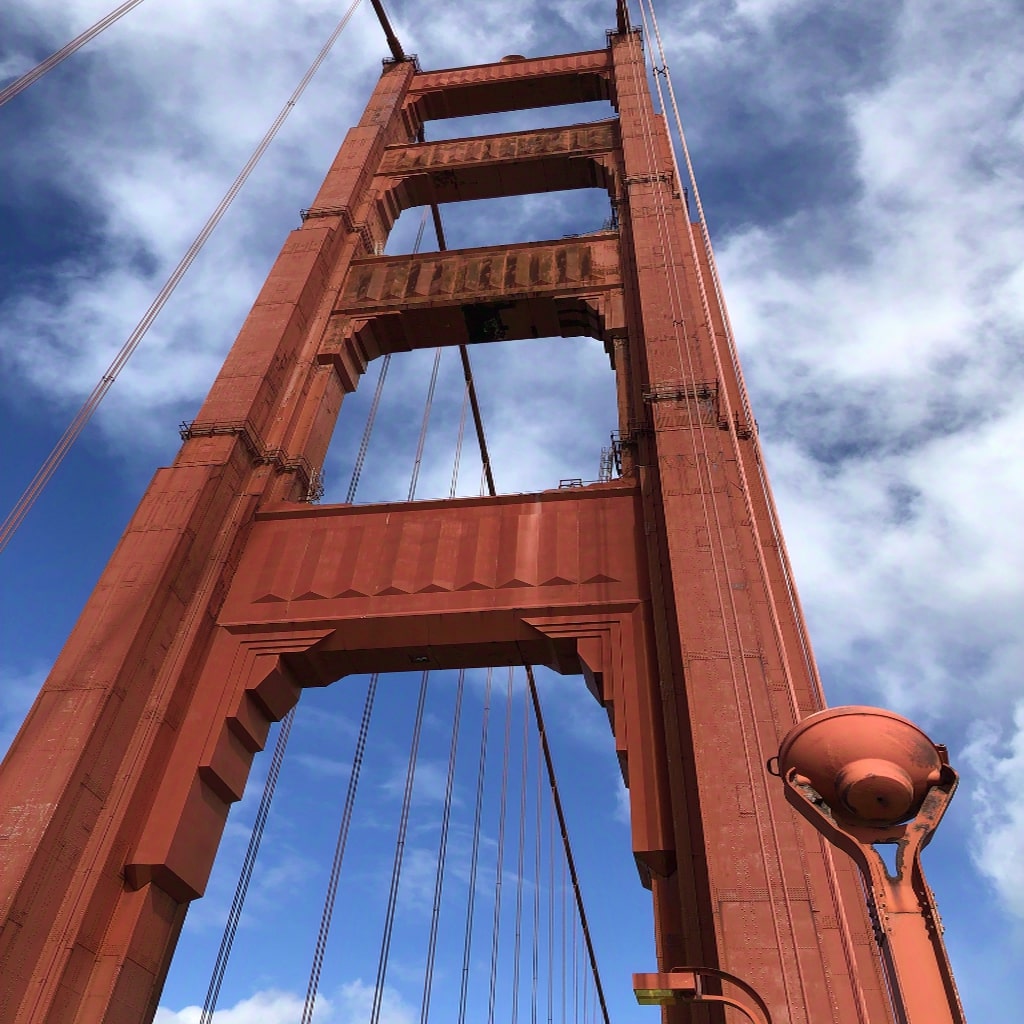}
      \\
      &
      \includegraphics[width=0.3\textwidth]{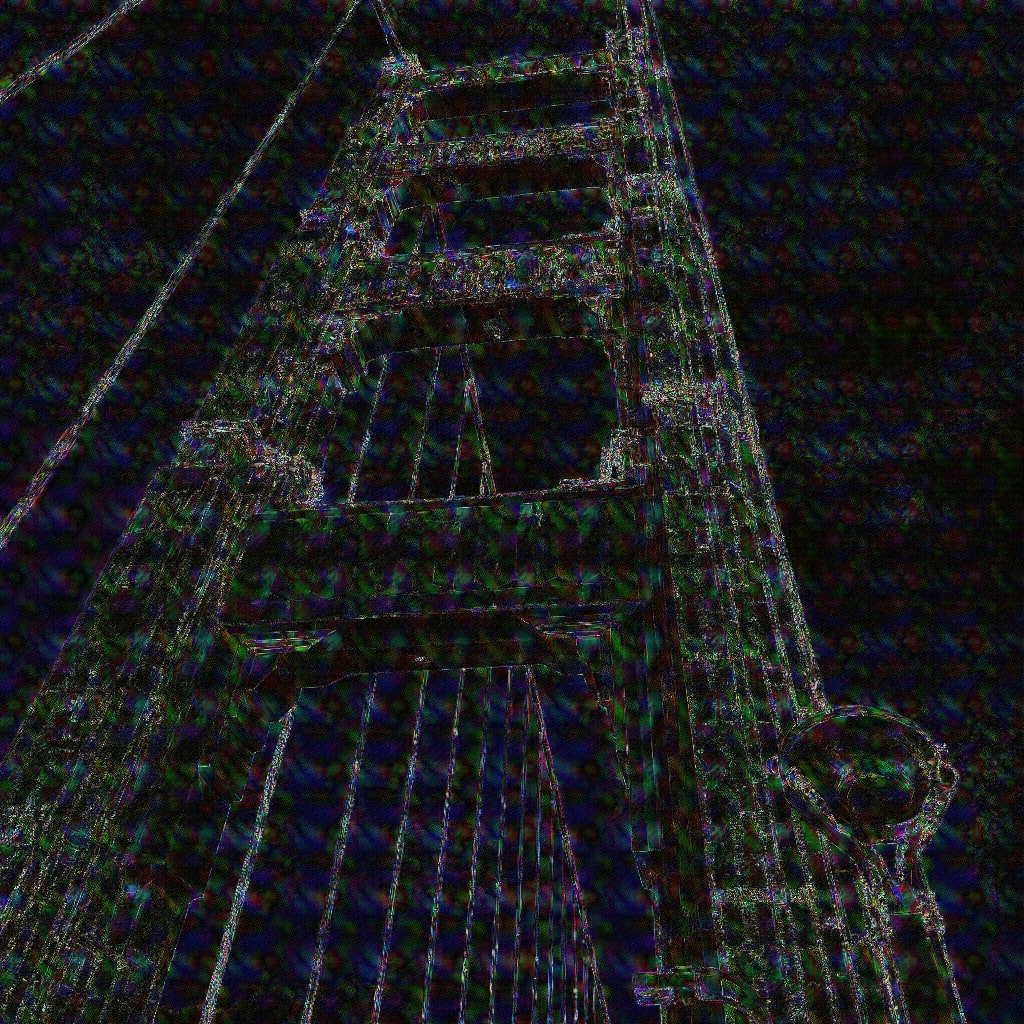}
      &
      \includegraphics[width=0.3\textwidth]{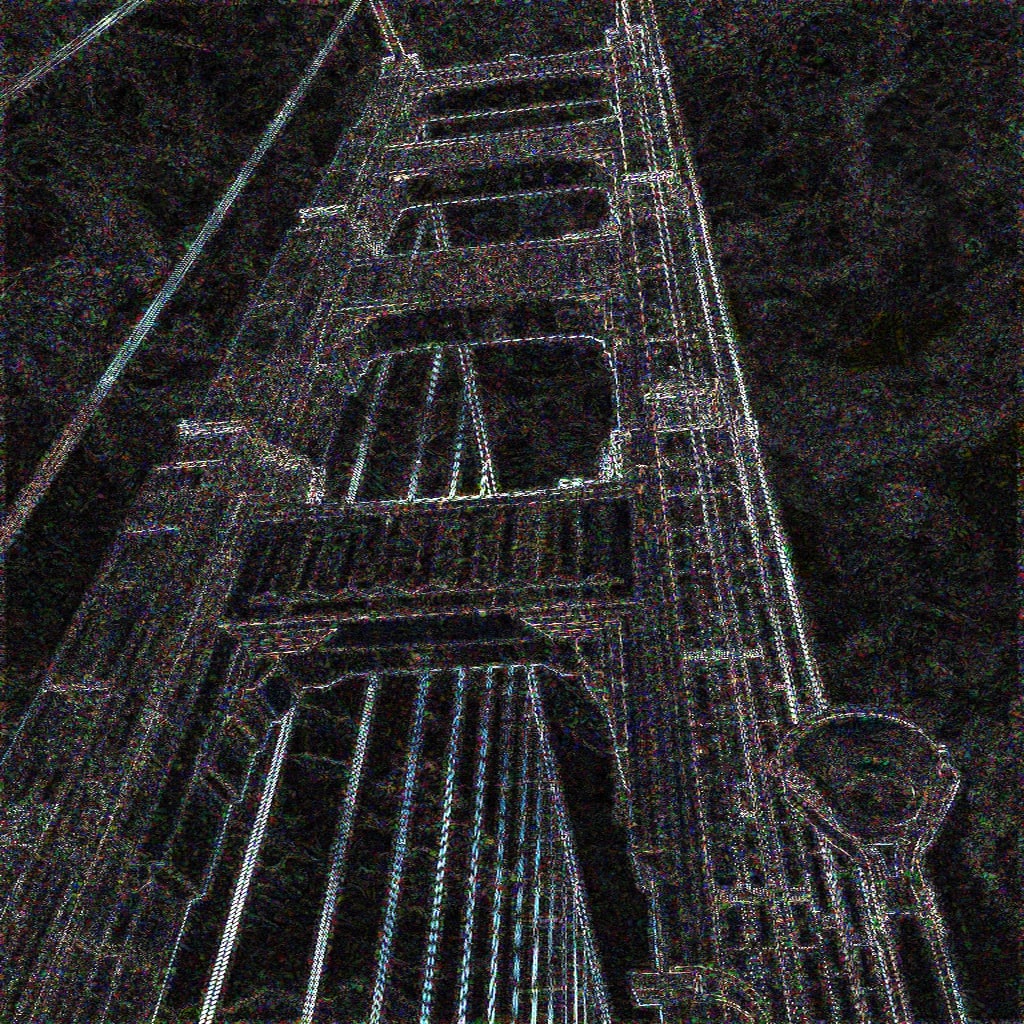}
      &
      \includegraphics[width=0.3\textwidth]{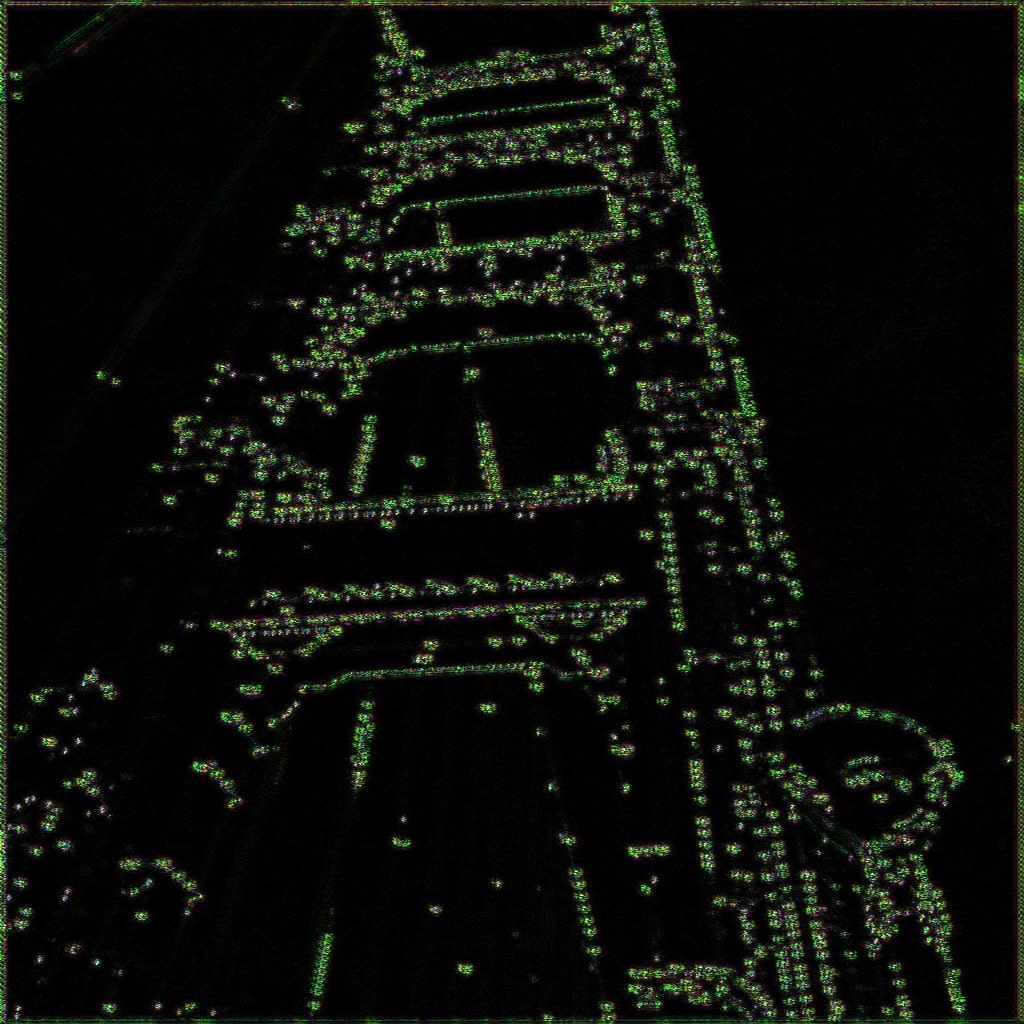}
      & 
      \includegraphics[width=0.3\textwidth]{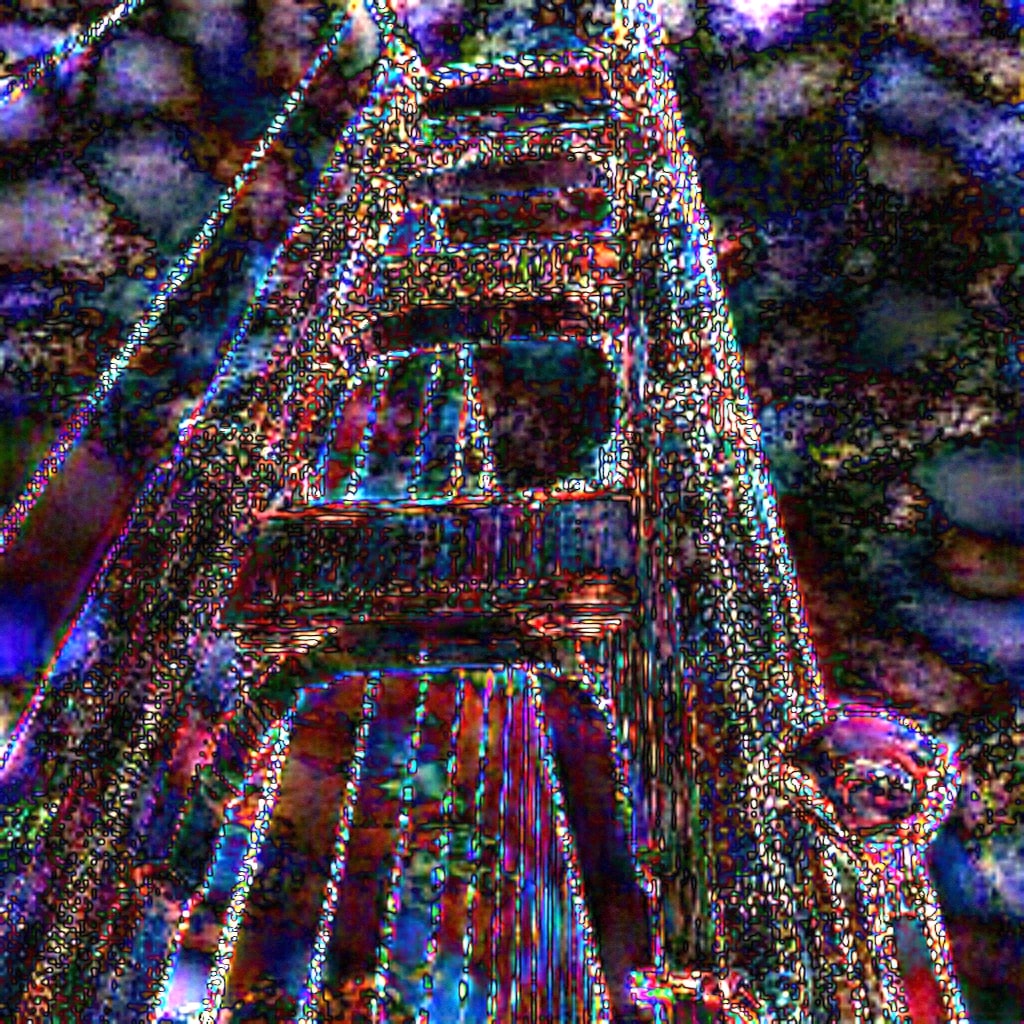}
      & 
      \includegraphics[width=0.3\textwidth]{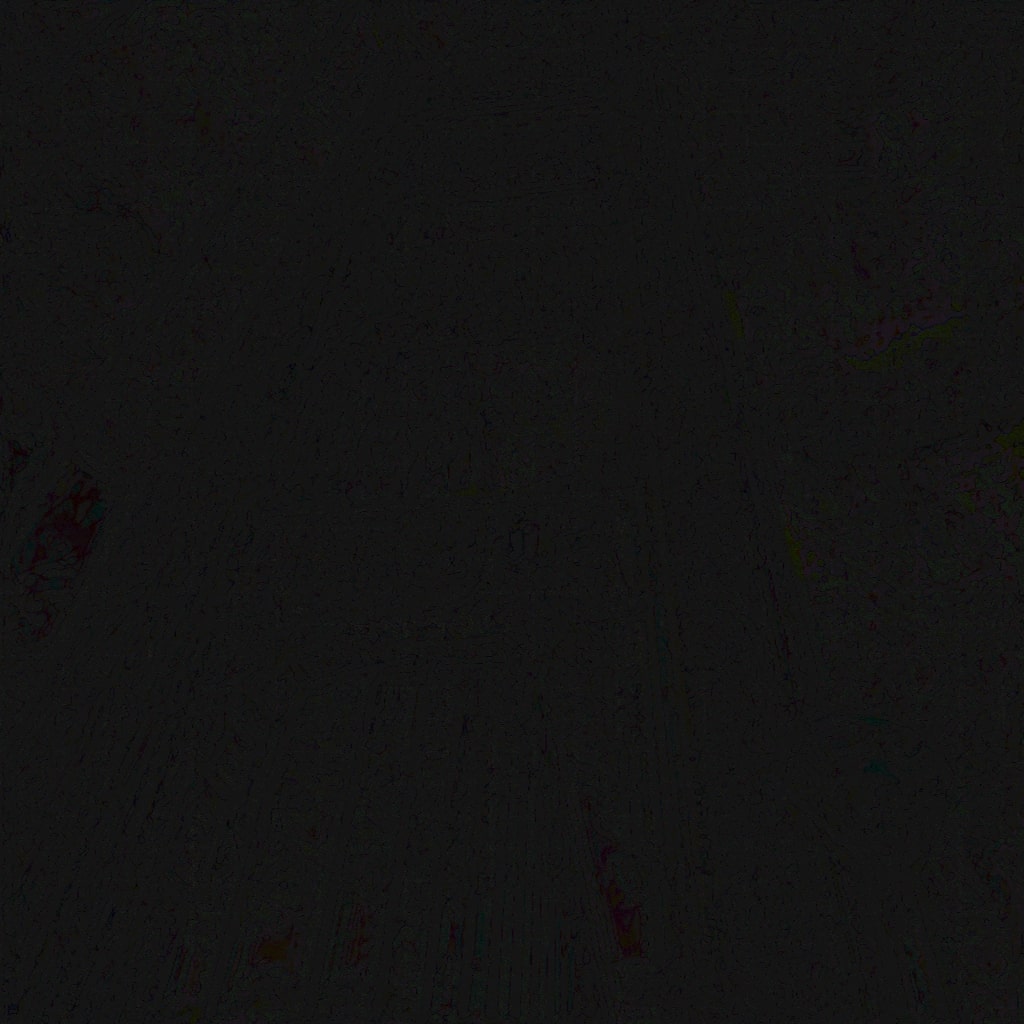}
      \\
      \includegraphics[width=0.3\textwidth]{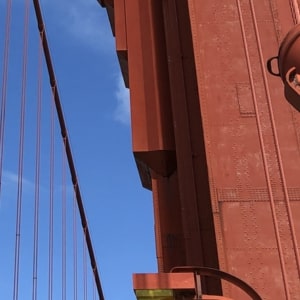}
      &
      \includegraphics[width=0.3\textwidth]{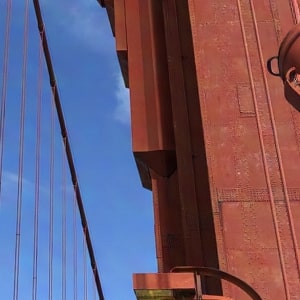}
      &
      \includegraphics[width=0.3\textwidth]{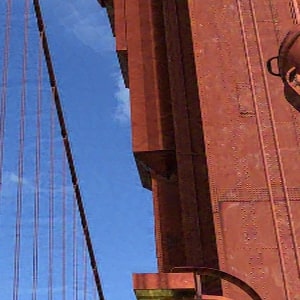}
      & 
      \includegraphics[width=0.3\textwidth]{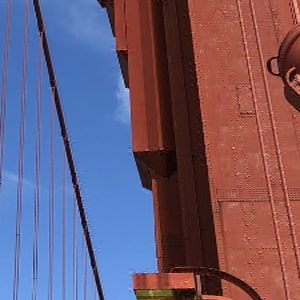}
      & 
      \includegraphics[width=0.3\textwidth]{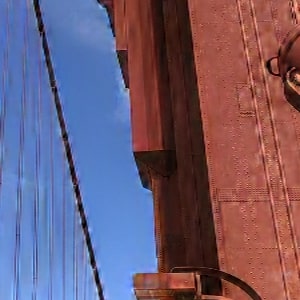}
      & 
      \includegraphics[width=0.3\textwidth]{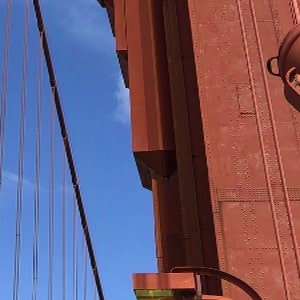}
      \\
      \bottomrule
      \end{tabular}}
      \captionof{figure}{Qualitative comparison of watermarked images using LaWa and other methods on natural images from CLIC~\cite{toderici2020workshop} dataset.
      } \label{fig:supp_fig5}
    \vspace{-20pt}
\end{table*}

\subsection{Image Watermarking Methods}
For DCT-DWt, we used the implementation in~\cite{rivaGANpretrained}. For SSL watermark~\cite{fernandez2022watermarking}, we set the number of optimization steps to 40 and the PSNR to 33. We used the provided pre-trained weights of the method for feature extractor network. For FNNS~\cite{kishore2021fixed}, we set the number of iterations to 10 and use the decoder of Stable Signature~\cite{fernandez2023stable} as the network used for the optimization process of FNNS. For RoSteaALS~\cite{bui2023rosteals}, the public pre-trained weights are for 100-bit watermarks. Therefore, we train two 32-bit and 48-bit versions of the method using the same dataset used for LaWa. Crop attack is not included in the default training process of RoSteaALS and we add this attack during the training. We use the default auto-encoder network used by RoSteaALS. For RivaGan~\cite{zhang2019robust} and HiDDen~\cite{zhu2018hidden}, we use the pre-trained weights provided in~\cite{rivaGANpretrained} and \cite{hiddenpretrained}, respectively. For Stable Signature~\cite{fernandez2023stable}, we follow the provided public code and fine-tune 10 KL-f8 auto-encoders using 10 different messages and create 1k images with each message.

\subsection{Learning-Based Attack}
In this attack, the watermarked image is passed through a pre-trained auto-encoder network in order to remove the watermark information from the reconstructed image by the auto-encoder. We select image compression auto-encoders Bmshj2018~\cite{balle2018variational} and Cheng2020~\cite{cheng2020learned} and use the weights available in CompressAI library zoo~\cite{begaint2020compressai}. We use $q=\{4,8\}$ and $q=\{3,6\}$ for Bmshj2018 and Cheng2020, respectively.
We also use image generation auto-encoders KL-VAE~\cite{rombach2022high} with $f=\{4,8,16,32\}$ and VQ-VQE~\cite{esser2021taming} with $f=\{4,8,16\}$. For these models, the publicly available pre-trained weights of~\cite{stableDiffusion14Git} are used.

\section{Discussion of Limitations}
LaWa addresses the problem of black-box image watermarking, where we assume users do not have access to the generative models. Thus, the main application scenario of LaWa is focused on image generation services. However, model owners may decide to make their models public, where users have white-box access to model weights. Although our intermediate watermarking modules are integrated as extra layers to the model, a hostile user may be able to remove these modules and achieve image generation without watermark. Furthermore, imperceptible image watermarking has generally limited robustness to newer learning-based attacks. We show that LaWa is robust to such attacks for some models. However, with the rapid development of image modification models, a hostile user can use such models to evade the watermark. In addition, learning-based image watermarking methods such as LaWa generally need to apply existing differentiable image distortions at training time. As a result, they may have a poor performance on unknown distortions and image modifications.

\begin{table*}[tb]
     \centering
    \resizebox{0.95\textwidth}{!}{ 
    \begin{tabular}{ c c c  c  c c}
    \toprule
    Original & Watermarked & Residual $\times$ 10 & Original & Watermarked & Residual $\times$ 10 \\
    \midrule
      \includegraphics[width=\setwidth]{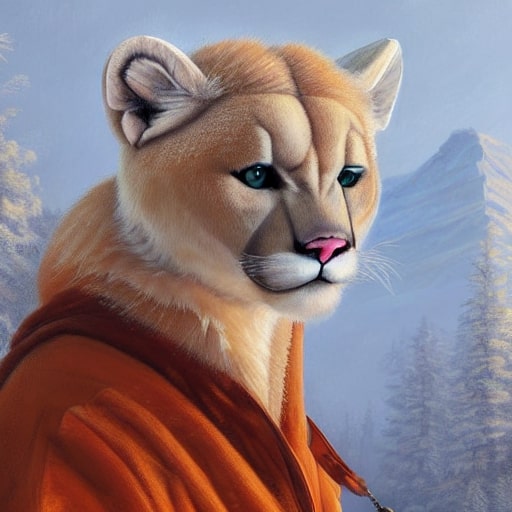}
      &
      \includegraphics[width=\setwidth]{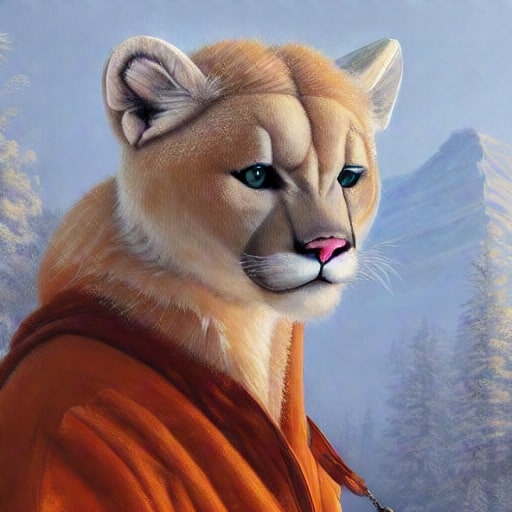}
      &
      \includegraphics[width=\setwidth]{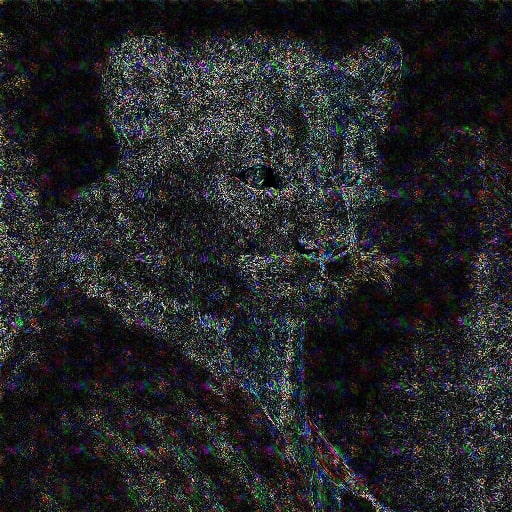}
      &
      \includegraphics[width=\setwidth]{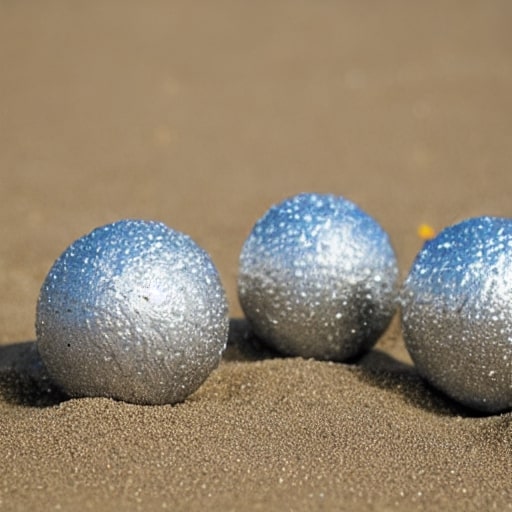}
      &
      \includegraphics[width=\setwidth]{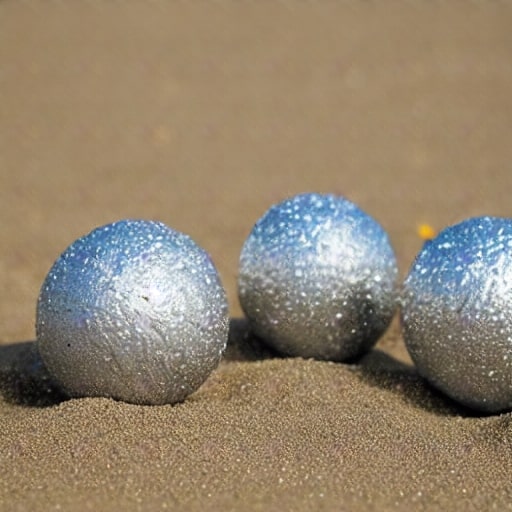}
      &
      \includegraphics[width=\setwidth]{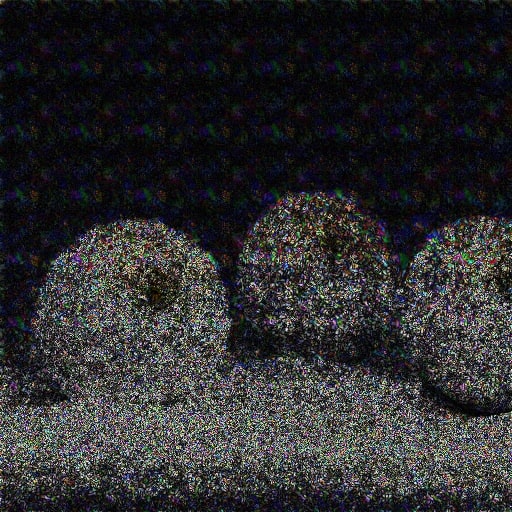}
      \\
      \includegraphics[width=\setwidth]{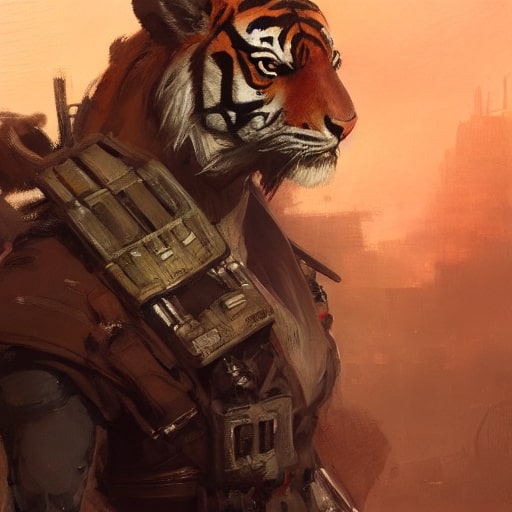}
      &
      \includegraphics[width=\setwidth]{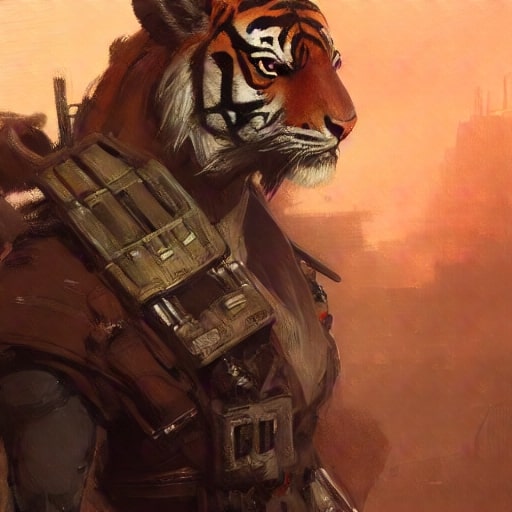}
      &
      \includegraphics[width=\setwidth]{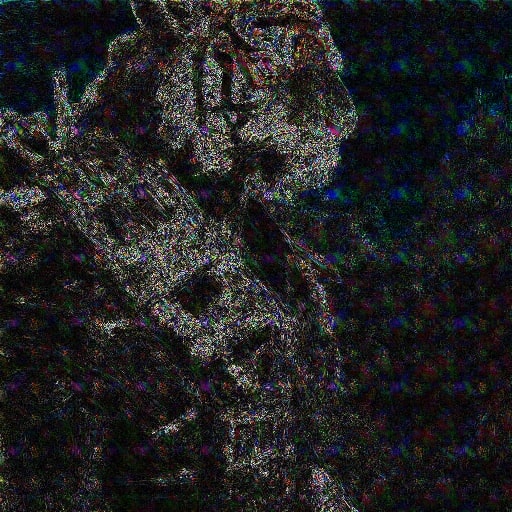}
      & 
      \includegraphics[width=\setwidth]{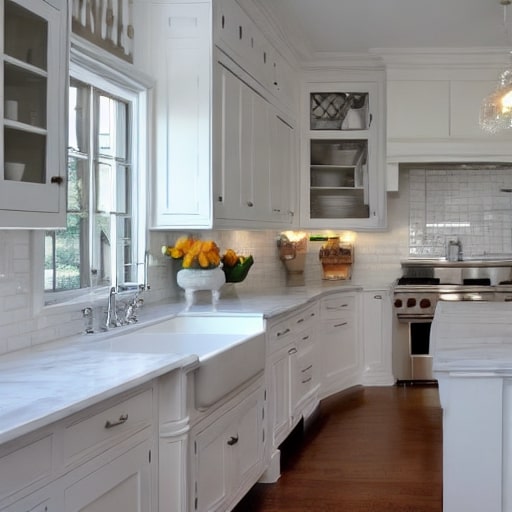}
      &
      \includegraphics[width=\setwidth]{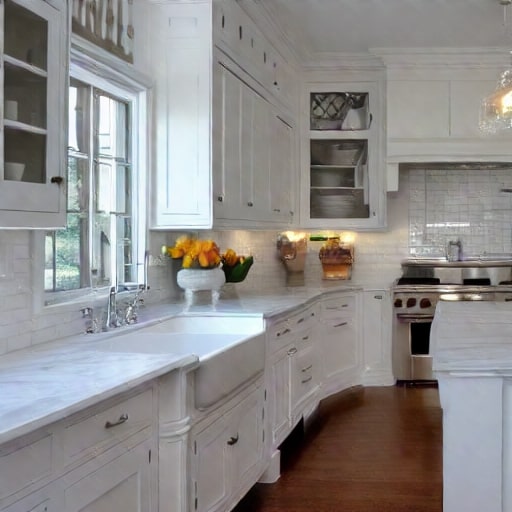}
      &
      \includegraphics[width=\setwidth]{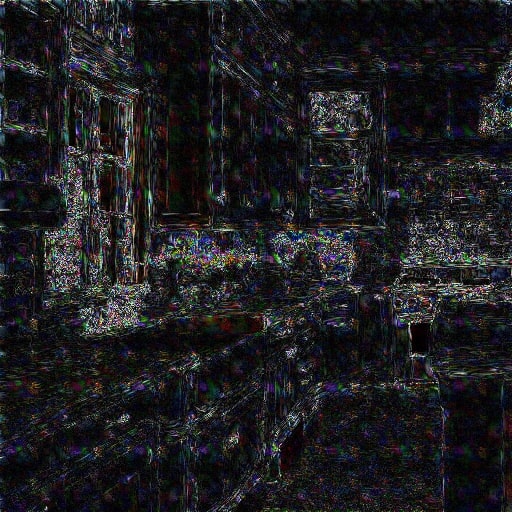}
      \\
      \includegraphics[width=\setwidth]{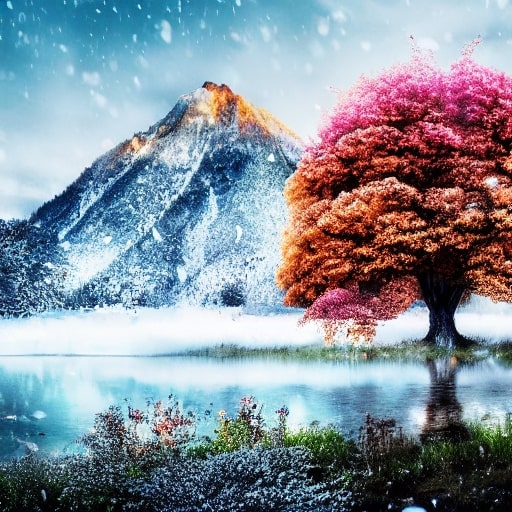}
      &
      \includegraphics[width=\setwidth]{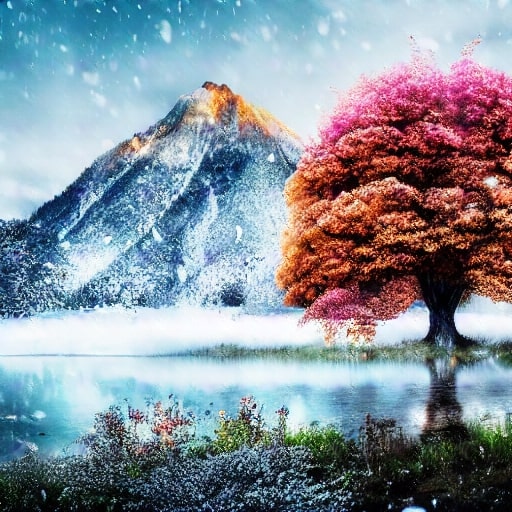}
      &
      \includegraphics[width=\setwidth]{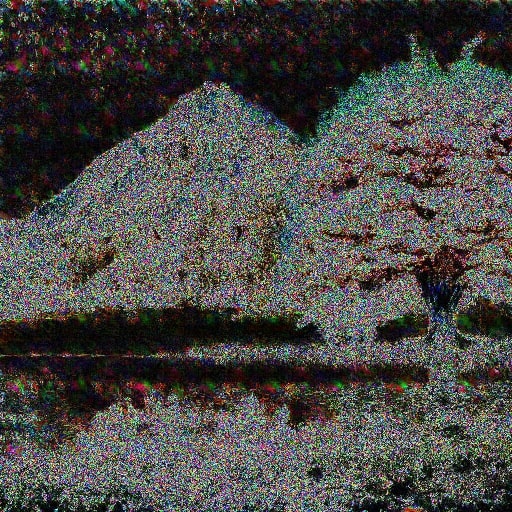}
      & 
      \includegraphics[width=\setwidth]{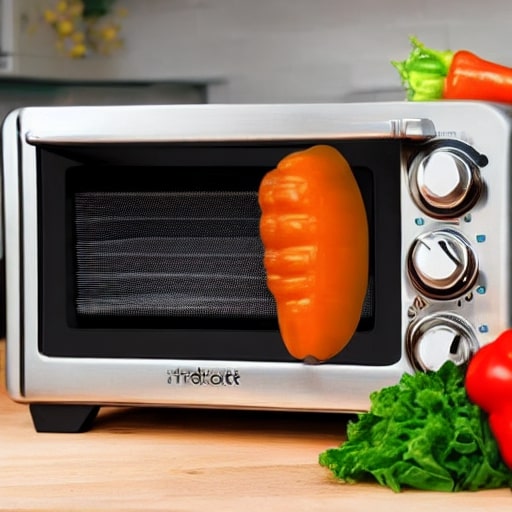}
      &
      \includegraphics[width=\setwidth]{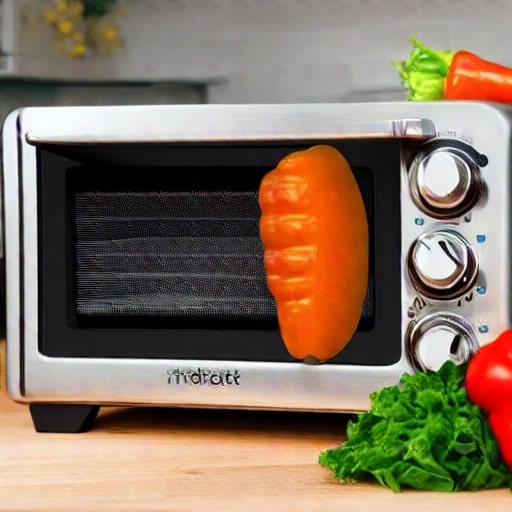}
      &
      \includegraphics[width=\setwidth]{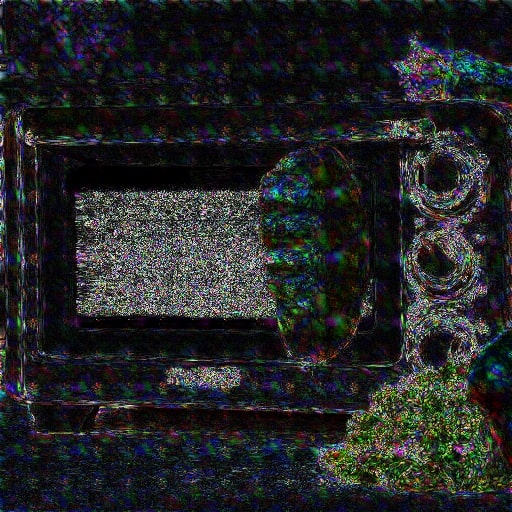}
      \\
      \includegraphics[width=\setwidth]{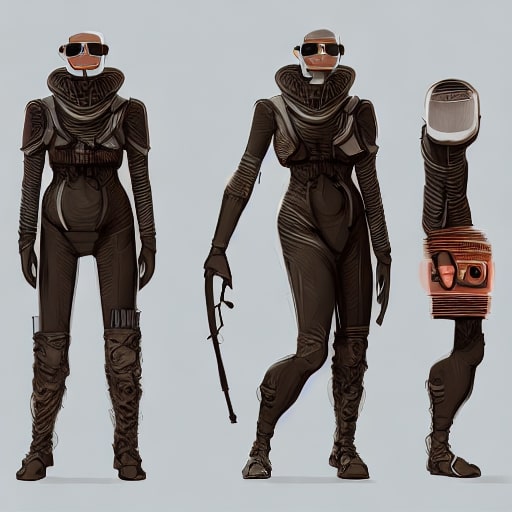}
      &
      \includegraphics[width=\setwidth]{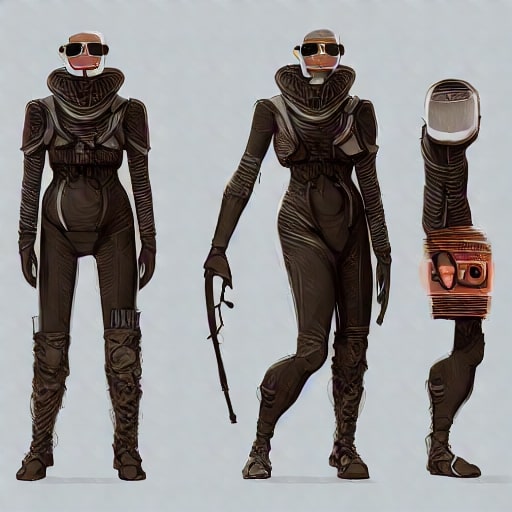}
      &
      \includegraphics[width=\setwidth]{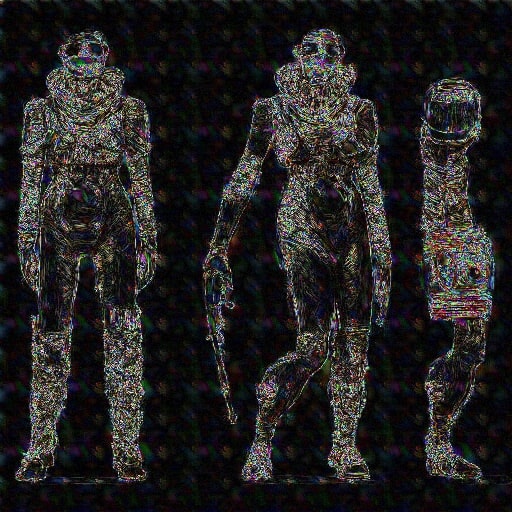}
      & 
      \includegraphics[width=\setwidth]{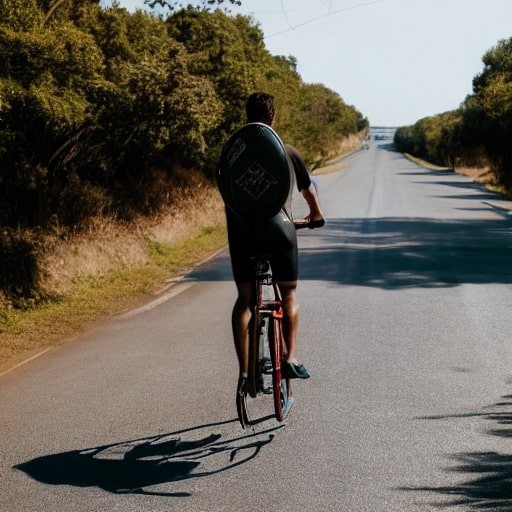}
      &
      \includegraphics[width=\setwidth]{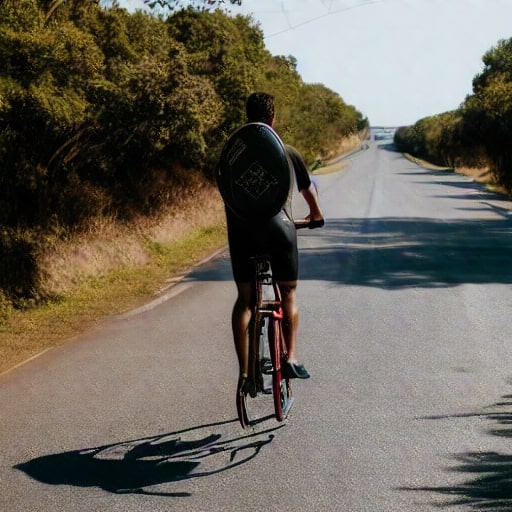}
      &
      \includegraphics[width=\setwidth]{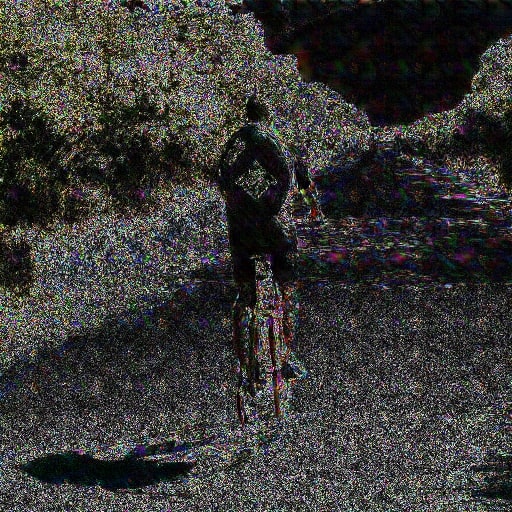}
      \\
      \includegraphics[width=\setwidth]{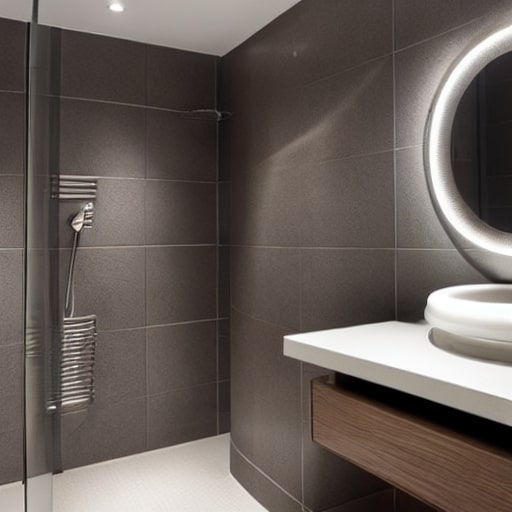}
      &
      \includegraphics[width=\setwidth]{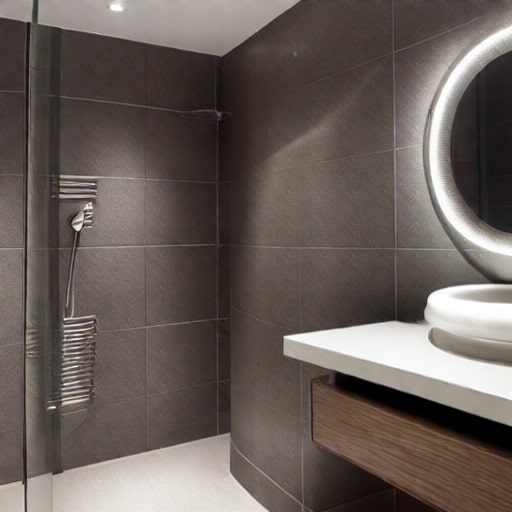}
      &
      \includegraphics[width=\setwidth]{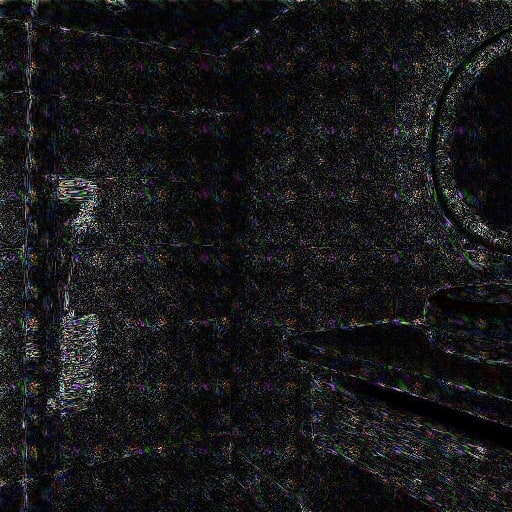}
      & 
      \includegraphics[width=\setwidth]{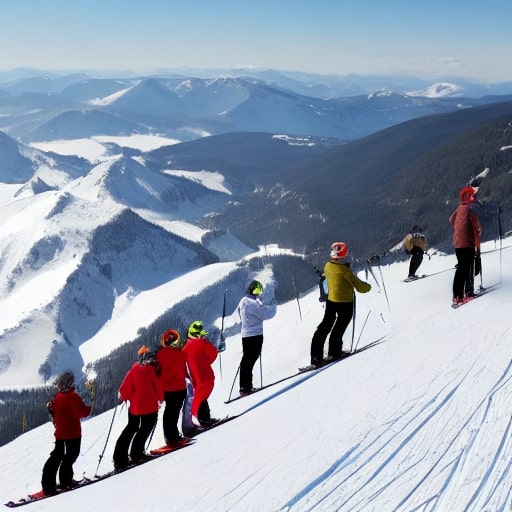}
      &
      \includegraphics[width=\setwidth]{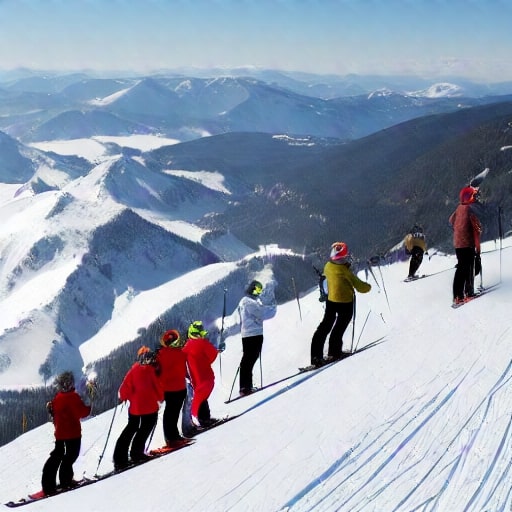}
      &
      \includegraphics[width=\setwidth]{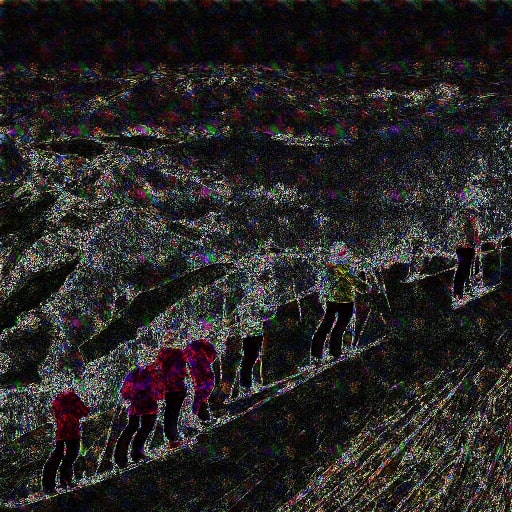}
      \\
      \includegraphics[width=\setwidth]{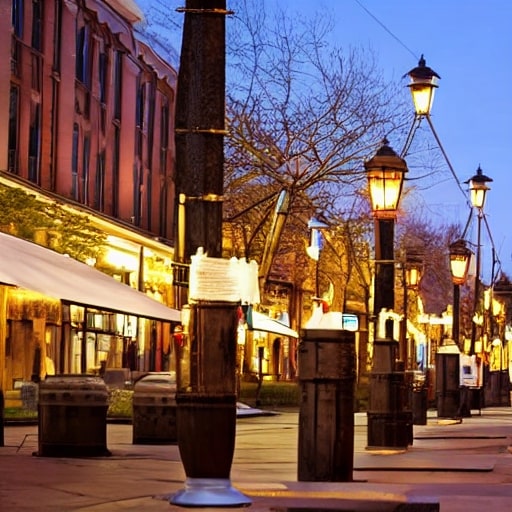}
      &
      \includegraphics[width=\setwidth]{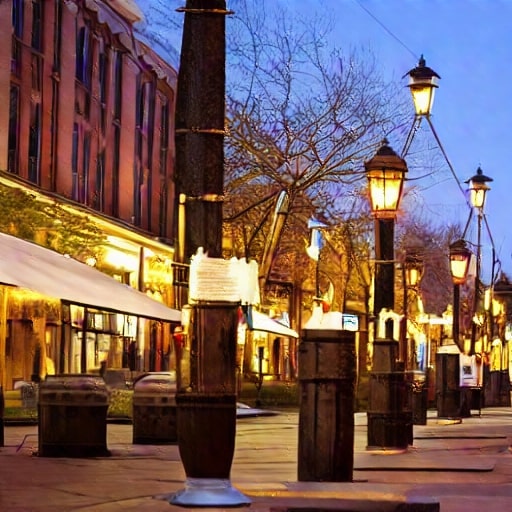}
      &
      \includegraphics[width=\setwidth]{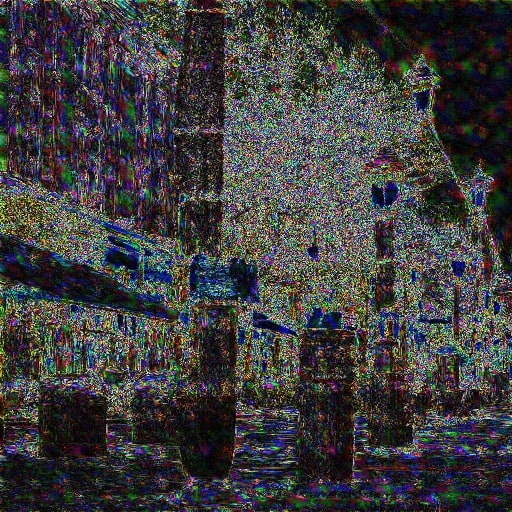}
      & 
      \includegraphics[width=\setwidth]{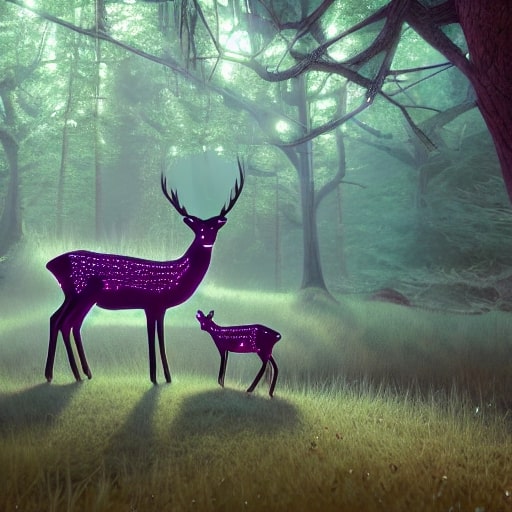}
      &
      \includegraphics[width=\setwidth]{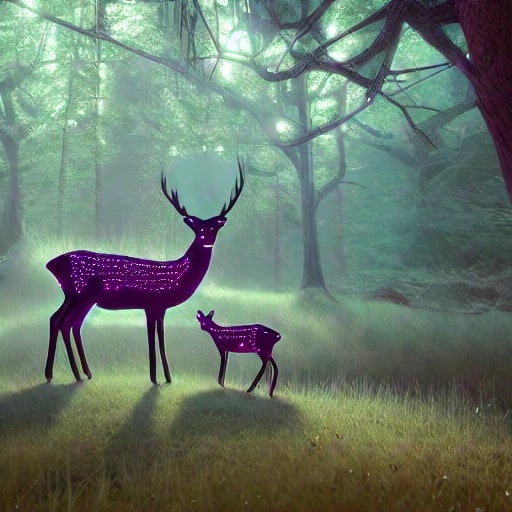}
      &
      \includegraphics[width=\setwidth]{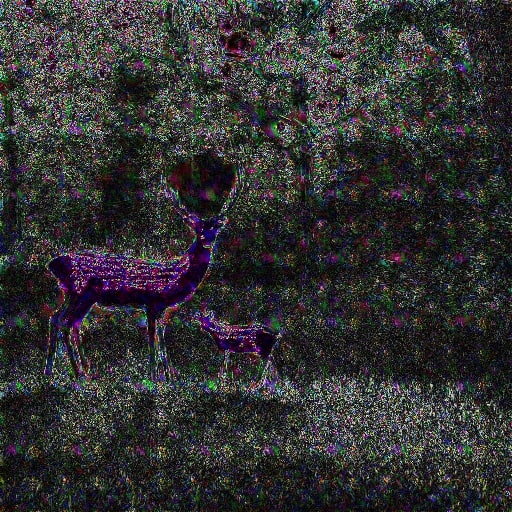}
      \\
      \end{tabular}}
      \captionof{figure}{More qualitative comparison between the original images and the LaWa watermarked
images along with their residuals (multiplied by 10).
      } \label{fig:supp_fig6}
      \vspace{-15pt}
\end{table*}

\end{document}